\pdfoutput=1

\documentclass[11pt]{article}

\usepackage[]{acl}

\usepackage{times}
\usepackage{latexsym}

\usepackage[T1]{fontenc}

\usepackage[utf8]{inputenc}

\usepackage{microtype}

\usepackage{inconsolata}

\usepackage{adjustbox}

\usepackage{graphicx}
\usepackage{scalerel} 
\usepackage{pgfplots}
\usepackage{pgfplotstable}
\usepgfplotslibrary{groupplots}
\usepgfplotslibrary{statistics}
\usetikzlibrary{matrix}
\pgfplotsset{compat=newest}
\usetikzlibrary{pgfplots.fillbetween,patterns}
\usepackage{tikz}
\usetikzlibrary{arrows,automata,calc,shapes, positioning,shadows,shadows.blur,shapes.geometric}
\usepackage{caption}
\usepackage{subcaption}

\usepackage{multirow}
\usepackage{colortbl}
\usepackage{booktabs}
\usepackage{arydshln} 
\usepackage{tabularx} 
\usepackage{threeparttable} 

\usepackage{amsmath}
\usepackage{amssymb}
\usepackage{bm} 
\usepackage{mathtools} 
\usepackage{amsfonts} 
\usepackage{nicefrac} 

\usepackage{pifont}
\usepackage{fontawesome} 
\usepackage{bbding}

\usepackage{cleveref} 
\crefformat{section}{\S#2#1#3}
\crefformat{subsection}{\S#2#1#3}
\crefformat{subsubsection}{\S#2#1#3}
\crefformat{appendix}{\S#2#1#3}

\usepackage{mdframed}

\usepackage{enumitem} 

\usepackage{soul}
\usepackage[most]{tcolorbox}
\definecolor{teal1}{HTML}{00426A}
\definecolor{teal2}{HTML}{008080}
\definecolor{teal3}{HTML}{64C989}
\definecolor{checkmarkgreen}{HTML}{008000}
\definecolor{hlyellow}{HTML}{F4F6D5}
\definecolor{hlcyan}{HTML}{C9EFF8}
\definecolor{hlgreen}{HTML}{DAF6DB}
\definecolor{hlpurple}{HTML}{F8DEF6}
\definecolor{plotbackground}{HTML}{E6E6E6}
\definecolor{lightgrey}{HTML}{F2F2F2}
\definecolor{darkgrey}{HTML}{404040}
\definecolor{electricblue}{HTML}{00AEEF}
\definecolor{purple}{HTML}{BF0040}
\definecolor{warm1}{HTML}{621708}
\definecolor{warm2}{HTML}{9A0070}
\definecolor{warm3}{HTML}{CE030A}
\definecolor{warm4}{HTML}{BB3E03}
\definecolor{warm5}{HTML}{F48C06}
\definecolor{underestimated}{HTML}{235E95}
\definecolor{overestimated}{HTML}{BE533B}

\NewDocumentCommand\medgenreadIcon{}{\resizebox{!}{3.55mm}{\includegraphics{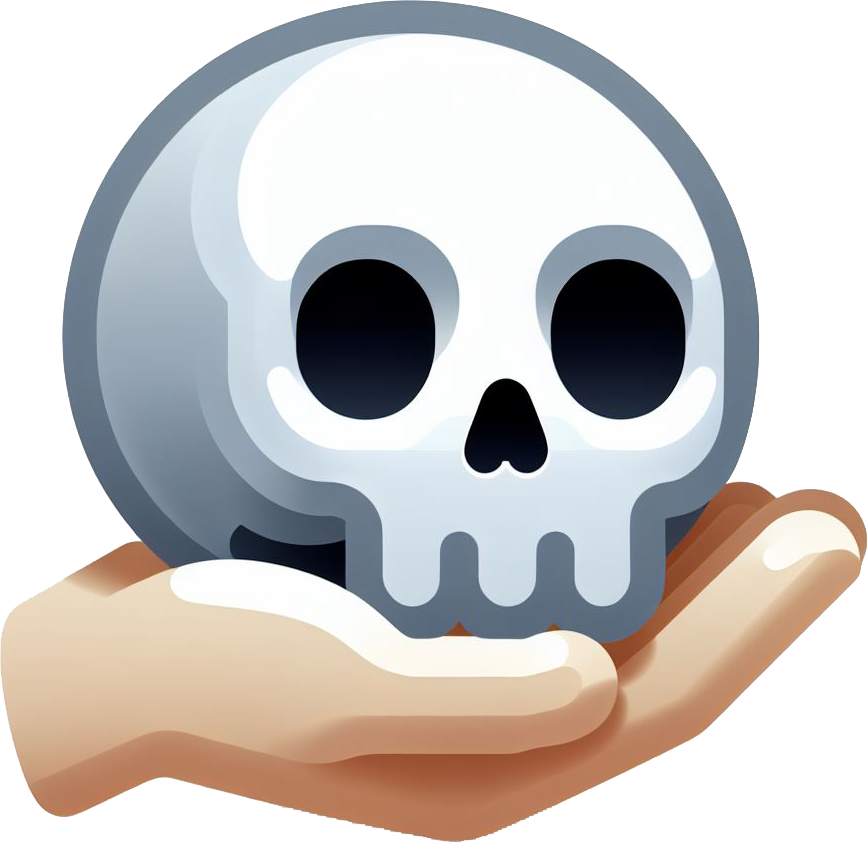}}}

\DeclareRobustCommand{\hlyellow}[1]{{\sethlcolor{hlyellow}\hl{#1}}}
\DeclareRobustCommand{\hlcyan}[1]{{\sethlcolor{hlcyan}\hl{#1}}}
\DeclareRobustCommand{\hlgreen}[1]{{\sethlcolor{hlgreen}\hl{#1}}}
\DeclareRobustCommand{\hlpurple}[1]{{\sethlcolor{hlpurple}\hl{#1}}}

\DeclareMathOperator*{\OPLUS}{\oplus}

\newcommand*\samethanks[1][\value{footnote}]{\footnotemark[#1]}

\newmdenv[backgroundcolor=lightgrey, 
          linecolor=lightgrey,
          innerleftmargin=10pt, 
          innerrightmargin=10pt, 
          innertopmargin=10pt, 
          innerbottommargin=10pt,
          skipabove=10pt,
          skipbelow=10pt,
          roundcorner=5pt,
          font=\small]{graybox}

\title{{\medgenreadIcon}\hspace{1.5mm}To Generate or to Retrieve? On the Effectiveness of Artificial Contexts\\for Medical Open-Domain Question Answering}

\author{Giacomo Frisoni$^{\spadesuit}$\thanks{\hspace{.04in}Equal contribution (co-first authorship).}\ \ \ \  \textbf{Alessio Cocchieri}$^{\spadesuit}$\samethanks\ \ \ \  \textbf{Alex Presepi}$^{\spadesuit}$\\\textbf{Gianluca Moro}$^{\spadesuit}$\samethanks\ \ \ \  \textbf{Zaiqiao Meng}$^{\diamondsuit}$\\
$^\spadesuit$Department of Computer Science and Engineering, University of Bologna \\
$^\diamondsuit$School of Computing Science, University of Glasgow \\
\texttt{$^\spadesuit$\{giacomo.frisoni, a.cocchieri, gianluca.moro\}@unibo.it}\\
\texttt{$^\spadesuit$alex.presepi@studio.unibo.it}\quad
\texttt{$^\diamondsuit$zaiqiao.meng@glasgow.ac.uk}}

\begin{document}
\maketitle
\begin{abstract}
Medical open-domain question answering demands substantial access to specialized knowledge. Recent efforts have sought to decouple knowledge from model parameters, counteracting architectural scaling and allowing for training on common low-resource hardware. The retrieve-then-read paradigm has become ubiquitous, with model predictions grounded on relevant knowledge pieces from external repositories such as PubMed, textbooks, and UMLS. An alternative path, still under-explored but made possible by the advent of domain-specific large language models, entails constructing artificial contexts through prompting. As a result, "to generate or to retrieve" is the modern equivalent of Hamlet's dilemma. This paper presents \textsc{MedGENIE}, the first generate-then-read framework for multiple-choice question answering in medicine. We conduct extensive experiments on MedQA-USMLE, MedMCQA, and MMLU, incorporating a practical perspective by assuming a maximum of 24GB VRAM. \textsc{MedGENIE} sets a new state-of-the-art in the open-book setting of each testbed, allowing a small-scale reader to outcompete zero-shot closed-book 175B baselines while using up to 706$\times$ fewer parameters. Our findings reveal that generated passages are more effective than retrieved ones in attaining higher accuracy.\footnote{Our code, fine-tuned models, and generated contexts are publicly available at \href{https://github.com/disi-unibo-nlp/medgenie}{https://github.com/unibo-nlp/medgenie}.}
\end{abstract}

\begin{figure}[h]
    \centering
    \includegraphics[width=\linewidth]{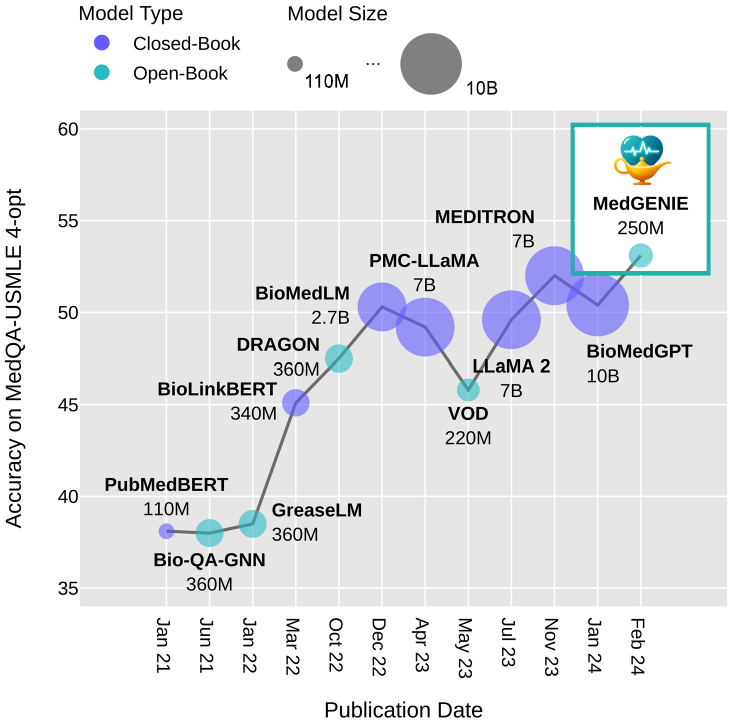}
    \caption{\textsc{MedGENIE} performance (Flan-T5-base, Fusion-In-Decoder) on USMLE-style questions. Comparison against fine-tuned open-source baselines with a maximum of 10B parameters, using the MedQA (4 options) test set. Model size displayed on a log scale.}
    \label{fig:abstract}
\end{figure}

\section{Introduction}

Question answering is a challenging task that requires complex reasoning on explicit constraints described in the question and unstated domain knowledge.
Open-domain question answering (ODQA) aims to tackle natural questions across various topics without predefined evidence \citep{chen-etal-2017-reading}.
This setting mirrors real-world scenarios where there cannot be a labeled passage for every potential user inquiry.
ODQA has particular significance in medicine due to the high-quality standards it demands, including a deep understanding of specialized terminology and background concepts, and an effective recall of expert insight for clinical decision making \citep{frisoni-etal-2022-bioreader}.

Recent efforts have transitioned from a \textit{closed-book} strategy, where models rely solely on their opaque parametric knowledge, to an \textit{open-book} alternative, allowing them to consult external sources for grounding.
In particular, the \textit{retrieve-then-read} framework is a common thread \citep{DBLP:journals/corr/abs-2101-00774,zhang-etal-2023-survey-efficient}, where the input is augmented with relevant knowledge chunks \textit{retrieved} from an external datastore, which can be unstructured (e.g., PubMed, textbooks) or structured (e.g., UMLS).
However, performance is highly dependent on the quality of the retriever.
Developing custom retrieval modules generally requires extensive question--context pairs \citep{karpukhin-etal-2020-dense} or intensive computational resources \citep{lee-etal-2019-latent}, particularly when dealing with massive sources \citep{DBLP:conf/kdd/GanGZSXXHCWS23}. 
Furthermore, the retrieved fragments may be incomplete or not specifically tailored to the query, leading to noise \citep{oh-thorne-2023-detrimental}.

In parallel, medical large language models (LLMs) have gained increasing research interests to aid professionals and improve patient care \citep{DBLP:journals/corr/abs-2311-05112}.
After pre-training on an extreme-scale collection of specialized text corpora, they implicitly encode an impressive amount of domain knowledge that can be evoked through prompting, akin to summoning a genie from a lamp.
This facilitates a paradigm shift towards a \textit{generate-then-read} approach, wherein contexts are directly \textit{generated} by an LLM.
Despite preliminary work \citep{DBLP:conf/iclr/0002IWXJ000023,su-etal-2023-context,zhang-etal-2023-merging}, there is an ongoing debate on ``whether generative augmentation is preferable to retrieval augmentation''.

In this paper, we introduce \textsc{MedGENIE}, the first \textit{generate-then-read} framework for multiple-choice medical ODQA.
Specifically, we study the effectiveness of grounding generalist LLMs and small language models (SLMs) on multi-view contexts generated by a medical LLM via in-context learning (ICL) and fine-tuning, respectively.
To foster accessibility and match prevalent hardware configurations, we assume a low-resource infrastructure with 24GB VRAM.

We evaluate \textsc{MedGENIE} on three standard ODQA benchmarks designed to quantify professional medical competencies: MedQA-USMLE \citep{DBLP:journals/corr/abs-2009-13081}, MedMCQA \citep{DBLP:conf/chil/PalUS22}, and MMLU-Medical \citep{DBLP:conf/iclr/HendrycksBBZMSS21}.
\textsc{MedGENIE} demonstrates significant performance gains, improving the accuracy of few-shot LLM readers on all testbeds by up to $\approx$16 points.
By fine-tuning the reader, \textsc{MedGENIE} allows Flan-T5-base to outcompete closed-book zero-shot 175B LLMs and supervised 10B baselines on MedQA (Figure~\ref{fig:abstract}), using up to 706$\times$ fewer parameters.
Furthermore, our research demonstrates a clear inclination of cutting-edge rerankers towards favoring generated contexts over retrieved ones.
When treated as knowledge sources or incorporated into human-curated ones, generated passages notably enhance the effectiveness of retrieve-then-read workflows (up to $\approx$6 extra points). RAGAS evaluation confirms the quality of generated contexts, even allowing for more faithful answers from the reader. Finally, we release a comprehensive dataset of $\approx$1 million artificial contexts in the medical field, adhering to principles of open science and encouraging further research endeavors.

\section{Related Work}

\paragraph{Medical Language Models}
Transformer-based language models have become the go-to solution for any task in medical NLP.
Early work on domain adaption focused on encoder-only models pre-trained on PubMed articles and abstracts, counting 40+ models introduced between 2019 and 2021 \citep{DBLP:journals/jbi/KalyanRS22}.
With the recent trend of scaling up pre-training data and model parameters, multiple studies have delved into medical LLMs, transitioning to decoder-only architectures and generative tasks \citep{DBLP:journals/corr/abs-2310-05694,DBLP:journals/corr/abs-2311-05112}.
Popular open-source milestones include ChatDoctor \citep{DBLP:journals/corr/abs-2303-14070}, PMC-LLaMA \citep{DBLP:journals/corr/abs-2304-14454}, and MEDITRON \citep{DBLP:journals/corr/abs-2311-16079}.
"Pre-train, prompt, and predict" is today's prevalent methodology for applying LLMs to new problems, circumventing the need for extensive fine-tuning on human-labeled examples, which are rarely accessible in clinical scenarios.
We are the first to examine the prowess of medical LLMs in generating background context for ODQA without supervision.

\paragraph{Open-Book Question Answering}
Closed-book LLMs, such as GPT-IV and Med-PaLM-2 540B, exhibit impressive results in ODQA with fine-tuning \citep{DBLP:journals/corr/abs-2305-09617} or prompting \citep{DBLP:journals/corr/abs-2303-13375,DBLP:journals/corr/abs-2311-16452} techniques, performing non-trivial reasoning steps \citep{DBLP:journals/corr/abs-2207-08143}.
As LLMs grow to achieve predictable accuracy gains \citep{DBLP:journals/corr/abs-2001-08361}, their computational demands exceed the resources of most product teams.
To counteract the scaling trend, reach satisfactory results with commodity hardware, and ensure greater control over knowledge, various open-book contributions have surfaced in medicine, comprising BioReader \citep{frisoni-etal-2022-bioreader}, DRAGON \citep{DBLP:conf/nips/YasunagaBR0MLL22}, and VOD \citep{DBLP:conf/icml/LievinMJW23}.
Nevertheless, to date, generate-then-read pipelines have predominantly undergone scrutiny solely within the general domain.
GenRead \citep{DBLP:conf/iclr/0002IWXJ000023} and CGAP \citep{zhang-etal-2023-merging} are among the first to demonstrate the efficacy of LLMs as strong context generators, focusing on datasets like Natural Questions, TriviaQA, and WebQuestions. They utilized LLMs such as InstructGPT (175B) and Megatron (530B) both as generators and readers. GenRead introduced a clustering-based prompting technique involving context generation, filtering, encoding, $K$-means clustering, and per-cluster sampling. While we acknowledge the efficacy of these techniques, as demonstrated in ad-hoc ablations (Appendix \ref{app:clustering_based_prompting}), we caution against their feasibility in real-world scenarios due to associated high costs.
Additionally, investigations into the fusion of retrieved and LLM-generated contexts for ODQA, exemplified by COMBO \citep{su-etal-2023-context}, have shown promising results in enhancing performance across aforementioned benchmarks. However, their approach entails utilizing two discriminators trained on silver labels to compute compatibility scores, with InstructGPT and ChatGPT as context generators and a FiD Flan-T5-large as the reader, scaling up to 10 contexts using four A40 GPUs, each equipped with 46 GB VRAM.

\section{Method}
\label{sec:method}

\begin{figure*}[h]
    \centering
    \includegraphics[width=\linewidth]{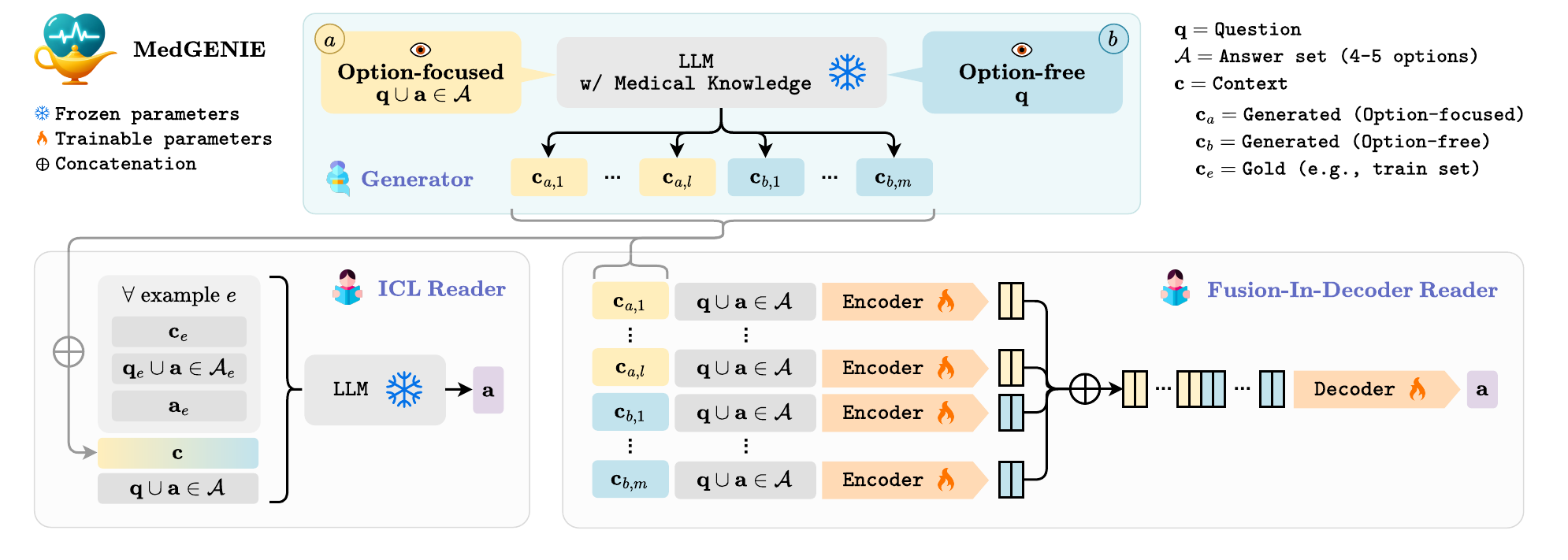}
    \caption{Overview of the \textsc{MedGENIE} framework. It generates multi-view artificial contexts with a specialized LLM (top), and then uses them to ground a prompted LLM or a fine-tuned SLM (bottom).}
    \label{fig:medgenie_pipeline}
\end{figure*}

In this section, we discuss our \textsc{MedGENIE} framework, illustrated in Figure~\ref{fig:medgenie_pipeline}. First, it prompts a medical LLM to furnish multi-view background contexts for a given question (\cref{subsec:context_gen}). Then, it diverges into two distinct paths depending on the reader type: ICL for LLMs (\cref{subsec:icl}), and Fusion-In-Decoder (FID) fine-tuning for SLMs (\cref{subsec:fid}). Both strategies equip readers with custom background passages, allowing them to tackle medical questions effectively even without prior knowledge.

\paragraph{Problem Statement}

In multiple-choice ODQA, a dataset record consists of a question $\mathbf{q}$ and an answer set $\mathcal{A}{=}\{\mathbf{a}_1, \dots, \mathbf{a}_{|\mathcal{A}|}\}$, all expressed in text.
Our goal is to identify which answer $\mathbf{a} \in \mathcal{A}$ is correct ($\hat{\mathbf{a}}$).
A closed-book solution would estimate $p_{\theta}(\mathbf{a}|\mathbf{q},\mathcal{A})$ parameterized by $\theta$, leaving considerable domain knowledge unexploited \citep{DBLP:journals/corr/abs-2204-10019}.
Conversely, we assume access to a set of record-specific artificial contexts and add an auxiliary variable $\mathcal{C}{=}\{\mathbf{c}_1, \dots, \mathbf{c}_{|\mathcal{C}|}\}$.
We thus move to a marginal likelihood with a reader-generator model: $p_{\theta}(\mathbf{a}|\mathbf{q},\mathcal{A}){=}\sum_i p_{\theta}(\mathbf{a}|\mathcal{C}_i,\mathbf{q},\mathcal{A})p_{\theta}(\mathcal{C}_i|\mathbf{q},\mathcal{A})$.
Since we cannot sum over all possible contexts, we approximate the maximum a posteriori estimation with a decoding strategy, using a single value for~$\mathcal{C}$.

\subsection{Multi-view artificial contexts}
\label{subsec:context_gen}

\begin{figure}[t]
    \centering
    \includegraphics[width=\linewidth]{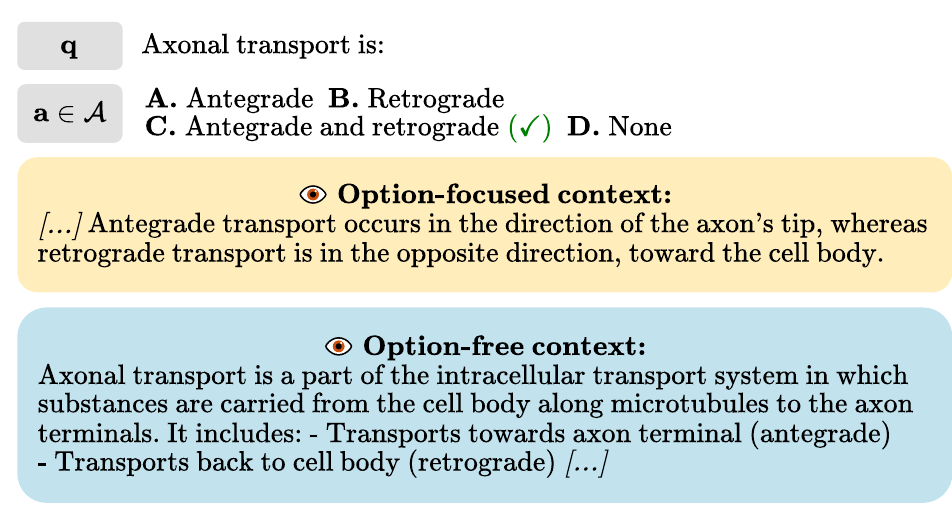}
    \caption{Example of multi-view context generation for a MedMCQA eval instance. The knowledge verbalized by a medical LLM is highly valuable in determining the correct answer (unseen by the generator).}
    \label{fig:multiview_context}
\end{figure}

In the first stage, we prompt a medical LLM to output $\mathcal{C}$.
The primary challenge in generating multiple synthetic contexts for the same question is to promote diversity and knowledge coverage while avoiding repetitive information.
Addressing this concern, \citet{DBLP:conf/iclr/0002IWXJ000023} proposed a three-step approach: (i) building a supporting repository with an artificial document for each $\mathbf{q}$ of the training set, (ii) encoding artificial documents and clustering them, (iii) given a training or inference query, sampling various question--document pairs from each cluster, and using them as ICL demonstrations to craft a series of topic-guided contexts.
Although we acknowledge that clustering-based prompting can amplify the effectiveness of \textsc{MedGENIE} (Appendix~\cref{app:clustering_based_prompting}), it introduces complexity, largely impacting time and memory costs.
To streamline the process and reduce resource constraints, we devise a multi-view prompt technique, dividing $\mathcal{C}$ into two segments, $\mathcal{C}_a$ and $\mathcal{C}_b$.
We first ask the LLM to generate a set of contexts $\{\mathbf{c}_{a,1}, \dots, \mathbf{c}_{a,l}\}$ conditioned on $\mathbf{q} \cup \mathbf{a} \in \mathcal{A}$ (\textit{option-focused}), and then a set of contexts $\{\mathbf{c}_{b,1}, \dots, \mathbf{c}_{b,m}\}$ conditioned solely on $\mathbf{q}$ (\textit{option-free}).
The former aims to provide information for each candidate answer, whereas the latter targets a broader context, encompassing pertinent domain concepts that could aid reasoning.
Figure \ref{fig:multiview_context} clarifies their distinction.

We rely on few-shot prompting \citep{DBLP:conf/nips/BrownMRSKDNSSAA20} to guide the LLM on two examples $\mathcal{E}$ meticulously curated by a healthcare professional in distinct and representative domains:
\begin{equation}
Pr_a{=}\left[\OPLUS_e^{|\mathcal{E}|} \left[ \mathbf{q}_{e} \oplus \mathcal{A}_e \oplus \mathbf{c}_e \right]\right] \oplus \mathbf{q} \oplus \mathcal{A}
\end{equation}
\begin{equation}
Pr_b{=}\left[\OPLUS_e^{|\mathcal{E}|} \left[ \mathbf{q}_{e} \oplus \mathbf{c}_e \right]\right] \oplus \mathbf{q}
\end{equation}
where $Pr_a$ is the option-focused prompt, $Pr_b$ the option-free one, and $\oplus$ denotes concatenation.
Full prompt templates are listed in Appendix \cref{app:prompt_templates}.

To promote intra-context diversity, we (i) use a random sampling decoding strategy with high temperature, ensuring a wider exploration of the model's output space, and (ii) penalize new tokens based on their frequency in the prompt and the inferred text. Details in Appendix~\cref{app:implementation_details}.

According to in-depth ablation studies (Appendix~\cref{app:contexts_number_type}), we set $l{=}3$ and $m{=}2$, corroborating the importance of each view.
Importantly, we preprocess each context by removing any answer-guessing phrase that could bias the reader.

\subsection{ICL reader (Unsupervised)}
\label{subsec:icl}

For the unsupervised setting, we use ICL which does not need the complete training data but only a few demonstration examples.
In particular, we feed an LLM reader with few-shot ODQA demonstrations and the test query preceded by its artificial context.
Inferring the prompt continuation indirectly accomplishes the \textit{unseen} ODQA task.
Mechanically, we build the following two-shot prompt:
\begin{equation}
\begin{aligned}
\left[\OPLUS_e^{|\mathcal{E}|} \mathbf{c}_e \oplus \mathbf{q}_e \oplus \mathcal{A}_e \oplus \hat{\mathbf{a}}_e \right] \oplus \mathbf{c} \oplus \mathbf{q} \oplus \mathcal{A}
\end{aligned}
\end{equation}
where $\mathcal{E}$ encloses two train instances of the target benchmark, and $\mathbf{c}_e$ is an option-focused context from the medical LLM---proved to be more effective than a human sample (Appendix~\cref{app:icl_example_selection}).
$\mathbf{c}$ is the concatenation of option-focused and option-free contexts from \cref{subsec:context_gen}.
Consistently, we assign the initial positions within $\mathbf{c}$ to option-focused contexts due to their higher probability of containing the correct answer. This conjecture is further validated by the outcomes of our experiments in \cref{subsec:gen_vs_ret}. Our design choice wants to prevent $\mathcal{C}_a$ from "getting lost in the middle" \citep{DBLP:journals/corr/abs-2307-03172}. Indeed, LLMs are prone to difficulty extracting relevant information when it is buried within the central portion of a lengthy context. Contrarily, accuracy improves when pertinent information is situated at either the beginning or end of the input context.

\subsection{Fine-tuned reader (Supervised)}
\label{subsec:fid}

While the ICL methodology unlocks the utilization of off-the-shelf LLMs and harnesses their innate language understanding abilities, it mandates loading billions of weights into memory, making inference alone expensive.
LLM readers could remarkably benefit from additional fine-tuning steps, but training generally requires up to 20 extra bytes per parameter, easily evolving prohibitive.
Therefore, for the supervised regime, we train a lightweight FID reader \citep{izacard-grave-2021-leveraging}.
Each ${<}\mathbf{c}, \mathbf{q}, \mathcal{A}{>}$ tuple is joined with special separator tokens, and processed independently by the encoder.
The representations produced by the last encoder layer are then concatenated and sent to the decoder layers.
In this way, the computation cost grows linearly with $|\mathcal{C}|$.
We keep the relative order of contexts fixed across all input pairs, always putting $\mathcal{C}_a$ in front of $\mathcal{C}_b$.

\section{Experimental Setup}

\subsection{Benchmarks}

We evaluate \textsc{MedGENIE} on three multiple-choice ODQA benchmarks (Table~\ref{table:benchmarks}), requesting multi-hop reasoning capabilities, and human-level reading comprehension skills.

\paragraph{MedQA-USMLE \citep{DBLP:journals/corr/abs-2009-13081}} English-language subset of MedQA tailored for medical license examination in the United States. It amalgamates various knowledge facets, such as patient profiles, disease symptoms, and drug dosage requirements. Each question entails a medical case history, vital signs (e.g., blood pressure, temperature), and eventual diagnostic evaluations (e.g., CT scan). It exists in two versions: 4 and 5 options. Following \citet{DBLP:journals/corr/abs-2311-16079}, in the supervised pipeline, we fine-tune the reader over the original 5-option train set for fair comparisons.

\paragraph{MedMCQA \citep{DBLP:conf/chil/PalUS22}} Highly comprehensive dataset sourced from Indian medical school entrance exams (AIIMS, NEET-PG). It features mixed question types, such as diagnosis and mathematical problems. Due to the absence of public answer labels for the test set, we employ the eval set as the main reference, adhering to prior literature \citep{DBLP:journals/corr/abs-2207-08143,DBLP:journals/corr/abs-2212-13138,DBLP:journals/corr/abs-2311-16452,DBLP:journals/corr/abs-2311-16079,DBLP:journals/corr/abs-2304-14454}.

\paragraph{MMLU-Medical \citep{DBLP:conf/iclr/HendrycksBBZMSS21}} MMLU is a multi-task zero-shot benchmark suite of 57 subsets spanning STEM, humanities, and social sciences. In the footsteps of \citet{DBLP:journals/corr/abs-2212-13138,DBLP:conf/icml/LievinMJW23,DBLP:journals/corr/abs-2311-16079}, we limit our analysis to 9 medically relevant subsets. As MMLU lacks training data, we fine-tune the FID reader over the MedMCQA train set and gauge out-of-domain generalization, securing fairness with \citet{DBLP:conf/icml/LievinMJW23,DBLP:journals/corr/abs-2311-16079}.

\vspace{2mm}

\textit{Terminology}. Throughout the remainder of this paper, we use the terms "MedQA" and "MMLU" to denote their medical subsets, aiming for succinctness. With regard to MedQA, unless otherwise specified, we refer to the common 4-option variant.

\begin{table}[!htb]
\centering
\adjustbox{width=\linewidth}{
\begin{tabular}{llll}
\hline
& \begin{tabular}{@{}l}\textbf{MedQA}\end{tabular} & \textbf{MedMCQA} & \begin{tabular}{@{}l}\textbf{MMLU}\end{tabular} \\
\hline
\# Train & 10,178 & 187,000 & -- \\
\# Eval & 11,450 & 4183 & -- \\
\# Test & 1273 & -- & 1862 \\
$|\mathcal{A}|$ & 4-5 & 4 & 4 \\
\# Subjects & Not labeled & 21 & 9 \\
\# Words / $\mathbf{q}$ & 116.6 & 12.7 & 36.2 \\
\hline
\end{tabular}}
\caption{Summary of the medical benchmarks.}
\label{table:benchmarks}
\end{table}

\subsection{Medical-expert generator}

Following exhaustive preliminary experiments (Appendix \cref{app:medical_llm_choice}), we select PMC-LLaMA-13B as the specialized generator model \citep{DBLP:journals/corr/abs-2304-14454}.\footnote{\href{https://huggingface.co/axiong/PMC\_LLaMA\_13B}{axiong/PMC\_LLaMA\_13B}}
Technically, PMC-LLaMA performs a two-step training above LLaMA \citep{DBLP:journals/corr/abs-2302-13971}. First, it infuses medical knowledge by continuing the pre-training on 4.8M academic papers and 30K medical books. Second, it undergoes medical instruction tuning on question answering, rationale, and conversation, also exploiting the MedQA and MedMCQA train sets.
Although the model has been exposed to the training data of some benchmarks, it is imperative to note that it was not explicitly trained to forge question-conditioned contexts.
Pre-aligning the selected medical expert generator towards the benchmark data is not strictly necessary.
The essential aspect is the latent knowledge encoded within the model parameters, which is frequently correlated with their quantity \citep{DBLP:conf/nips/TirumalaMZA22,DBLP:conf/iclr/CarliniIJLTZ23}.
\citet{DBLP:journals/corr/abs-2311-16452} operated CoT and self-consistency prompting in conjunction with major voting to recall knowledge within generalist foundational models and surpass domain-specialized counterparts. However, these methodologies demonstrate efficacy when applied to the largest versions of a foundational model \citep{DBLP:journals/corr/abs-2212-13138,DBLP:journals/corr/abs-2207-08143,DBLP:journals/corr/abs-2311-16079}, preferring domain-specific solutions at smaller magnitudes. In light of this, we opt for the largest medical LLM runnable on consumer GPU hardware. The average length of a single PMC-LLaMA context is $\approx$200 words; see Appendix~\cref{app:context_distributional_stats} for distributional statistics.

\subsection{Readers}

\textsc{MedGENIE} is \textit{reader-agnostic} but, for this study, we adopt the following implementation choices.
\begin{itemize}[noitemsep, topsep=0pt]
    \item \textbf{ICL}. We test LLaMA-2-chat (7B) \citep{DBLP:journals/corr/abs-2302-13971}\footnote{\href{https://huggingface.co/meta-llama/Llama-2-7b-chat-hf}{meta-llama/Llama-2-7b-chat-hf}} and Zephyr-$\beta$ (7B) \citep{DBLP:journals/corr/abs-2310-16944},\footnote{\href{https://huggingface.co/HuggingFaceH4/zephyr-7b-beta}{HuggingFaceH4/zephyr-7b-beta}} both supporting a context window length of 4K tokens. Additionally, we examine more recent models with increased context windows: LLaMA-3-instruct (8B)\footnote{\href{https://huggingface.co/meta-llama/Meta-Llama-3-8B-Instruct}{meta-llama/Meta-Llama-3-8B-Instruct}} with 8K tokens and Phi-3-mini-instruct (3.8B) with 128K tokens.\footnote{\href{https://huggingface.co/microsoft/Phi-3-mini-128k-instruct}{microsoft/Phi-3-mini-128k-instruct}}
    \item \textbf{FID}. We train a FID Flan-T5-base (250M) \citep{DBLP:journals/corr/abs-2210-11416}.\footnote{\href{https://huggingface.co/google/flan-t5-base}{google/flan-t5-base}}
\end{itemize}

\begin{graybox}
This article has evolved over two distinct versions, each exploring different LLM ICL readers:
\vspace{-5pt}
\begin{enumerate}[leftmargin=0.1em, noitemsep]
    \item[] \textbf{v1 - February 2024:} Experiments and ablation studies conducted utilizing LLaMA-2 and Zephyr.
    \item[] \textbf{v2 - June 2024:} Integration of the newly released LLaMA-3 and Phi-3-mini in the core experiments, with up-to-date baselines, namely BioMistral, MeditronLlama-Slerp, LLaMA-3-Meditron, and OpenBioLLM.
\end{enumerate}
\end{graybox}

\begin{table*}[!t]
\centering
\adjustbox{width=\linewidth}{
\begin{threeparttable}[b]
\begin{tabular}{cllllllll}
\hline
& \textbf{Model} & \textbf{Ground (Source)}\tnote{*} & \textbf{Learning} & \textbf{Params} & \textbf{MedQA} & \textbf{MedMCQA} & \textbf{MMLU} & \textbf{AVG ($\downarrow$)} \\
\hline
\multirow{9}{*}{\includegraphics[width=.35cm]{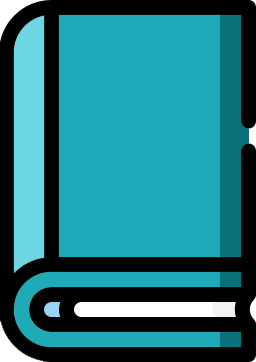}} & 
\textcolor{blue}{LLaMA-3-Instruct} & $\varnothing$ & 1-shot & 8B & 60.6 & 55.7	& 69.8 & 62.0 \\
& \textcolor{orange}{Phi-3-mini} & $\varnothing$ & 1-shot	& 3.8B & 55.1 & 53.5 & 70.3 & 59.6 \\
& MEDITRON\tnote{1} & $\varnothing$ & Fine-tuned & 7B & 52.0 & \textbf{59.2} & 55.6 & 55.6 \\
& PMC-LLaMA\tnote{1} & $\varnothing$ & Fine-tuned & 7B & 49.2 & 51.4 & 59.7 & 53.4 \\
& LLaMA-2\tnote{1} & $\varnothing$ & Fine-tuned & 7B & 49.6 & 54.4 & 56.3 & 53.4 \\
& \textcolor{violet}{Zephyr-$\beta$} & $\varnothing$ & 2-shot & 7B & 49.3 & 43.4 & 60.7 & 51.1 \\
& Mistral-Instruct\tnote{1} & $\varnothing$ & 3-shot & 7B & 41.1 & 40.2 & 55.8 & 45.7 \\
& \textcolor{teal2}{LLaMA-2-chat} & $\varnothing$ & 2-shot & 7B & 36.9 & 35.0 & 49.3 & 40.4 \\
& Codex\tnote{2} & $\varnothing$ & 0-shot & 175B & 52.5 & 50.9 & -- & -- \\
\hline
\multirow{8}{*}{\includegraphics[width=.6cm]{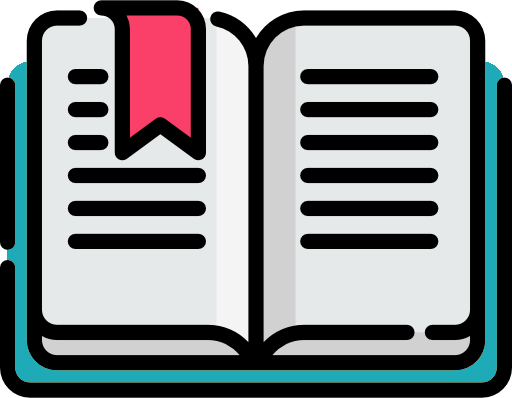}} & 
			
\textcolor{orange}{\textbf{MedGENIE-Phi-3-mini}} & G (\small{PMC-LLaMA}) & 1-shot &	3.8B & 	\textbf{64.7} \textcolor{orange}{(+9.6)} & 54.1 \textcolor{orange}{(+0.6)} & \textbf{70.8} \textcolor{orange}{(+0.5)} & \textbf{63.2} \textcolor{orange}{(+3.6)}\\
& \textcolor{blue}{\textbf{MedGENIE-LLaMA-3-Instruct}} & G (\small{PMC-LLaMA}) & 1-shot & 8B  &  \underline{63.1} \textcolor{blue}{(+2.5)} & 56.2 \textcolor{blue}{(+0.5)} & \underline{68.9} \textcolor{blue}{(-0.9)} & \underline{62.7} \textcolor{blue}{(+0.7)} \\
& \textcolor{violet}
{\textbf{MedGENIE-Zephyr-$\beta$}} & G (\small{PMC-LLaMA}) & 2-shot & 7B & 59.7 \textcolor{violet}{(+10.4)} & 51.0 \textcolor{violet}{(+7.6)} & 66.1 \textcolor{violet}{(+5.4)} & 58.9 \textcolor{violet}{(+7.8)} \\
& \textbf{MedGENIE-FID-Flan-T5} & G (\small{PMC-LLaMA}) & Fine-tuned & 250M & 53.1 & 52.1 & 59.9 & 55.0 \\
& Zephyr-$\beta$ & R (\small{MedWiki}) & 2-shot & 7B & 50.5 & 47.0 & \textbf{66.9} & 54.8 \\
& VOD\tnote{3} & R (\small{MedWiki}) & Fine-tuned & 220M & 45.8 & \underline{58.3} & 56.8 & 53.6 \\
& \textcolor{teal2}{\textbf{MedGENIE-LLaMA-2-chat}} & G (\small{PMC-LLaMA}) & 2-shot & 7B & 52.6 \textcolor{teal2}{(+15.7)} & 44.8 \textcolor{teal2}{(+9.8)} & 58.8 \textcolor{teal2}{(+9.5)} & 52.1 \textcolor{teal2}{(+11.7)} \\
& Mistral-Instruct & R (\small{MedWiki}) & 2-shot & 7B & 45.1 & 44.3 & 58.5 & 49.3 \\
& LLaMA-2-chat & R (\small{MedWiki}) & 2-shot & 7B & 37.2 & 37.2 & 52.0 & 42.1 \\
\hline
\multirow{2}{*}{\faUser} & \multicolumn{4}{l}{\textit{Human (passing score)}} & \textit{60.0} & \textit{50.0} & \textit{--} & \\
& \multicolumn{4}{l}{\textit{Human (expert score)}} & \textit{87.0} & \textit{90.0} & \textit{89.8} & \\
\hline
\end{tabular}
\begin{tablenotes}[para]
\item[1] \citep{DBLP:journals/corr/abs-2311-16079} \item[2] \citep{DBLP:journals/corr/abs-2207-08143} \item[3] \citep{DBLP:conf/icml/LievinMJW23}\\
\item[*] Whether the input is augmented with external knowledge to ground model predictions; $\varnothing$ = No Grounding, R = Retrieved, G = Generated. While external knowledge sources for RAG (i.e., Ground=R) are the same for each benchmark, the contexts generated by an LLM (i.e., Ground=G) are specific for the questions of each benchmark.\\
\textbf{Note:} The influence (i.e., $\Delta$ Acc.) of grounding LLMs with knowledge generated under the MedGENIE framework is highlighted with colors.
\end{tablenotes}
\end{threeparttable}}
\caption{Main accuracy results of \textsc{MedGENIE} (ICL with LLaMa-3-Instruct, Phi-3-mini, LLaMA-2-chat or Zephyr-$\beta$, and FID with Flan-T5) against open-source closed-book and open-book baselines. Evaluation on MedQA (test), MedMCQA (eval), and MMLU (test) benchmarks. Bold and underline denote the best and second-best scores.}
\label{table:main_results}
\end{table*}

\subsection{Evaluation}

\paragraph{Multiple Choice Prompting and Direct Completion} 

In alignment with recent literature \citep{DBLP:journals/corr/abs-2311-16452,DBLP:journals/corr/abs-2311-16079,DBLP:journals/corr/abs-2207-08143,DBLP:journals/corr/abs-2212-13138}, as anticipated in \cref{sec:method}, we utilize multiple-choice prompting (MCP).
In MCP, the model is presented with both $\mathbf{q}$ and $\mathcal{A}$, where each option is bound to a proxy symbol (A, B, C, D, E), and the model only has to predict a single token.\footnote{The prediction may be longer (e.g., "(A) Vitamin B12"), but only the mentioned symbol is considered as the answer.}
This diverges from earlier cloze prompting (CP) strategies \citep{DBLP:conf/nips/BrownMRSKDNSSAA20,J1WhitePaper,DBLP:conf/icml/DuHDTLXKZYFZFBZ22,DBLP:journals/corr/abs-2201-11990,DBLP:journals/jmlr/ChowdheryNDBMRBCSGSSTMRBTSPRDHPBAI23}, where only $\mathbf{q}$ is passed to the model and the candidate answers are scored independently.
Unlike CP, MCP (i) does not conflate the likelihood of an answer with the likelihood of its text, (ii) does not require normalization procedures, (iii) explicitly contrasts different answer options, and (iv) only requires a single forward pass; see \citep{DBLP:conf/iclr/RobinsonW23}.
Since we focus on evaluating the pure impact of artificial contexts, we operate under the strict assumption of \textit{directly} inferring the correct answer with this additional input signal. We do not resort to alternative schemas, such as CoT or self-consistency CoT, beyond the scope of our research work.
For reproducibility, we adopt greedy decoding as in \citet{DBLP:journals/corr/abs-2311-16079}.

\paragraph{Metrics} We use Accuracy (\% of correctly solved questions) as our main metric across all benchmarks. We resort to Recall@$K$ for measuring the \% of retrieved generated contexts rather than gold-chunked ones after top-$K$ retrieval and reranking.

\subsection{Baselines}

We juxtapose \textsc{MedGENIE} with two categories of models.
\textit{Closed-book}. We prioritize medium-sized LLM baselines ($\leq$8B), either pre-trained or fine-tuned on medical data, fostering an equitable and resource-aware comparison with our models.
\textit{Open-book}. We incorporate all the contributions to our knowledge, going beyond text-only grounding.

We further incorporate baselines implemented and run by us. For the \textit{closed-book} category, we conduct few-shot inferences utilizing up-to-date generalist LLM backbones. For \textit{open-book}, we explore augmenting our backbone models with MedWiki \citep{DBLP:conf/icml/LievinMJW23}, a collection of 293,593 medical articles from English Wikipedia, built by querying each option of MedMCQA and MedQA against the Wikipedia API.

\section{Results}

Our core findings are delineated in Table~\ref{table:main_results}.
For space reasons, we report key baselines with available results for each benchmark. To ensure complete and impartial evaluations, we also document current state-of-the-art (SOTA) models outside the intersection.
We refer the reader to Appendix~\cref{app:full_baselines} for full benchmark-specific baselines.

\subsection{Artificial grounding effect}

We direct our attention to the accuracy gap between \textsc{MedGENIE} models and their non-grounded version (cf. teal-colored scores in Table~\ref{table:main_results}).

\paragraph{Impact on ICL Reader}

Our framework significantly enhances the performance of all the considered reader models. Specifically, LLaMA-2-chat, Zephyr-$\beta$, and Phi-3-mini show average improvements of +11.7, +7.8, and +3.6 points, respectively. The impact on MedQA is particularly notable, with improvements of +15.7, +9.8, and +9.6 points, respectively. Artificial grounding further elevates Phi-3-mini, establishing a new SOTA in MedQA and delivering the best overall performance. The combination of \textsc{MedGENIE} with the strong reasoning capabilities and extended context window of Phi-3 fully unlocks the potential of this lightweight model.
The results for LLaMA-2-chat and Zephyr-$\beta$ indicate that the advantages of \textsc{MedGENIE} are particularly prominent in models without prior domain-specific knowledge. In contrast, newer models, such as LLaMA-3, already demonstrate a robust medical background, likely due to their extensive pre-training phase over 15 trillion tokens. As a result, the impact of \textsc{MedGENIE} is less significant.
Few-shot models struggle against fine-tuned versions on MedMCQA, highlighting the necessity of dedicated training for achieving high scores. Surprisingly, the ICL \textsc{MedGENIE} models excel in both MedQA and MMLU, outperforming fully fine-tuned LLMs. Notably, MedGENIE-LLaMA-2-chat surpasses the fine-tuned LLaMA-2-chat by 3 points in MedQA and 2.5 points in MedMCQA. Similarly, MedGENIE-Zephyr-$\beta$ outperforms the strongest fine-tuned alternative, MEDITRON, by +7.7 points in MedQA and +10.5 points in MMLU.
Our approach eliminates the need for expensive full fine-tuning of 7B models, which requires 56GB VRAM when using a standard AdamW optimizer. Instead, ICL \textsc{MedGENIE} operates efficiently on a standard consumer GPU with 14GB VRAM in half precision.

\paragraph{Impact on FID Reader} 

MedGenie-FID-Flan-T5 achieves 53.1\% accuracy on MedQA, with outstanding efficiency.
It outshines 7B baselines fine-tuned on the train set of target benchmarks, such as MEDITRON (+1.1), PMC-LLaMA (+3.9), LLaMA-2-chat (+3.5), and even CODEX 175B in zero-shot settings (+0.6)---leveraging up to 706$\times$ fewer parameters.
In MedMCQA, it significantly exceeds ICL models, such as Zephyr-$\beta$ (+8.7), Mistral-Instruct (+11.9), and LLaMA-2-chat (+17.1). It remains superior to fully fine-tuned PMC-LLaMA (+0.7) and the zero-shot CODEX 175B model (+1.2), but struggles to compete with previous SOTA holders: MEDITRON (-7.1) and VOD (-6.2). We motivate this gap with the huge difference in size and pre-knowledge, other than the more notional nature of the MedMCQA questions, where MedWiki grounding is highly effective.
In MMLU, it again shows exceptional performance, outperforming all fine-tuned baselines on the MedMCQA train set. This highlights the generalization power of \textsc{MedGENIE} compared to larger and more costly LLMs.
Zooming out, our findings corroborate the \textit{tiny titan} nature of Flan-T5 observed by other researchers \citep{DBLP:journals/corr/abs-2402-00841}.

\subsection{Generated vs retrieved context}
\label{subsec:gen_vs_ret}

\textsc{MedGENIE} tops all prior retrieve-then-read solutions across all benchmarks except for MedMCQA, where it is only rivalled by VOD. Generated contexts yield higher accuracy compared to Wikipedia and textbook chunks, as well as UMLS subgraphs. Impressively, artificial augmentation applied to LLaMA-2-chat results in +10 points (avg) compared to MedWiki augmentation.

\paragraph{Reranking Preference}

A crucial aspect of our framework is to ensure top-tier contexts. However, the use of a 13B model as the context generator raises concerns about its ability to deliver adequate quality contexts, especially compared to stronger closed and open-source alternatives such as GPT-4 and LLaMA-70B. These concerns arise from the risk of hallucinations or less accurate text compared to factual information stored in knowledge bases.
Human evaluation is worth exploring, but it would require considerable resources and expertise, which may not be readily accessible in our case.
Hence, we propose using a SOTA reranker, BGE-large \citep{DBLP:journals/corr/abs-2309-07597},\footnote{\href{https://huggingface.co/BAAI/bge-reranker-large}{BAAI/bge-reranker-large}} to determine whether our artificial contexts are deemed more relevant than the top contexts (chunks) fetched from MedWiki.
Given a question $\mathbf{q}$, we take the 5 multi-view contexts $\mathcal{C}$ generated by PMC-LLaMA. Next, we retrieve the 10 MedWiki contexts most similar to $\mathbf{q}$ (PubMedBERT \citep{DBLP:journals/health/GuTCLULNGP22}, cosine similarity). Subsequently, we create 15 $\mathbf{q}$--context pairs to feed the reranker, registering their relevance score.
Finally, we apply Recall@$K$ to quantify the \% of generated (relevant) contexts in the top-$K$ positions according to the reranker. Figure~\ref{fig:reranker} portrays the results.
The reranker strongly favors our artificial contexts over the retrieved ones across all three benchmarks.
Precisely, we achieve a Recall@1 of 91\%, 98\%, and 96\% on MedQA, MedMCQA, and MMLU, respectively.
Option-focused contexts contribute predominantly to this preference. However, as $K$ rises, our recall begins to diminish, and the significance of option-free contexts amplifies, thus advocating for our multi-view approach.

\begin{figure}[!t]
    \centering
    
    \begin{subfigure}[t]{\linewidth}
        \centering
        \begin{tikzpicture}
            \fontsize{8}{8}\selectfont
            \node[inner sep=2pt, align=center] {\begin{tabular}{ll}
                \ref{plot:medqa_all} MedQA (all) &
                \ref{plot:medqa_option_free} MedQA (option-free) \\
                \ref{plot:medmcqa_all} MedMCQA (all) &
                \ref{plot:medmcqa_option_free} MedMCQA (option-free) \\
                \ref{plot:mmlu_all} MMLU (all) &
                \ref{plot:mmlu_option_free} MMLU (option-free)
            \end{tabular}};
        \end{tikzpicture}
    \end{subfigure}\\[1mm]
    
    \begin{subfigure}[t]{\linewidth}
        \hspace{-2mm}
        \centering
        \begin{tikzpicture}
            \begin{axis}[
                every minor tick={very thin, gray},
                width=\linewidth, height=4.5cm,
                axis line style={gray},
                grid=both,
                grid style=dashed,
                xtick={1, 3, 5, 8},
                xlabel = {$K$},
                xlabel style={yshift=0.2cm},
                ylabel = {Recall@$K$},
                ytick={0, 20, ..., 100},
                enlarge y limits=0.15,
                axis background/.style={fill=plotbackground!50},
                every tick label/.append style={font=\fontsize{9}{9}\selectfont},
                every x tick/.style={/pgfplots/major tick length=0pt},
                ylabel near ticks,
                xlabel near ticks,
                label style={font=\small},
                ymajorgrids=true,
                ]
            \addplot[color=teal1, line width=1.5pt] coordinates {
                (1, 90.65)
                (3, 82.67)
                (5, 71.92)
                (8, 53.74)
            }; \label{plot:medqa_all}
            \addplot[color=teal1, dashed, line width=1.5pt] coordinates {
                (1, 28.51)
                (3, 28.25)
                (5, 26.69)
                (8, 20.38)
            }; \label{plot:medqa_option_free}
        
            \addplot[color=teal2, line width=1.5pt] coordinates {
                (1, 98.40)
                (3, 94.82)
                (5, 85.17)
                (8, 58.92)
            }; \label{plot:medmcqa_all}
            \addplot[color=teal2, dashed, line width=1.5pt] coordinates {
                (1, 16.21)
                (3, 22.80)
                (5, 29.63)
                (8, 22.34)
            }; \label{plot:medmcqa_option_free}
            
            \addplot[color=teal3, line width=1.5pt] coordinates {
                (1, 96.40)
                (3, 91.98)
                (5, 81.44)
                (8, 57.50)
            }; \label{plot:mmlu_all}
            \addplot[color=teal3, dashed, line width=1.5pt] coordinates {
                (1, 23.25)
                (3, 26.23)
                (5, 29.00)
                (8, 21.76)
            }; \label{plot:mmlu_option_free}
            \end{axis}
        \end{tikzpicture}
    \end{subfigure}
    \caption{Percentage of multi-view generated contexts compared to MedWiki-retrieved contexts in the top-$K$ positions of a BGE-large reranker.}
    \label{fig:reranker}
\end{figure}

\begin{table}[!t]
\centering
\adjustbox{width=\linewidth}{
\begin{threeparttable}[b]
\begin{tabular}{llllll}
\toprule
\textbf{KB}\tnote{*} & \textbf{G}\tnote{*} & \textbf{llama2}\tnote{1} & \textbf{mistral}\tnote{2} & \textbf{zephyr}\tnote{3} \\ 
\midrule      
\rowcolor{lightgrey}\multicolumn{6}{c}{MedQA} \\
\midrule
4.5M & -- & 37.2 & 45.1 & 50.4\\
4.5M & 96K\tnote{$\dagger$} & 41.4 \textcolor{teal1}{(+4.2)} & 45.6 \textcolor{teal1}{(+0.5)}& 50.8 \textcolor{teal1}{(+0.4)}\\
4.5M & 2M\tnote{$\ddagger$} & 40.8 \textcolor{teal2}{(+3.6)} & 45.9 \textcolor{teal2}{(+0.8)} & 51.2 \textcolor{teal2}{(+0.8)}\\
\midrule
\rowcolor{lightgrey}\multicolumn{6}{c}{MedMCQA} \\
\midrule
4.5M & -- & 37.3 & 44.3 & 47.1\\
4.5M & 96K\tnote{$\dagger$} & 41.8 \textcolor{teal1}{(+4.5)} & 48.1 \textcolor{teal1}{(+3.8)}& 50.3 \textcolor{teal1}{(+3.2)}\\
4.5M & 2M\tnote{$\ddagger$} & 43.7 \textcolor{teal2}{(+6.4)} & 49.8 \textcolor{teal2}{(+5.5)} & 50.9 \textcolor{teal2}{(+3.8)}\\
\midrule
\rowcolor{lightgrey}\multicolumn{6}{c}{MMLU (out-of-domain)}\\
\midrule
4.5M & -- & 52.0 & 58.5 & 66.9\\
4.5M & 96K\tnote{$\dagger$} & 53.5 \textcolor{teal1}{(+1.5)} & 58.3 & 67.4 \textcolor{teal1}{(+0.5)}\\
4.5M & 2M\tnote{$\ddagger$} & 53.0 \textcolor{teal2}{(+1.0)} & 58.4 & 67.0 \textcolor{teal2}{(+0.1)}\\
\bottomrule
\end{tabular}
\begin{tablenotes}[para]
\item[1] LLaMA-2-chat (7B) \item[2] Mistral-Instruct (7B) \item[3] Zephyr-$\beta$ (7B) \item[$\dagger$] MedQA$+$MedMCQA train set \item[$\ddagger$] MedQA$+$MedMCQA train+test set \\
\item[*] Number of chunks of MedWiki (KB) and the generated sources (G). VectorDB details in Appendix~\cref{app:context_rag}.\\
\textbf{Note:} The positive influence (i.e., $\Delta$ Acc.) of retrieving from artificial contexts is highlighted with colors.
\end{tablenotes}
\end{threeparttable}}
\caption{Accuracy on MedQA (test), MedMCQA (eval), and MMLU (test) with a retrieve-then-read pipeline based on ICL and MedWiki as external knowledge base, progressively enriched with artificial contexts.}
\label{table:rag}
\end{table}

\paragraph{Artificial Contexts for Retrieval Augmented Generation}

Our analysis delves deeper into assessing the efficacy of synthetic contexts by implementing them within a standard retrieve-then-read pipeline.
The objective is to discern whether their combination with factual data can improve overall performance.
We analyze two different settings.
First, we supplement MedWiki with artificial contexts from the MedQA test set and the MedMCQA eval set. We assess whether a small portion of synthetic data may impact a much larger body of factual data.
Second, we broaden the knowledge base to cover contexts generated from the MedQA and MedMCQA train set questions. Although such contexts may not pertain to the benchmark test sets, they might offer additional insights to improve accuracy.
For each setting, we evaluate retrieval-augmented ICL readers on all three benchmarks.
We refrain from considering MMLU contexts to ensure a complete \textit{out-of-domain} validation, enabling us to understand the validity of artificial passages when obtained from a different domain.
The results are pointed out in Table~\ref{table:rag}.
Mixed generation$+$retrieval knowledge sources consistently improve accuracy on all benchmarks, except Mistral-Instruct on MMLU. Combined with Zephyr-$\beta$, we obtain 67.4, pushing the SOTA on MMLU. Despite these positive gains, scores fall short in competing with the generate-then-read paradigm of \textsc{MedGENIE}. We attribute this discrepancy to the retriever influence and the inevitable degradation deriving from segmentation.

\paragraph{RAGAS Evaluation}

When tasked with generating extensive medical content, the context generator may produce significant hallucinations. Demonstrating a preference for generated contexts over those retrieved from MedWiki might not be sufficient, as the preference could be attributed to coherence rather than factual accuracy. To validate our findings further, we evaluate both generated and retrieved contexts using the RAGAS library \citep{es-etal-2024-ragas}. RAGAS offers various reference-free metrics for RAG pipelines, following the LLM-as-a-judge paradigm. Particularly we focus on:
\begin{itemize}
    \item \textbf{Context Recall}:
    Measure the extent to which the retrieved/generated context aligns with the ground-truth (GT) answer. Each sentence in the GT answer (generally one) is analyzed to determine whether it can be attributed to the retrieved context or not.
    \begin{equation}
        \footnotesize
        CR = \frac{|\text{GT sentences linked to the context}|} {|\text{GT sentences}|}
    \end{equation}

    \item \textbf{Context Precision}:
    For each chunk in retrieved/generated context, check if it is relevant or not relevant to arrive at the GT answer for the given question. Ideally, all relevant chunks should appear in the highest ranks.
    \begin{equation}
        \footnotesize
        CP@K = \frac{\sum_{k=1}^{K}{P@k \cdot v_k}}{|\text{Relevant chunks in the top-\(K\)}|}
    \end{equation}
    \begin{equation}
        \footnotesize
        P@k = \frac{\text{true positives}@k}{(\text{true positives}@k + \text{false positives}@k)}
    \end{equation}
    where $K$ is the total number of chunks in the context and $v_k \in \{0,1\}$ is the relevance indicator at rank $k$.

    \item \textbf{Faithfulness}:
    The factual consistency of the generated answer is evaluated by comparing its claims against the provided context to determine if they can be accurately inferred.
    \begin{equation}
        \footnotesize
        F = \frac{|\text{Generated answer claims implied by the context}|}{|\text{Generated answer claims}|}
    \end{equation}
\end{itemize}

Each metric gives score in a $[0, 1]$ boundary; the higher, the better.
We utilize \texttt{gpt-4-turbo-2024-04-09}. Due to the high API costs, we limit our evaluation to a sample of MedQA, the most relevant testbed in our study. Table \ref{tab:ragas} reports the performance of Zephyr-$\beta$ on 150 randomly selected instances, where both the generated and retrieved contexts direct the LLM to produce correct answers. We also apply RAGAS to 50 random instances where both sources lead to wrong answers. The results reaffirm the superiority of artificial contexts, achieving up to +27.2 $CR$, +39.3 $CP$, +35.9 $F$.

\begin{table}[htb!]

\adjustbox{width=\linewidth}{
\centering
\begin{tabular}{cccccc}
\hline
& \textbf{Metric}  & \textbf{\# Samples} & \textbf{Answer} & \textbf{G} & \textbf{R} \\
\hline
\multirow{5}{*}{\includegraphics[width=.45cm]{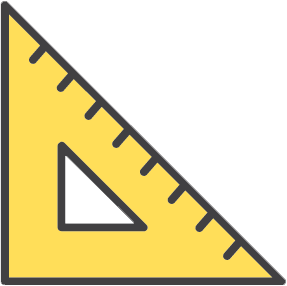}} & $CR$ & 150 & \textit{correct} & \href{https://smith.langchain.com/public/08a26eea-7c7c-4f69-9698-daf62635ba02/r}{93.4} & \href{https://smith.langchain.com/public/16807dd2-cf7d-473b-86f8-cf6d825e17d7/r}{76.2} \\
& $CP$  & 150 & \textit{correct} & \href{https://smith.langchain.com/public/4d4249bf-f432-4dd5-afc3-1a5fdb42aa7b/r}{87.9} & \href{https://smith.langchain.com/public/7ff3d138-adc6-4cd8-8a49-5fe2175a6b0a/r}{48.6} \\
& $F$  & 150 & \textit{correct} & \href{https://smith.langchain.com/public/08a26eea-7c7c-4f69-9698-daf62635ba02/r}{59.7} & \href{https://smith.langchain.com/public/16807dd2-cf7d-473b-86f8-cf6d825e17d7/r}{23.8} \\
\cline{2-6}
& $CR$  & 50 & \textit{wrong} & \href{https://smith.langchain.com/public/8f117018-a91c-4fe0-b67e-a1960bf4d204/r}{59.2} & \href{https://smith.langchain.com/public/4fb7929a-e6a0-4e7a-ae87-42eb197d35c8/r}{32.0} \\
& $CP$  & 50 & \textit{wrong} & \href{https://smith.langchain.com/public/0ee14f13-02df-44ca-928b-50b27eb0eea9/r}{55.3} & \href{https://smith.langchain.com/public/4c7fc330-3974-4e05-b90c-f8edf23663de/r}{29.5} \\
\hline
\end{tabular}}
\caption{RAGAS evaluation on MedQA. For transparency, scores are linked to the LangSmith pages providing detailed run information, including selected instances, prompts, predictions, and inference time.}
\label{tab:ragas}
\end{table}

\section{Conclusion}

This paper introduces \textsc{MedGENIE}, a novel fully-generative framework for medical ODQA, with a particular emphasis on resource-constrained environments. Through comprehensive experimentation with three standard medical benchmarks, \textsc{MedGENIE} demonstrates substantial performance improvements over the existing closed-book and open-book methodologies. By injecting multi-view contexts from a medical LLM via ICL or lightweight fine-tuning, \textsc{MedGENIE} achieves new SOTA results, even surpassing expensive fine-tuned LLMs or the largest zero-shot models with up to 706$\times$ fewer parameters. Furthermore, our research highlights the effectiveness of artificial passages in enhancing \textit{retrieve-then-read} workflows, showcasing the potential of generated contexts to surpass or assist retrieved counterparts. From a wider angle, \textsc{MedGENIE} represents a promising approach to address the intricate challenges of medical ODQA, laying the foundation for future advancements in the field. 

\section*{Ethical Statement}

We honor and support the ACL Code of Ethics. LLMs offer significant social benefits, but also pose potential risks. Safety and trustability are paramount in the medical domain. This paper uses knowledge embedded within LLM parameters to tackle open-domain questions. Unlike retrieve-then-read methods reliant on curated external corpora, using models to generate contextual documents may inadvertently amplify inherent biases and deviate from clinical and societal norms, potentially leading to the dissemination of convincing medical misinformation. Therefore, we advocate for a cautious approach, recommending manual scrutiny of any output by domain experts before real-world utilization. This ethical precaution is vital to avoid disseminating potentially erroneous or misleading information, especially in the clinical and scientific communities.

\section*{Limitations and Future Work}

Despite achieving SOTA performance on the benchmarks presented, our MedGENIE pipeline has several limitations that warrant attention. By replacing the retrieval component with a generative one, we lose the ability to update the knowledge state, which remains frozen to that of the LLM training time. This stands in contrast to the standard retrieve-then-read approach, which allows for the incorporation of new documents as new information becomes available. In a rapidly evolving field like medicine, this ability to quickly adapt by adding temporally recent documents or documents from new domains is important to cope with scientific information overload \citep{landhuis-2016-info-overload}. In our method, the responsibility of retaining all knowledge rests entirely on the LLM, and incorporating new knowledge would probably require retraining the context generator.

Our method generates context for any given question, even when the medical LLM lacks knowledge. This may produce a noisy context with inaccurate or irrelevant information. We retain all retrieved contexts without relevance filtering, a strategy that could notably enhance overall performance if implemented.
Moreover, the efficacy of our ICL pipeline hinges on the reader's ability to process long prompts within the context window, which may not always be the case for every LLM. This limitation can impact results, particularly when dealing with complex or verbose contexts.

Recent literature suggests solutions for these limitations. \citet{raft} introduced Retrieval Augmented Fine Tuning (RAFT), a training approach that enhances the model’s ability to answer questions in an \textit{open-book} in-domain setting. RAFT trains the model to disregard irrelevant documents when given a question and a set of retrieved documents. \citet{when-to-retrieve} propose ADAPT-LLM, teaching an LLM to generate a \texttt{<RET>} token when it does not know the answer, triggering retrieval only when necessary. Interestingly, our work suggests potential improvements for both approaches. For RAFT, it might be interesting to explore whether fine-tuning can benefit from augmentation with artificially generated contexts. For the ADAPT-LLM approach, introducing a second token, \texttt{<GEN>}, could be valuable when the retrieved context does not contain the answer, prompting the generation of a context from an auxiliary LLM.

Previous work, such as GenRead \citep{DBLP:conf/iclr/0002IWXJ000023}, has explored the zero-shot pipeline using the same model for both generator and reader, following a self-distillation approach. This suggests that a model may indeed benefit from directly extracting knowledge embedded within its own parameters. We acknowledge the significance of assessing the performance of our generator, PMC-LLama-13B (awq), in the role of a reader. Nonetheless, we chose not to pursue this evaluation due to the model's restricted context window of merely 2,000 tokens. Expanding the input prompt with artificial contexts would rapidly exhaust this limited context capacity. However, with recent advancements in LLMs that possess a stronger medical background and an extended context window of 8K tokens (e.g., LLaMA 3 \footnote{\href{https://www.meditron.io/}{https://www.meditron.io/}}, OpenBioLLM \footnote{\href{https://huggingface.co/aaditya/Llama3-OpenBioLLM-8B}{aaditya/Llama3-OpenBioLLM-8B}}), this approach warrants further exploration as future work.

Finally, evaluating the effectiveness of multiple-round reasoning coupled with context generation, as opposed to other methodologies such as multiple-hop reasoning with retrieval (e.g., DSP \citet{DSP}), could provide valuable insights into optimizing QA systems for the complexities of medical domain tasks.


\section*{Acknowledgements}

This research is partially supported by (i) the Complementary National Plan PNC-I.1 "Research initiatives for innovative technologies and pathways in the health and welfare sector" D.D. 931 of 06/06/2022, DARE---DigitAl lifelong pRevEntion initiative, code PNC0000002, CUP B53C22006450001, (ii) the PNRR---M4C2---Investment 1.3, Extended Partnership PE00000013, FAIR---Future Artificial Intelligence Research, Spoke 8 "Pervasive AI," funded by the European Commission under the NextGeneration EU program, (iii) the European Commission and the Italian MIMIT through the Chips JU TRISTAN project (G.A. 101095947). We thank Andrea Zecca, Antonio Lopez, Matteo Vannucchi, and Stefano Colamonaco for their valuable support in ablation studies with retrieve-then-read pipelines.

\bibliography{references}

\newpage

\appendix

\section{Implementation Details}
\label{app:implementation_details}

\paragraph{Hardware Configuration} We ran each experiment on an in-house workstation having one Nvidia GeForce RTX3090 GPU with 24GB of dedicated memory, 64GB of RAM, and an Intel® Core™ i9-10900X1080 CPU @ 3.70GHz.

\paragraph{Checkpoints} All the models trained are open source and have permissive licenses for this study: Apache-2.0 (PMC-LLaMA, Mistral-Instruct, Flan-T5), MIT (Zephyr-$\beta$, Phi-3-mini), Meta License (LLaMA-2, LLaMA-3). We initialized them with the corresponding pre-trained versions available in the HuggingFace Transformers library.\footnote{\url{https://github.com/huggingface/transformers}}

\paragraph{Medical-Expert Generator} To accommodate memory limitations, we applied post-training Activation-aware Weight Quantization (AWQ) \citep{DBLP:journals/corr/abs-2306-00978} in 4 bits to PMC-LLaMA 13B.

\paragraph{Inference} For context generation and ICL experiments, we utilized the vLLM library.\footnote{\url{https://github.com/vllm-project/vllm}}
Since vLLM does not currently support inference for sequence-to-sequence models, we employed the Transformer library for the FID Flan-t5-base.
Specifically, generating $\mathcal{C}$, we set a random decoding strategy with a temperature of 0.9, a frequency penalty of 1.95, and a maximum of 512 new tokens.
We returned 3 output contexts for a given option-focused prompt and 2 output contexts for a given option-free prompt.
We increased the inference batch size to 5 and generated all contexts in parallel with one inference call to the LLM.
Thus, the overall latency remains the same as when using a single context.
To ensure reproducibility, during the evaluation of both ICL and supervised readers, we adopted a greedy decoding strategy with the random state set to 0.

\paragraph{FID Training} In our implementation, we successfully scaled up to 5 independently encoded question--context pairs on 24GB VRAM. For each pair, we used a maximum context window of 1024 tokens for MedQA and 600 tokens for MedMCQA. During training, for MedQA, we chose a batch size of 1 with 4 accumulation steps, totaling 40,712 training steps, and executed evaluations every 10,178 steps. Similarly, for MedMCQA, we employed a GPU batch size of 2 with 2 accumulation steps, trained over 182,816 steps, and evaluated every 22,852 steps. For both benchmarks, we employed a linear scheduler with a warmup phase comprising 10\% of the total training steps. The learning rate was set to 5e-5, using an AdamW optimizer with a weight decay of 0.01.

\section{Context Generation Insights}

\subsection{Human Evaluation of Context Generator}
\label{app:medical_llm_choice}

We compare three LLMs as candidate generators, two medical (PMC-LLaMA-13B, BioMedGPT-7B) and one generalist (LLaMA-2-Chat-13B), by enlisting the expertise of five external medical professionals with NLP competencies.
We exclude MEDITRON due to the release of pre-trained checkpoints only.
Motivated by \cite{fabbri-etal-2019-multi,huang-etal-2023-swing,narayan-etal-2018-dont}, we use a direct comparison strategy, proven to be more reliable, sensitive, and less labor-intensive than rating scales.
We consider a pool of 100 questions sampled in a balanced manner from MedQA, MedMCQA, and MMLU.
For each, experts are presented with contexts inferred by 2 out of 3 LLMs.
We generate one context for question, allocating 60\% of the evaluation to option-focused and 40\% to option-free. In order to have also insights about the quality of the contexts generated, 
we ask experts to select the best all-around context considering relevance, factuality and succinctness, optionally providing a text comment; a ``tie'' is declared if an expert perceives the two contexts as of equal quality. This approach inherently assesses context quality, with high-quality contexts being those that fully satisfy our evaluation criteria.
Finally, the total number of mandatory preference labels per participant is 300.
We randomize the order of pairs and context sources per question to guard the rating against being gamed.
The final score of each LLM is the percentage of times that its contexts are selected as the best, minus the percentage of times they are not.
The results indicate PMC-LLaMA as the best choice, garnering 41\% preference, surpassing LLaMA-2-Chat-13B with 36\% and BioMedGPT-7B with 33\%. The average Kendall's $\tau$ coefficients among all inter-annotator agreements is 0.72.
Table~\ref{table:medical_expert_eval} provides an annotation example.

\begin{table*}[!htb]
\small
\centering
\begin{adjustbox}{width=\linewidth}
\begin{tabularx}{\textwidth}{p{2.4cm}|X}
\hline
\multicolumn{2}{c}{\textbf{MedMCQA}[84d89a4f-db04-402d-88ab-ca39f6eee8bd]}\\
\hline
\textbf{Question} & Respiratory rhythm generation center is located at: A. Dorsal respiratory group \textbf{B. Pre-Botzinger complex} C. Ventral respiratory neurons D. Pneumotaxic center \\
\hline
\textit{PMC-LLaMA \textbf{(C1)}}\newline \textcolor{teal}{Rank: \#1} & Respiratory rhythm generation occurs at the Pre-Botzinger complex, located on either side of the medulla between the nucleus ambiguus and the lateral reticular nucleus. This complex is primarily involved in inspiration/inhalation. It fires bursts of action potentials at a frequency that oscillates with inspiration, controlling respiratory rate. The dorsal respiratory group (DRG), ventral respiratory neurons (VR), and pneumotaxic center also play a role in regulating breathing but not specifically associated with rhythm generation as Pre-Botzinger complex is responsible for generating inspiratory drive to spinal motor circuits controlling thoracic movements associated with breathing.\\
\hline
\textit{BioMedGPT-LM-7B \textbf{(C2)}}\newline \textcolor{teal}{Rank: \#3} & The pre-Botzinger complex is located in the caudal medulla and plays a critical role in the generation of respiratory rhythm. It receives input from the pontine respiratory groups and integrates them into a steady discharge pattern. This discharge pattern is used to control the duration and frequency of the respiratory cycle. The pre-Botzinger complex is a group of cells that reside in the caudal medulla and are responsible for generating the respiratory rhythm. This complex is also known as the \"pre-BotC\" or \"pre-Botzinger,\" named after the German anatomist and physiologist, Botzinger. The pre-BotC receives input from the pneumotaxic center and integrates it with the input from the pontine respiratory groups. The pre-Botzinger complex is a key component of the respiratory rhythm generation center, located in the caudal medulla. The question in the quiz is asking for the location of the pre-Botzinger complex, which is a group of neurons in the caudal medulla that play a critical role in respiration. \\
\hline
\textit{LLaMA-2-chat-13B \textbf{(C3)}}\newline \textcolor{teal}{Rank: \#2} & The Pre-Botzinger complex (PBC) is a group of neurons located in the medulla oblongata that play a crucial role in generating the respiratory rhythm. The PBC receives inputs from various sources, including sensory information from the body and inputs from the central nervous system, and integrates this information to generate the rhythmic activity that drives breathing. The dorsal respiratory group (DRG) is a group of neurons located in the dorsal part of the medulla oblongata that are involved in regulating breathing, but they are not the primary source of respiratory rhythm generation. The ventral respiratory neurons are also involved in breathing regulation, but they are not the main source of rhythm generation.The pneumotaxic center is a region in the medulla oblongata that helps regulate breathing by integrating sensory information and inputs from the central nervous system, but it is not the primary source of respiratory rhythm generation. \\
\hline
\textbf{Expert judge} & \textcolor{teal}{C1 stands out as the most relevant and concise response, directly stating the location of the Pre-Botzinger complex and contrasting its role with other potential options. This response effectively addresses the question with clarity.
C3 offers a comprehensive explanation of the Pre-Botzinger complex's role in respiratory rhythm generation, though it is slightly longer and lacks the direct comparison found in C1. 
C2, while informative, is less effective in directly addressing the question due to its narrower focus on the Pre-Botzinger complex without explicitly comparing its function with other potential locations. In summary, C1 provides the most direct and succinct response, followed by C3 for its thoroughness, and then C2 for its informative content but lack of direct comparison.}
\\
\hline
\end{tabularx}
\end{adjustbox}
\caption{Example of medical expert evaluation on the contexts generated by three LLMs.}
\label{table:medical_expert_eval}
\end{table*}

\subsection{Artificial Context Distributional Statistics}
\label{app:context_distributional_stats}

\begin{figure}[!htb]
    \centering
    \begin{subfigure}[t]{\linewidth}
        \centering
        \includegraphics[width=\linewidth]{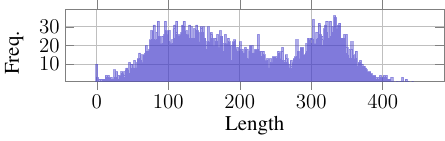}
        \caption{MedQA (4 options), test}
    \end{subfigure}\\[1mm]
    \begin{subfigure}[t]{\linewidth}
        \centering
        \includegraphics[width=\linewidth]{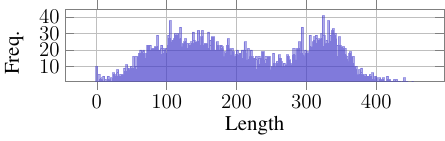}
        \caption{MedQA (5 options), test}
    \end{subfigure}\\[1mm]
    \begin{subfigure}[t]{\linewidth}
        \centering
        \includegraphics[width=\linewidth]{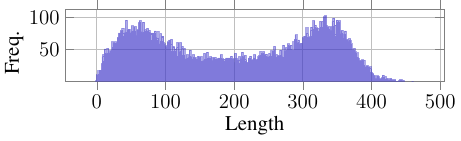}
        \caption{MedMCQA, eval}
    \end{subfigure}\\[1mm]
    \begin{subfigure}[t]{\linewidth}
        \centering
        \includegraphics[width=\linewidth]{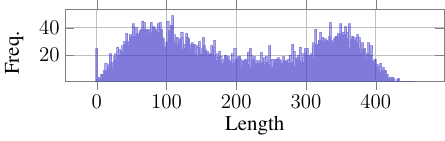}
        \caption{MMLU, test}
    \end{subfigure}
    \caption{Word-level length distribution of PMC-LLaMA artificial contexts.}
    \label{fig:context_distribution}
\end{figure}

Figure~\ref{fig:context_distribution} plots the length distribution of the artificial contexts used in this study, focusing on the test sets of our benchmarks (eval set for MedMCQA).
We tokenize contexts into words using NLTK \citep{bird-loper-2004-nltk}.
In summary, the average number of words for MedQA-4opt, MedQA-5opt, MedMCQA, and MMLU is 207, 207, 203, and 202, respectively. The maximum number of words is 442, 452, 460, and 453.

\subsection{Number and Type of Generated Contexts}
\label{app:contexts_number_type}

Table~\ref{table:impact-option-free} displays the tangible impact of incorporating option-free contexts alongside option-focused contexts.
Consistently, this integration enhances the performance of every model considered across all benchmarks, with the total number of contexts playing an essential role.
We observe that given more contexts, our framework generally achieves greater performance.
Particularly noteworthy is the experiment conducted on MedQA with the ICL \textsc{MedGENIE} models.
Here, we demonstrate that employing 3 option-focused + 2 option-free contexts yields superior accuracy compared to using only 5 option-focused contexts.
Specifically, there's an accuracy enhancement of +1.4 and +0.6 for MedGENIE-LLaMA-2-chat and MedGENIE-Zephyr-$\beta$, respectively.

\begin{table*}[!htb]
\centering
\adjustbox{max width=.8\textwidth}{
\begin{threeparttable}[b]
\begin{tabular}{ccccccccc}
    \hline
    \textbf{k} ($\uparrow$) & \textbf{option-focused} & \textbf{option-free} & \textbf{MedQA (5opt)} & \textbf{MedQA} & \textbf{MedMCQA} & \textbf{MMLU} \\
    \hline
    \rowcolor{lightgrey}\multicolumn{8}{c}{\textsc{MedGENIE-FID-Flan-T5}} \\
    \hline
     1 &  \CheckmarkBold & \XSolidBrush & 40.69  & 45.8 &  46.2 & 51.5\\
     2 & \CheckmarkBold & \XSolidBrush & 42.58  & 48.0 &  48.5 & 56.0\\
     3 & \CheckmarkBold &  \XSolidBrush  & 44.07  & 50.8 &  50.6 & 58.5\\
     4 &\CheckmarkBold &  \CheckmarkBold & 44.62  & 52.1 &  51.6 & 60.0\\
     5 & \CheckmarkBold &  \CheckmarkBold  & 45.56  & 53.1 &  52.1 & 59.9 \\
    \hline
    \rowcolor{lightgrey}\multicolumn{9}{c}{\textsc{MedGENIE-LLaMA-2-chat}} \\
    \hline
     1  & \CheckmarkBold & \XSolidBrush & 42.2  & 48.5  & 41.9 & 54.6 \\
    2  & \CheckmarkBold & \XSolidBrush & 43.8 & 50.0  &  43.3 & 55.4\\
    3  & \CheckmarkBold & \XSolidBrush & 45.7  & 51.1  & 44.1 & 55.6\\
    4  & \CheckmarkBold & \CheckmarkBold & 44.7  & 51.1  & 44.3 & 58.0 \\
    5  & \CheckmarkBold & \XSolidBrush & -  & 51.2 & - & - \\
    5  & \CheckmarkBold & \CheckmarkBold & 46.0  & 52.6 & 44.8 & 58.8 \\
    \hline
    \rowcolor{lightgrey}\multicolumn{9}{c}{\textsc{MedGENIE-Zephyr-$\beta$}} \\
    \hline
    1  & \CheckmarkBold & \XSolidBrush & 50.2  & 56.7  & 48.7 & 62.9 \\
    2  & \CheckmarkBold & \XSolidBrush & 52.2 & 58.0  & 49.9  & 61.3\\
    3  & \CheckmarkBold & \XSolidBrush & 52.5  & 58.8  & 50.8 & 63.7\\
    4  & \CheckmarkBold & \CheckmarkBold & 54.6 & 57.5  & 50.8 & 65.3 \\
    5  & \CheckmarkBold & \XSolidBrush & -  & 59.2 & - & - \\
    5  & \CheckmarkBold & \CheckmarkBold & 54.9  & 59.7  & 51.0  & 66.1 \\
    \hline
\end{tabular}
\end{threeparttable}}
\caption{Performance variation across \textit{k} different numbers of contexts  during evaluation. For \textsc{MedGENIE-FID-Flan-T5}, contexts are intended to be combined with the input question and encoded independently of each other. Conversely, for \textsc{MedGENIE-LLaMA 2-chat} and \textsc{MedGENIE-Zephyr-$\beta$}, they are passed within the input prompt as a single concatenated context before appending the test question. The first three contexts are always to be considered only \textit{option-focused}, while the fourth and fifth ones may be \textit{option-free}. Therefore, when $k=4$, only the fourth context is option-free. When $k=5$ and \textit{option-free} is flagged, it implies that both the fourth and fifth contexts are \textit{option-free}.}
\label{table:impact-option-free}
\end{table*}

\section{In-Context Learning Sample Selection}
\label{app:icl_example_selection}

Selecting appropriate samples for the prompts is essential to generate high-quality contexts.
Recent studies suggest strategies that aim to uncover more effective contextual demonstrations by retrieving examples that closely match the semantics of each input test \citep{DBLP:conf/acl-deelio/LiuSZDCC22}.
However, we are constrained by the need to first consider each demonstration as a context--question pair, rather than solely focusing on the question.
Secondly, we need to ensure that the context is not too long to avoid (i) saturation of the context window when appending the new test input, which is the result of concatenating five other contexts; (ii) the "lost in the middle" phenomenon; (iii) excessive increase in inference costs.
As a result, we employ a strategy similar to that pursued by \citet{DBLP:journals/corr/abs-2311-16079} for MEDITRON, adapted based on the characteristics of the reader model considered.
For models with lower context window capability, such as Zephyr-$\beta$ and LLaMA-2-chat, we follow these steps:
\begin{itemize}[noitemsep]
  \item We sample three questions from the training set and generate a context with PMC-LLaMA for each of them.
  \item We ensure accurate filtering of only relevant content from the generated text.
  \item Additionally, we involve a medical expert to craft pairs of questions with relevant contexts for each benchmark considered.
  \item Consequently, we end up with four different pairs of shots for each model in each benchmark.
 \item We consider two shots as in-context demonstrations.
\end{itemize}

Tested pairs for MedQA and MedMCQA are documented in Table~\ref{table:shots_medqa} and Table~\ref{table:shots_medmcqa}, respectively.
Additionaly, Table~\ref{table:best_icl_prompt} presents how the performances of each ICL \textsc{MedGENIE} model are affected by each pair, unveiling the accuracy effectiveness of artificial examples.
 
Finally, For LLMs with higher context window capacity, such as LLaMA-3-instruct and Phi-3-mini-instruct, our approach is slightly different:
\begin{itemize}[noitemsep]
  \item We sample three questions from the training set and generate a context with PMC-LLaMA for each of them.
  \item We consider only one long-context artificial shot as in-context demonstration.
\end{itemize}
 
This refined strategy ensures effective handling of context constraints while generating accurate and informative demonstrations for each model and benchmark scenario.

\begin{table*}[!htb]
\centering
\adjustbox{width=.95\textwidth}{
\begin{threeparttable}[b]
\small
\begin{tabular}{p{1cm}p{7cm}p{7cm}}
\hline
\multicolumn{3}{c}{\textsc{MedQA}}\\
\hline
\textbf{Context} & \textbf{Shot 1} & \textbf{Shot 2} \\
\hline & \\[-1.5ex]
\textit{Human (H)} & \textsc{Context:} Nitrofurantoin is a commonly used antibiotic for the treatment of uncomplicated urinary tract infections (UTIs) in pregnant women. It is considered safe during pregnancy and is effective against common pathogens causing UTIs. Ampicillin and ceftriaxone are not the first-line choices for treating uncomplicated UTIs, and doxycycline is contraindicated in pregnancy due to potential adverse effects on fetal development.

\vspace{1mm}\textsc{Question:}
A 23-year-old pregnant woman... Which of the following is the best treatment for this patient? A. Ampicillin, B. Ceftriaxone, C. Doxycycline, \textbf{D. Nitrofurantoin}
& \textsc{Context: }Placing the infant in a supine position on a firm mattress while sleeping is the recommended precaution to reduce the risk of sudden infant death syndrome (SIDS). This position helps maintain clear airways and minimizes the risk of suffocation. Avoiding practices such as covering the infant excessively, using devices to maintain sleeping position, and prohibiting pacifier use during sleep are not recommended and may pose additional risks. 

\vspace{1mm} \textsc{Question: }
A 3-month-old baby died suddenly... Which of the following precautions could have prevented the death of the baby? \textbf{A. Placing the infant in a supine position on a firm mattress while sleeping}, B. Keeping the infant covered and maintaining a high room temperature, C. Application of a device to maintain the sleeping position, D. Avoiding pacifier use during sleep\\
\hline & \\[-1.5ex]
\textit{Artificial (A1)} & \textsc{Context:} Most outpatient physicians treat asymptomatic bacteriuria with sulfate-based cephalosporins such as nitrofurantoin (100 mg BID for 7 days) or cephalexin (500mg tid for 7 days). Both drugs are considered safe during pregnancy. 

\vspace{1mm}\textsc{Question:}
A 23-year-old pregnant woman... Which of the following is the best treatment for this patient? A. Ampicillin, B. Ceftriaxone, C. Doxycycline, \textbf{D. Nitrofurantoin}
& \textsc{Context:} Sudden infant death syndrome (SIDS) is the unexpected, sudden death of a child under one year old. An autopsy does not show an explainable cause of death in cases with SIDS. Placing the child in a supine position on a firm mattress while sleeping decreases the risk of SIDS by preventing potential hazards such as soft bedding material or entrapment risks that could compromise respiration.

\vspace{1mm} \textsc{Question: }
A 3-month-old baby died suddenly... Which of the following precautions could have prevented the death of the baby? \textbf{A. Placing the infant in a supine position on a firm mattress while sleeping}, B. Keeping the infant covered and maintaining a high room temperature, C. Application of a device to maintain the sleeping position, D. Avoiding pacifier use during sleep \\
\hline & \\[-1.5ex]
\textit{Artificial (A2)} & \textsc{Context:} Acute hypoxic respiratory failure in the setting of recent surgery for femur fracture suggests pulmonary embolism as the most likely pathogenesis. The histologic section demonstrates a thromboembolus lodged in the lumen of a pulmonary artery. Thrombotic or embolic phenomenon has occurred which led to sudden cardiac arrest (pulmonary\u00a0passive congestion, ischemia, and hypertension are unlikely given that no CAD event or myocardial infarction preceded this acute event). 

\vspace{1mm}\textsc{Question:}
A pulmonary autopsy specimen... Which of the following is the most likely pathogenesis for the present findings? \textbf{A. Thromboembolism}, B. Pulmonary ischemia, C. Pulmonary hypertension, D. Pulmonary passive congestion
& \textsc{Context:} Women with von Willebrand disease (vWD) often present with menorrhagia and easy bruising. The platelet count is usually normal, but the bleeding time and PTT are prolonged. Hemophilia A, lupus anticoagulant, protein C deficiency, or factor V deficiency would not present with these findings on the PTT test. 

\vspace{1mm} \textsc{Question: }
A 20-year-old woman... Which of the following is the most likely cause of this patient\u2019s symptoms? A. Hemophilia A, B. Lupus anticoagulant, C. Protein C deficiency, \textbf{D. Von Willebrand disease}\\
\hline & \\[-1.5ex]

\textit{Artificial (A3)} & \textsc{Context:} The presence of splenomegaly and the finding of immature granulocytic cells in the bone marrow are consistent with this diagnosis. Chronic myeloid leukemia (CML) is characterized by an abnormality involving the ABL1 gene on chromosome 9q, which results in unregulated tyrosine kinase activity. The JAK-STAT pathway, loss of function of the APC gene, altered expression of retinoic acid receptor genes, or induced expression PDGFRA are not associated with CML; these abnormalities can be seen in other types of leukemia or myelodysplastic syndromes (MDS). 

\vspace{1mm} \textsc{Question: }
A 52-year-old... Which of the following mechanisms is most likely responsible for this patient's condition? A. Cytokine-independent activation of the JAK-STAT pathway, B. Loss of function of the APC gene, C. Altered expression of the retinoic acid receptor gene, \textbf{D. Unregulated expression of the ABL1 gene}
& \textsc{Context:} Post-translational modifications (PTMs) are covalent modifications to a polypeptide following its synthesis by the ribosome. The chemically-tagged protein mentioned in the question acts as an E3 ubiquitin ligase by catalyzing the attachment of ubiquitin molecules to lysine residues on targeted proteins, marking them for degradation. Glycosylation involves adding sugar molecules; phosphorylation/dephosphorylation adds or removes phosphate groups and carboxylation involves adding carbon dioxide. Ubiquitination modifies a protein through addition of small, globular proteins called ubiquitins through isopeptide bonds.

\vspace{1mm} \textsc{Question: }
An investigator is studying... Which of the following post-translational modifications has most likely occurred? A. Glycosylation, B. Phosphorylation, \textbf{C. Ubiquitination}, D. Carboxylation\\
\hline
\end{tabular}
\end{threeparttable}}
\caption{Tested 2-shot in-context learning examples for MedQA. The prompts marked with (H) are created by a medical expert, while A* refers to relevant portions of contexts generated by PMC-LLaMA.}
\label{table:shots_medqa}
\end{table*}

\begin{table*}[!htb]
\centering
\begin{threeparttable}[b]
\small
\begin{tabular}{p{1cm}p{7cm}p{7cm}}
\hline
\multicolumn{3}{c}{\textsc{MedMCQA} and \textsc{MMLU}}\\
\hline
\textbf{Context} & \textbf{Shot 1} & \textbf{Shot 2} \\
\hline & \\[-1.5ex]
\textit{Human (H)} & \textsc{Context:} Persistent obstruction of urine flow, as seen in benign prostatic hyperplasia, can result in increased pressure within the urinary system. This elevated pressure may eventually lead to atrophy of the kidney parenchyma. Atrophy refers to the shrinking or reduction in size of an organ or tissue, and in this context, it reflects the consequence of long-term obstruction on the affected kidney.

\vspace{1mm}\textsc{Question:} Chronic urethral obstruction due to benign prismatic hyperplasia can lead to the following change in kidney parenchyma A. Hyperplasia, B. Hyperophy, \textbf{C. Atrophy}, D. Dyplasia

& \textsc{Context:} Vitamin B12, also known as cobalamin, is primarily found in animal products such as meat, fish, eggs, and dairy. It is not naturally present in significant amounts in plant-based foods, making it essential for individuals following a vegetarian or vegan diet to obtain this vitamin through fortified foods or supplements. In contrast, Vitamin C, Vitamin B7 (biotin), and Vitamin D can be obtained from both animal and plant sources. 

\vspace{1mm} \textsc{Question: }Which vitamin is supplied from only animal source: \textbf{A. Vitamin B12}, B. Vitamin B7, C. Vitamin C, D. Vitamin D
\\
\hline & \\[-1.5ex]
\textit{Artificial (A1)} & \textsc{Context:} Hyperviscosity is a condition where the blood becomes abnormally thick, hindering its ability to flow properly. Cryoglobulinemia is a condition characterized by abnormal antibodies in the blood (antibodies are specialized cells that recognize and attack foreign invaders). These abnormal antibodies become solid at cold temperatures and lead to clumping of red blood cells, an increase in viscosity, and subsequent obstruction of small vessels.

\vspace{1mm}\textsc{Question:} Hyper viscosity is seen in \textbf{A. Cryoglobulinemia}, B. Multiple myeloma, C. MGUS, D. Lymphoma

& \textsc{Context:} Inversion of the foot refers to a foot in which its sole faces medially. Since the plantar surface of the foot is in contact with the ground at all times, this condition occurs when one or more muscles responsible for moving it become tight or weak, resulting in an alteration in normal alignment.

\vspace{1mm} \textsc{Question:} In inversion of the foot, the sole will face: A. Upwards, B. Downwards, C. Laterally, \textbf{D. Medially}
\\
\hline & \\[-1.5ex]
\textit{Artificial (A2)} & \textsc{Context:} Antibiotic prophylaxis for infective endocarditis is indicated in individuals with predisposing cardiac conditions. In this scenario, determining if an isolated secundum ASD and mitral valve prolapse without regurgitation are associated with the potential risk of developing infective endocarditis requires further information. The presence of a prior coronary aery bypass graft and coarctation of aoa are both established indications for antibiotic prophylaxis due to their association with infective endocarditis risk.

\vspace{1mm}\textsc{Question:} Antiboiotic Prophylaxis for infective endocarditis is indicated in: A. Isolated secundum ASD, B. Mitral valve prolapse without regurgitation, C. Prior coronary aery bypass graft, \textbf{D. Coarctation of aoa}

& \textsc{Context:} The Anterolateral portal is also known as the lateral portal. It is used for viewing the patellofemoral joint, inserting probe or laser for soft-tissue procedures.

\vspace{1mm} \textsc{Question:} Anterolateral ahroscopy of knee is for: \textbf{A. To see patellofemoral aiculation}, B. To see the posterior cruciate ligament, C. To see the anterior poion of lateral meniscus, D. To see the periphery of the posterior horn of medial meniscus\\
\hline & \\[-1.5ex]

\textit{Artificial (A3)} & \textsc{Context:} Ligamentum teres is a degenerative string of tissue that exists in the fetal remnant of umbilical vein. In adults, it runs along the inferior margin of the liver and functions as both an anatomic landmark and as part of a ligamentous structure that connects the falciform ligament with the round, triangular, and coronary ligaments.

\vspace{1mm} \textsc{Question:} Ligament teres is a remnant of ? A. Ductus aeriosus, B. Umbilical aery, \textbf{C. Umbilical vein}, D. Ductus venosus

& \textsc{Context:} The Magic syndrome refers to the presence of mouth and genital ulcers. Although initially thought to represent a distinct disease entity, it is now recognized as a subset of Behcet disease.

\vspace{1mm} \textsc{Question:} Magic syndrome is seen in: \textbf{A. Behcet disease}, B. Aphthous major, C. Herpetiform, D. Bloom syndrome
\\
\hline
\end{tabular}
\end{threeparttable}
\caption{Tested 2-shot in-context learning examples for MedMCQA and MMLU. The prompts marked with (H) are created by a medical expert, while A* refers to relevant portions of contexts generated by PMC-LLaMA.}
\label{table:shots_medmcqa}
\end{table*}

\begin{table}[!htb]
\centering
\adjustbox{width=\linewidth}{
\begin{tabular}{lcccc}
\hline
\textbf{Dataset} & \textbf{H} & \textbf{A1} & \textbf{A2} & \textbf{A3} \\
\hline
\rowcolor{lightgrey}\multicolumn{5}{c}{MedGENIE-LLaMA-2-chat} \\
\hline
MedQA  &  52.4  & \textbf{52.6}  & 51.5 & 51.6 \\
MedMCQA  & 44.1 & \textbf{44.8} & 44.3 & 44.7 \\
MMLU & 56.5 & 58.2 & \textbf{58.8} & 56.6 \\
\hline
\rowcolor{lightgrey}\multicolumn{5}{c}{MedGENIE-Zephyr-$\beta$} \\
\hline
MedQA  &  \textbf{59.7}  & 58.9  & 58.8 & 57.8 \\
MedMCQA & 50.5 & 50.8 & \textbf{51.0} & 50.8 \\
MMLU & 65.1 & \textbf{66.1} & 66.0 & 65.3 \\
\hline
\end{tabular}
}
\caption{Effect of each tested pair of shots on the accuracy of each ICL \textsc{MedGENIE} model. H, A1, A2, A3 refer to the shot examples provided in Table \ref{table:shots_medqa} and Table \ref{table:shots_medmcqa} for corresponding benchmarks. Best results are in bold.}
\label{table:best_icl_prompt}
\end{table}

\section{Clustering-Based Prompting}
\label{app:clustering_based_prompting}

The cluster-based methodology by \citet{DBLP:conf/iclr/0002IWXJ000023} can be detailed as follows:
\begin{enumerate}[noitemsep,topsep=0pt]
    \item ask an LLM to generate a background context for each question in the training split (alternatively, we can retrieve a context from a knowledge source), obtaining a set of question--context pair;
    \item perform inference with the LLM by leveraging the generated context, and maintain from the initial set of pairs only those for which the LLM answered correctly;
    \item encode each question--context pair from the filtered set;
    \item use $K$-means to cluster all the embedding vectors obtained into $K$ classes, where the number of classes matches the number of documents that need to be generated eventually;
    \item randomly select $n$ question--context pairs from each cluster;
    \item present each different $n$ question--context pairs to the LLM as in-context demonstrations for generating a context from a given test question.
\end{enumerate}
This approach ensures that the LLM is exposed to different distributions of examples, resulting in generated contexts that cover various (uncontrolled) perspectives.
We investigate the potential benefits deriving from the integration of this methodology into \textsc{MedGENIE} ICL pipeline, tested on MedQA.
For point (1), we implement the strategy using both contexts retrieved from MedWiki and contexts artificially generated with PMC-LLaMA.
We set $K{=}5$ and $n{=}3$.
The results are summarized in Table~\ref{table:clustering_based_prompt}.
Integration leads to a notable improvement of up to +2.4 points in accuracy.
However, it comes with a considerably higher cost compared to \textsc{MedGENIE} alone, primarily due to the additional operations required at points (1) and (2).
These operations are unnecessary for \textsc{MedGENIE}, making it a more cost-effective option, especially in scenarios with larger training sets, such as MedMCQA.

\begin{table}[!htb]
\centering
\adjustbox{width=\linewidth}{
\begin{threeparttable}[b]
\begin{tabular}{lcll}
\hline
\textbf{Retrieval source} & \textbf{Cluster-based} & \textbf{LLaMA-2-chat} & \textbf{Zephyr-$\beta$} \\
\hline
-- & \XSolidBrush & 52.6  & 59.7  \\
MedWiki  & \CheckmarkBold & 52.3  & 59.8 \textcolor{teal1}{(+0.1)} \\
PMC-LLaMA  & \CheckmarkBold & \textbf{54.3} \textcolor{teal2}{(+1.7)} & \textbf{62.1} \textcolor{teal2}{(+2.4)}\\
\hline
\end{tabular}
\begin{tablenotes}[para]
\textbf{Note:} The positive influence (i.e., $\Delta$ Acc.) of incorporating clustering-based prompting within the \textsc{MedGENIE} is highlighted using colors.
\end{tablenotes}
\end{threeparttable}}
\caption{Comparison of clustering-based prompting for context generation with the standard full-generative \textsc{MedGENIE} ICL pipeline on the MedQA test set.}
\label{table:clustering_based_prompt}
\end{table}

\section{Full Baselines}
\label{app:full_baselines}

In Table~\ref{table:medqa_results}, Table~\ref{table:medmcqa_results}, and Table~\ref{table:mmlu_results}, we present a fine-grained comparison of MedGENIE models with closed and open-book alternatives known in the literature, both fine-tuned and in zero/few-shot modes.
It should be noted that the baseline results provided by \citet{DBLP:journals/corr/abs-2207-08143} for MedMCQA should only be regarded as a reference point, as they relate to a subset of 1K samples from the eval~set.

\begin{table*}[!htb]
\centering
\adjustbox{width=.85\linewidth}{
\begin{threeparttable}[b]
\begin{tabular}{cllllc}
\hline
& \textbf{Model} & \textbf{Ground (Source)} & \textbf{Learning} & \textbf{Params} & \textbf{MedQA ($\downarrow$)} \\
\hline
\multirow{31}{*}{\includegraphics[width=.35cm]{figures/closed_book.png}} & LLaMA-3-Instruct & $\varnothing$ & 1-shot & 8B & 60.6 \\
& LLaMA-3-Meditron\tnote{14} & $\varnothing$ & ?\tnote{**} & 8B & 60.6 \\
& OpenBioLLM\tnote{15} & $\varnothing$ & ?\tnote{**} & 8B & 59.0 \\
& Phi-3-mini & $\varnothing$ & 1-shot & 3.8B & 55.1 \\
& Codex\tnote{1} & $\varnothing$ & 0-zhot & 175B & 52.5 \\
& MEDITRON\tnote{2} & $\varnothing$ & Fine-tuned & 7B & 52.0 \\
& BioMistral DARE\tnote{12}	& $\varnothing$ & Fine-tuned & 7B & 51.1 \\
& BioMistral\tnote{12}	& $\varnothing$	& Fine-tuned & 7B & 50.6 \\
& BioMedGPT\tnote{3} & $\varnothing$ & \textit{k}-shot\tnote{**} & 10B & 50.4 \\
& BioMedLM\tnote{4} & $\varnothing$ & Fine-tuned & 2.7B & 50.3 \\
& PMC-LLaMA\tnote{*} & $\varnothing$ & Fine-tuned & 13B & 50.2 \\
& LLaMA-2\tnote{2} & $\varnothing$ & Fine-tuned & 7B & 49.6 \\
& Zephyr-$\beta$ & $\varnothing$ & 2-shot & 7B & 49.6 \\
& Zephyr-$\beta$\tnote{2} & $\varnothing$ & 3-shot & 7B & 49.2 \\
& PMC-LLaMA\tnote{2} & $\varnothing$ & Fine-tuned & 7B & 49.2 \\
& Flan-PaLM\tnote{4} & $\varnothing$ & 5-shot & 62B & 46.1 \\
& InstructGPT\tnote{1} & $\varnothing$ & 0-shot & 175B & 46.0 \\
& Vicuna 1.3\tnote{1} & $\varnothing$ & 0-shot & 33B & 45.2 \\
& BioLinkBERT\tnote{4} & $\varnothing$ & Fine-tuned & 340M & 45.1 \\
& Galactica & $\varnothing$ & 0-shot & 120B & 44.4 \\
& LLaMA-2\tnote{1} & $\varnothing$ & 0-shot & 70B & 43.4 \\
& Guanaco\tnote{1} & $\varnothing$ & 0-shot & 33B & 42.9 \\
& LLaMA-2-chat\tnote{1} & $\varnothing$ & 0-shot & 70B & 42.3 \\
& Vicuna 1.5\tnote{1} & $\varnothing$ & 0-shot & 65B & 41.6 \\
& Mistral-Instruct\tnote{2} & $\varnothing$ & 3-shot & 7B & 41.1 \\
& PaLM\tnote{4} & $\varnothing$ & 5-shot & 62B & 40.9 \\
& Guanaco\tnote{1} & $\varnothing$ & 0-shot & 65B & 40.8 \\
& MeditronLlama-Slerp\tnote{13} & $\varnothing$ & \textit{k}-shot\tnote{**} & 7B & 39.20  \\
& Falcon-Instruct\tnote{1} & $\varnothing$ & 0-shot & 40B & 39.0 \\
& Vicuna 1.3\tnote{1} & $\varnothing$ & 0-shot & 13B & 38.7 \\
& PubMedBERT\tnote{4} & $\varnothing$ & Fine-tuned & 110M & 38.1 \\
& LLaMA-2-chat & $\varnothing$ & 2-shot & 7B & 37.2 \\
& BioBERT\tnote{5} & $\varnothing$ & Fine-tuned & 110M & 36.7 \\
& MTP-Instruct\tnote{1} & $\varnothing$ & 0-shot & 30B & 35.1 \\
& GPT-Neo\tnote{4} & $\varnothing$ & Fine-tuned & 2.5B & 33.3 \\
\hline
\multirow{15}{*}{\includegraphics[width=.6cm]{figures/open_book.png}} & \textbf{MedGENIE-Phi-3-mini} & G (\small{PMC-LLaMA}) & 1-shot &	3.8B & 	\textbf{64.7} \\
& \textbf{MedGENIE-LLaMA-3-Instruct} & G (\small{PMC-LLaMA}) & 1-shot &	8B & \underline{63.1} \\
& \textbf{MedGENIE-Zephyr-$\beta$} & G (PMC-LLaMA) & 2-shot & 7B & 59.7 \\
& \textbf{MedGENIE-FID-Flan-T5} & G (PMC-LLaMA) & Fine-tuned & 250M & 53.1 \\
& \textbf{MedGENIE-LLaMA-2-chat} & G (PMC-LLaMA) & 2-shot & 7B & 52.6 \\
& Codex\tnote{1} & R (Wikipedia) & 0-shot & 175B & 52.5 \\
& GPT-3.5-Turbo\tnote{6} & R (Wikipedia) & \textit{k}-shot\tnote{**} & -- & 52.3 \\
& Zephyr-$\beta$ & R (Textbooks) & 2-shot & 7B & 51.4 \\
& Zephyr-$\beta$ & R (MedWiki) & 2-shot & 7B & 50.4 \\
& DRAGON\tnote{7} & R (UMLS) & Fine-tuned & 360M & 47.5 \\
& InstructGPT\tnote{1} & R (Wikipedia) & 0-shot & 175B & 47.3 \\
& VOD\tnote{8} & R (MedWiki) & Fine-tuned & 220M & 45.8 \\
& Mistral-Instruct & R (MedWiki) & 2-shot & 7B & 45.1 \\
& BioReader\tnote{9} & R (PubMed-RCT) & Fine-tuned & 230M & 43.0 \\
& GreaseLM\tnote{10} & R (UMLS) & Fine-tuned & 359M & 38.5 \\
& QA-GNN\tnote{11} & R (UMLS) & Fine-tuned & 360M & 38.0 \\
& LLaMA-2\tnote{6} & R (Wikipedia) & \textit{k}-shot\tnote{**} & 13B & 37.6\\
& LLaMA-2-chat & R (Textbooks) & 2-shot & 7B & 37.9 \\
& LLaMA-2-chat & R (MedWiki) & 2-shot & 7B & 37.2 \\
\hline
\end{tabular}
\begin{tablenotes}[para]
\item[1] \citep{DBLP:journals/corr/abs-2207-08143} \item[2] \citep{DBLP:journals/corr/abs-2311-16079} \item[3] \citep{DBLP:journals/corr/abs-2308-09442} \item[4] \citep{DBLP:journals/corr/abs-2212-13138} \item[5] \citep{DBLP:journals/bioinformatics/LeeYKKKSK20} \item[6] \citep{DBLP:journals/corr/abs-2309-02233} \item[7] \citep{DBLP:conf/nips/YasunagaBR0MLL22} \item[8] \citep{DBLP:conf/icml/LievinMJW23} \item[9] \citep{frisoni-etal-2022-bioreader} \item[10] \citep{DBLP:journals/corr/abs-2201-08860}  \item[11] \citep{yasunaga-etal-2021-qa}\\
\item[12] \citep{biomistral}
\item[13] \citep{merge-kit}
\item[14] \href{https://www.meditron.io/}{https://www.meditron.io}
\item[15] \href{https://huggingface.co/aaditya/Llama3-OpenBioLLM-8B}{aaditya/Llama3-OpenBioLLM-8B} \\
\item[*] AWQ 4-bit inference. 
\item[**] Lack of inference details. 
\end{tablenotes}
\end{threeparttable}} 
\caption{Accuracy comparison on MedQA. Bold and underline denote the best and second-best scores.}
\label{table:medqa_results}
\end{table*}

\begin{table*}[!htb]
\centering
\adjustbox{width=.9\linewidth}{
\begin{threeparttable}[b]
\begin{tabular}{cllllc}
\hline
& \textbf{Model} & \textbf{Ground (Source)} & \textbf{Learning} & \textbf{Params} & \textbf{MedMCQA ($\downarrow$)} \\
\hline
\multirow{26}{*}{\includegraphics[width=.35cm]{figures/closed_book.png}} & MEDITRON\tnote{1} & $\varnothing$ & Fine-tuned & 7B & \textbf{59.2}\\
& LLaMA-3-Meditron\tnote{9} & $\varnothing$ & ?\tnote{**} & 8B & 58.4 \\
& PMC-LLaMA\tnote{*} & $\varnothing$ & Fine-tuned  & 13B & 57.6 \\
& OpenBioLLM\tnote{10} & $\varnothing$ & ?\tnote{**} & 8B & 56.9 \\
& LLaMA-3-Instruct & $\varnothing$ & 1-shot & 8B & 55.7  \\
& LLaMA-2\tnote{1}  & $\varnothing$ & Fine-tuned & 7B  & 54.4\\
& Phi-3-mini & $\varnothing$ & 1-shot	& 3.8B & 53.5  \\
& Galactica\tnote{2} & $\varnothing$ & Fine-tuned & 120B & 52.9 \\
& PMC-LLaMA\tnote{1}  & $\varnothing$ & Fine-tuned & 7B  & 51.5 \\
& BioMedGPT\tnote{3}  & $\varnothing$ & Fine-tuned & 10B  & 51.4 \\
& Codex\tnote{4} & $\varnothing$ & 0-shot & 175B & 50.9\tnote{**}\\
& BioMistral DARE\tnote{7}	& $\varnothing$ & Fine-tuned & 7B & 48.7 \\
& BioMistral\tnote{7}	& $\varnothing$	& Fine-tuned & 7B & 48.1 \\
& Flan-PaLM\tnote{2} & $\varnothing$ & 5-shot & 62B & 46.2 \\
& InstructGPT\tnote{4} & $\varnothing$ & 0-shot & 175B & 44.0\tnote{**}\\
& PaLM\tnote{2} & $\varnothing$ & 5-shot & 62B & 43.4 \\
& Zephyr-$\beta$\tnote{1} & $\varnothing$ & 3-shot & 7B  & 43.0\\
& Llama-2\tnote{4} & $\varnothing$ & 0-shot & 70B & 42.8\tnote{**}\\  
& Zephyr-$\beta$ & $\varnothing$ & 2-shot & 7B  & 42.5\\
& Llama-2-chat\tnote{4} & $\varnothing$ & 0-shot & 70B & 41.8\tnote{**}\\
& Vicuna 1.5\tnote{4} & $\varnothing$ & 0-shot & 13B & 41.5\tnote{**}\\
& Mistral-Instruct\tnote{1} & $\varnothing$ & 3-shot & 7B & 40.2 \\
& Vicuna 1.3\tnote{4} & $\varnothing$ & 0-shot & 65B & 38.3\tnote{**}\\ 
& Vicuna 1.3\tnote{4} & $\varnothing$ & 0-shot & 65B & 38.0\tnote{**}\\  
& Guanaco\tnote{4} & $\varnothing$ & 0-shot & 33B & 37.4\tnote{**}\\ 
& MeditronLlama-Slerp\tnote{8} & $\varnothing$ & \textit{k}-shot\tnote{**} & 7B & 36.9  \\
& Guanaco\tnote{4} & $\varnothing$ & 0-shot & 65B & 36.7\tnote{**}\\
& LLaMA-2-chat\tnote{4} & $\varnothing$ & 0-shot & 13B & 36.6\tnote{**}\\
& LLaMA-2-chat & $\varnothing$ & 2-shot & 7B  & 35.1\\
& MPT-Instruct\tnote{4} & $\varnothing$ & 0-shot & 20B & 34.6\tnote{**}\\
& LLaMA-2\tnote{4} & $\varnothing$ & 0-shot & 13B & 31.7\tnote{**}\\
& Falcon-Instruct\tnote{4} & $\varnothing$ & 0-shot & 20B & 30.0\tnote{**}\\
& GPT-NeoX\tnote{4} & $\varnothing$ & 0-shot & 20B & 27.8\tnote{**}\\
\hline
\multirow{11}{*}{\includegraphics[width=.6cm]{figures/open_book.png}} & VOD \tnote{5}&  R (MedWiki) & Fine-tuned & 220M & \underline{58.3} \\
& \textbf{MedGENIE-LLaMA-3-Instruct} & G (\small{PMC-LLaMA}) & 1-shot &	8B & 56.2 \\
& \textbf{MedGENIE-Phi-3-mini} & G (\small{PMC-LLaMA}) & 1-shot &	3.8B & 	54.1 \\
& \textbf{MedGENIE-FID-Flan-T5} & G (PMC-LLaMA) & Fine-tuned & 250M & 52.1 \\
& \textbf{MedGENIE-Zephyr-$\beta$} & G (PMC-LLaMA) & 2-shot & 7B & 51.0 \\
& Zephyr-$\beta$ & R (MedWiki) & 2-shot & 7B & 47.1 \\
& InstructGPT \tnote{4}  &  R (Wikipedia) & 0-shot & 175B & 46.7\tnote{**}\\
& \textbf{MedGENIE-LLaMA-2-chat} & G (PMC-LLaMA) & 2-shot & 7B & 44.8 \\
& Mistral-Instruct & R (MedWiki) & 2-shot & 7B & 44.3 \\
& PubmedBERT \tnote{6}  & R (Pubmed) & Fine-tuned & 110M & 43.0\\

& SciBERT \tnote{6} & R (Pubmed) & Fine-tuned & 110M  & 41.0 \\
& BioBERT \tnote{6} & R (Pubmed) & Fine-tuned & 110M  & 39.0 \\
& LLaMA-2-chat & R (MedWiki) & 2-shot & 7B & 37.3 \\
\hline
\end{tabular}
\begin{tablenotes}[para]
\item [1] \citep{DBLP:journals/corr/abs-2311-16079} \item[2] \citep{DBLP:journals/corr/abs-2211-09085} \item [3] \citep{DBLP:journals/corr/abs-2308-09442} \item[4] \citep{DBLP:journals/corr/abs-2207-08143} \item[5] \citep{DBLP:conf/icml/LievinMJW23} \item[6] \citep{DBLP:conf/chil/PalUS22}
\item[7] \citep{biomistral}
\item[8] \citep{merge-kit}
\item[9] \href{https://www.meditron.io/}{https://www.meditron.io}
\item[10] \href{https://huggingface.co/aaditya/Llama3-OpenBioLLM-8B}{aaditya/Llama3-OpenBioLLM-8B} \\
\item [*] AWQ 4-bit inference. \item[**] Estimated using 1k samples.
\end{tablenotes}
\end{threeparttable}} 
\caption{Accuracy comparison on MedMCQA. Bold and underline denote the best and second-best scores.}
\label{table:medmcqa_results}
\end{table*}

\begin{table*}[!htb]
\centering
\adjustbox{width=.85\linewidth}{
\begin{threeparttable}[b]
\begin{tabular}{cllllc}
\hline
& \textbf{Model} & \textbf{Ground (Source)} & \textbf{Learning} & \textbf{Params} & \textbf{MMLU ($\downarrow$)} \\
\hline
\multirow{8}{*}{\includegraphics[width=.35cm]{figures/closed_book.png}} & Phi-3-mini & $\varnothing$ & 1-shot	& 3.8B & 70.3 \\
& LLaMA-3-Instruct & $\varnothing$ & 1-shot & 8B & 69.8 \\
& Zephyr-$\beta$\tnote{1} & $\varnothing$ & 3-shot & 7B  & \underline{63.3}\\
& Galactica\tnote{2} & $\varnothing$ & Fine-tuned\tnote{*} & 120B & 61.8 \\
& Zephyr-$\beta$ & $\varnothing$ & 2-shot & 7B  & 60.5\\
& PMC-LLaMA\tnote{1} & $\varnothing$ & Fine-tuned\tnote{*}  & 7B & 59.7 \\
& LLaMA-2\tnote{1} & $\varnothing$ & Fine-tuned\tnote{*} & 7B  & 56.3\\
& Mistral-Instruct\tnote{1} & $\varnothing$ & 3-shot & 7B & 55.8 \\
& MEDITRON\tnote{1}  & $\varnothing$ & Fine-tuned\tnote{*} & 7B &  55.6 \\
& LLaMA-2-chat & $\varnothing$ & 2-shot & 7B  & 49.4\\
\hline
\multirow{7}{*}{\includegraphics[width=.6cm]{figures/open_book.png}} & \textbf{MedGENIE-Phi-mini} & G (\small{PMC-LLaMA}) & 1-shot &	8B & \textbf{70.8} \\
& \textbf{MedGENIE-LLaMA-3-Instruct} & G (PMC-LLaMA) & 2-shot & 7B & 68.9 \\
& Zephyr-$\beta$ & R (MedWiki) & 2-shot & 7B & 66.9 \\
& \textbf{MedGENIE-FID-Flan-T5} & G (PMC-LLaMA) & Fine-tuned\tnote{*} & 250M & 59.9  \\
& \textbf{MedGENIE-LLaMA-2-chat} & G (PMC-LLaMA) & 2-shot & 7B & 58.8  \\
& Mistral-Instruct & R (MedWiki) & 2-shot & 7B & 58.5 \\
& VOD\tnote{2} & R (MedWiki) & Fine-tuned\tnote{*} & 220M & 56.8 \\
& LLaMA-2-chat & R (MedWiki) & 2-shot & 7B & 52.0 \\
\hline
\end{tabular}
\begin{tablenotes}[para]
\item [1] \citep{DBLP:journals/corr/abs-2311-16079} \item [2] \citep{DBLP:conf/icml/LievinMJW23}\\
\item [*] Fine-tuned on the MedMCQA train set.
\end{tablenotes}
\end{threeparttable}} 
\caption{Accuracy comparison on MMLU-Medical. Bold and underline denote the best and second-best scores.}
\label{table:mmlu_results}
\end{table*}

\section{Robustness}

\subsection{Sensitivity to Question Subject}
\label{sec:a_sensitivity_subject}

In Figure~\ref{fig:topic_distribution}, we present a detailed analysis of the \textit{per-subject} performance of the \textsc{MedGENIE} models across MedMCQA (21 medical subjects) and MMLU (9 health and biology subjects).
For MedMCQA, the analysis is done in comparison with our baselines: Zephyr-$\beta$ and LLaMA-2-chat in 2-shot prompting.
For MMLU, we extend the analysis to known results from MEDITRON, PMC-LLaMA, and VOD, all fine-tuned over the train set of MedMCQA.
In this case, we consider Zephyr-$\beta$ in 3-shot prompting \citep{DBLP:journals/corr/abs-2311-16079} due to its higher accuracy.
Please note that the subcategory \texttt{professional\_medicine} of MMLU corresponds to the MedQA questions.

\begin{figure*}[!htb]
    \centering
    
    \begin{subfigure}[t]{\linewidth}
        \centering
        \begin{tikzpicture}
            \fontsize{6.5}{6.5}\selectfont
            \node[draw=black, inner sep=2pt, align=center] {\begin{tabular}{llll}
                \textbf{Ours:} & \ref{plot:medgenie_fid} MedGENIE-FID-Flan-T5 (250M, Fine-tuned) & \ref{plot:medgenie_llama} MedGENIE-LLaMA-2-chat (7B, ICL) & \ref{plot:medgenie_zephyr} MedGENIE-Zephyr-$\beta$ (7B, ICL)\\
                \arrayrulecolor{darkgrey}\hline
                \textbf{Baselines:} & \ref{plot:pmcllama_13} PMC-LLaMA (13B, Fine-tuned) & \ref{plot:llama2} LLaMA-2 (7B, ICL) & \ref{plot:meditron} Meditron (7B, Fine-tuned)\\
                & \ref{plot:vod} VOD (220M, Fine-tuned) & \ref{plot:zephyr} Zephyr-$\beta$ (7B, ICL) 
            \end{tabular}};
        \end{tikzpicture}
    \end{subfigure}\\[1cm]
    \begin{subfigure}[t]{\linewidth}
        \centering
        \begin{tikzpicture}[node font=\small, remember picture]
            \begin{axis}[
                title={\textbf{MedMCQA}},
                width=\linewidth, height=4cm,
                major grid style={gray},
                minor grid style={gray},
                axis background/.style={fill=plotbackground!50},
                axis x line=bottom,
                symbolic x coords={Physiology, Medicine, Biochemistry, Ophthalmology, Anatomy, Pediatrics, Pathology, Dental, Microbiology, Anaesthesia, Radiology},
                xticklabel style={align=right},
                xticklabels={\textbf{Physiology}\\(4.1\%), \textbf{Medicine}\\(7.1\%), \textbf{Biochemistry}\\(4.1\%), \textbf{Ophthalmology}\\(1.4\%), \textbf{Anatomy}\\(5.6\%), \textbf{Pediatrics}\\(5.6\%), \textbf{Pathology}\\(8.1\%), \textbf{Dental}\\(31.5\%), \textbf{Microbiology}\\(2.9\%), \textbf{Anaesthesia}\\(0.8\%), \textbf{Radiology}\\(1.7\%)},
                xtick=data,
                xlabel=\empty,
                xticklabel style={rotate=40},
                enlarge x limits=0.08,
                axis y line=left,
                ymin=0, ymax=85,
                ytick={0, 10, ..., 80},
                ymajorgrids=true,
                scaled y ticks = false,
                ylabel={Accuracy},
                ybar,
                bar width=2.8pt,
                nodes near coords,
                point meta=rawy, 
                every node near coord/.append style={
                    font=\fontsize{3}{4}\selectfont,
                    /pgf/number format/precision=0,
                    /pgf/number format/set thousands separator={},
                    rotate=90,
                    anchor=west,
                },
                every tick label/.append style={font=\fontsize{7}{7}\selectfont},
            ]
            
            \addplot[fill=teal1, draw=none] coordinates {(Physiology, 53.8) (Medicine, 58.0) (Biochemistry, 59.6) (Ophthalmology, 63.8) (Anatomy, 59.4) (Pediatrics, 51.7) (Pathology, 60.8) (Dental, 45.2) (Microbiology, 54.1) (Anaesthesia, 38.2) (Radiology, 40.6)};

            \addplot[fill=teal2, draw=none] coordinates {(Physiology, 49.1) (Medicine, 51.5) (Biochemistry, 53.2) (Ophthalmology, 46.6) (Anatomy, 46.2) (Pediatrics, 44.0) (Pathology, 56.1) (Dental, 37.1) (Microbiology, 54.1) (Anaesthesia, 20.6) (Radiology, 37.7)};

            \addplot[fill=teal3, draw=none] coordinates {(Physiology, 59.6) (Medicine, 52.5) (Biochemistry, 64.9) (Ophthalmology, 62.1) (Anatomy, 51.7) (Pediatrics, 49.6) (Pathology, 62.3) (Dental, 42.6) (Microbiology, 51.6) (Anaesthesia, 29.4) (Radiology, 59.4)};

            \addplot[fill=warm2, draw=none] coordinates {(Physiology, 37.4) (Medicine, 36.6) (Biochemistry, 42.7) (Ophthalmology, 37.9) (Anatomy, 33.8) (Pediatrics, 35.5) (Pathology, 34.7) (Dental, 33.0) (Microbiology, 41.0) (Anaesthesia, 29.4) (Radiology, 30.4)};

            \addplot[fill=warm5, draw=none] coordinates {(Physiology, 52.6) (Medicine, 45.8) (Biochemistry, 56.7) (Ophthalmology, 50.0) (Anatomy, 36.3) (Pediatrics, 41.0) (Pathology, 52.8) (Dental, 36.3) (Microbiology, 45.9) (Anaesthesia, 14.7) (Radiology, 53.6)};
            
            \end{axis}
        \end{tikzpicture}
    \end{subfigure}\\
    \begin{subfigure}[t]{\linewidth}
        \centering
        \begin{tikzpicture}[node font=\small, remember picture]
            \begin{axis}[
                width=\linewidth, height=4cm,
                major grid style={gray},
                minor grid style={gray},
                axis background/.style={fill=plotbackground!50},
                axis x line=bottom,
                symbolic x coords={Gynaecology \& Obstetrics, Pharmacology, Social \& Preventive Medicine, ENT, Surgery, Forensic Medicine, Psychiatry, Skin, Orthopaedics, Unknown},
                xticklabel style={align=right},
                xticklabels={\textbf{Gynaecology \&}\\\textbf{Obstetrics}\\(5.4\%), \textbf{Pharmacology}\\(5.8\%), \textbf{Social \& Preventive}\\\textbf{Medicine}\\(3.1\%), \textbf{ENT}\\(1.3\%), \textbf{Surgery}\\(8.8\%), \textbf{Forensic Medicine}\\(1.6\%), \textbf{Psychiatry}\\(0.4\%), \textbf{Skin}\\(0.4\%), \textbf{Orthopaedics}\\(0.5\%), \textbf{Unknown}\\(0.1\%)},
                xtick=data,
                xlabel=\empty,
                xticklabel style={rotate=40},
                enlarge x limits=0.08,
                axis y line=left,
                ymin=0, ymax=85,
                ytick={0, 10, ..., 80},
                ymajorgrids=true,
                scaled y ticks = false,
                ylabel={Accuracy},
                ybar,
                bar width=2.8pt, 
                nodes near coords,
                point meta=rawy, 
                every node near coord/.append style={
                    font=\fontsize{3}{4}\selectfont,
                    /pgf/number format/precision=0,
                    /pgf/number format/set thousands separator={},
                    rotate=90,
                    anchor=west,
                },  
                every tick label/.append style={font=\fontsize{7}{7}\selectfont},
            ]
            
            \addplot[fill=teal1, draw=none] coordinates {(Gynaecology \& Obstetrics, 51.3) (Pharmacology, 61.3) (Social \& Preventive Medicine, 48.1) (ENT, 50.9) (Surgery, 53.7) (Forensic Medicine, 47.8) (Psychiatry, 68.8) (Skin, 35.3) (Orthopaedics, 55.0) (Unknown, 0.0)};

            \addplot[fill=teal2, draw=none] coordinates {(Gynaecology \& Obstetrics, 44.6) (Pharmacology, 52.7) (Social \& Preventive Medicine, 38.0) (ENT, 47.2) (Surgery, 46.3) (Forensic Medicine, 46.3) (Psychiatry, 62.5) (Skin, 47.1) (Orthopaedics, 50.0) (Unknown, 0.0)};

            \addplot[fill=teal3, draw=none] coordinates {(Gynaecology \& Obstetrics, 50.4) (Pharmacology, 61.7) (Social \& Preventive Medicine, 49.6) (ENT, 45.3) (Surgery, 52.0) (Forensic Medicine, 49.3) (Psychiatry, 62.5) (Skin, 52.9) (Orthopaedics, 60.0) (Unknown, 50.0)};

            \addplot[fill=warm2, draw=none] coordinates {(Gynaecology \& Obstetrics, 34.4) (Pharmacology, 35.4) (Social \& Preventive Medicine, 36.4) (ENT, 28.3) (Surgery, 31.7) (Forensic Medicine, 47.8) (Psychiatry, 50.0) (Skin, 52.9) (Orthopaedics, 45.0) (Unknown, 0.0)};

            \addplot[fill=warm5, draw=none] coordinates {(Gynaecology \& Obstetrics, 41.5) (Pharmacology, 57.2) (Social \& Preventive Medicine, 45.0) (ENT, 49.1) (Surgery, 44.2) (Forensic Medicine, 43.3) (Psychiatry, 50.0) (Skin, 41.2) (Orthopaedics, 30.0) (Unknown, 0.0)}; \label{plot:zephyr}
            
            \end{axis}
        \end{tikzpicture}
    \end{subfigure}\\
    \begin{subfigure}[t]{\linewidth}
        \centering
        \begin{tikzpicture}[node font=\small, remember picture]
            \begin{axis}[
                title={\textbf{MMLU}}, 
                width=\linewidth, height=4cm,
                major grid style={gray},
                minor grid style={gray},
                axis background/.style={fill=plotbackground!50},
                axis x line=bottom,
                symbolic x coords={Anatomy, Clinical Knowledge, College Biology, College Medicine, High School Biology, Medical Genetics, Nutrition, Professional Medicine, Virology},
                xticklabel style={align=right},
                xticklabels={\textbf{Anatomy}\\(7.2\%), \textbf{Clinical Knowledge}\\(14.2\%), \textbf{College Biology}\\(7.7\%), \textbf{College Medicine}\\(9.2\%), \textbf{High School Biology}\\(16.6\%), \textbf{Medical Genetics}\\(5.3\%), \textbf{Nutrition}\\(16.4\%), \textbf{Professional Medicine}\\(14.6\%), \textbf{Virology}\\(8.9\%)},
                xtick=data,
                xlabel=\empty,
                xticklabel style={rotate=40},
                enlarge x limits=0.08,
                axis y line=left,
                ymin=35, ymax=85,
                ytick={40, 50, ..., 80},
                ymajorgrids=true,
                scaled y ticks = false,
                ylabel={Accuracy},
                ybar,
                bar width=2.8pt,
                nodes near coords,
                point meta=rawy, 
                every node near coord/.append style={
                    font=\fontsize{3}{4}\selectfont,
                    /pgf/number format/precision=0,
                    /pgf/number format/set thousands separator={},
                    rotate=90,
                    anchor=west,
                },
                every tick label/.append style={font=\fontsize{7}{7}\selectfont},
            ]
            
            \addplot[fill=teal1, draw=none] coordinates {(Anatomy, 67.2) (Clinical Knowledge, 59.9) (College Biology, 56.6) (College Medicine, 46.5) (High School Biology, 61.5) (Medical Genetics, 69.7) (Nutrition, 63.0) (Professional Medicine, 66.1) (Virology, 46.7)}; \label{plot:medgenie_fid}

            \addplot[fill=teal2, draw=none] coordinates {(Anatomy, 58.2) (Clinical Knowledge, 57.6) (College Biology, 57.3) (College Medicine, 52.9) (High School Biology, 57.6) (Medical Genetics, 60.6) (Nutrition, 59.7) (Professional Medicine, 55.0) (Virology, 48.5)}; \label{plot:medgenie_llama}

            \addplot[fill=teal3, draw=none] coordinates {(Anatomy, 45.5) (Clinical Knowledge, 63.3) (College Biology, 65.0) (College Medicine, 65.7) (High School Biology, 68.3) (Medical Genetics, 64.6) (Nutrition, 72.8) (Professional Medicine, 66.1) (Virology, 61.8)}; \label{plot:medgenie_zephyr}

            \addplot[fill=warm1, draw=none] coordinates {(Anatomy, 50.7) (Clinical Knowledge, 53.0) (College Biology, 62.2) (College Medicine, 50.0) (High School Biology, 58.9) (Medical Genetics, 60.6) (Nutrition, 61.6) (Professional Medicine, 49.0) (Virology, 55.8)}; \label{plot:pmcllama_13}

            \addplot[fill=warm2, draw=none] coordinates {(Anatomy, 48.5) (Clinical Knowledge, 59.8) (College Biology, 53.1) (College Medicine, 50.0) (High School Biology, 62.5) (Medical Genetics, 71.8) (Nutrition, 61.3) (Professional Medicine, 53.5) (Virology, 41.2)}; \label{plot:llama2}

            \addplot[fill=warm3, draw=none] coordinates {(Anatomy, 49.3) (Clinical Knowledge, 57.2) (College Biology, 53.8) (College Medicine, 44.8) (High School Biology, 62.5) (Medical Genetics, 64.6) (Nutrition, 63.6) (Professional Medicine, 55.4) (Virology, 38.2)}; \label{plot:meditron}

            \addplot[fill=warm4, draw=none] coordinates {(Anatomy, 58.5) (Clinical Knowledge, 58.5) (College Biology, 59.7) (College Medicine, 46.8) (High School Biology, 55.2) (Medical Genetics, 76.0) (Nutrition, 56.5) (Professional Medicine, 57.4) (Virology, 42.2)}; \label{plot:vod}

            \addplot[fill=warm5, draw=none] coordinates {(Anatomy, 56.0) (Clinical Knowledge, 64.0) (College Biology, 66.4) (College Medicine, 60.5) (High School Biology, 55.2) (Medical Genetics, 68.7) (Nutrition, 56.5) (Professional Medicine, 64.9) (Virology, 45.5)}; \label{plot:zephyr-2}
            
            \end{axis}
        \end{tikzpicture}
    \end{subfigure}
    \caption{Fine-grained MedGENIE performance on MedMCQA (21 medical subjects) and MMLU-Medical (9 health and biology subjects). The frequency percentage for each subject is shown in brackets. In MedMCQA, Zephyr-$\beta$ denotes our experiment with 2-shot prompting. Conversely, for MMLU, Zephyr-$\beta$ refers to performance from \citep{DBLP:journals/corr/abs-2311-16079} in 3-shot.}
    \label{fig:topic_distribution}
\end{figure*}

\subsection{Sensitivity to Option Order}
\label{app:sensitivity_order}

Table~\ref{table:medgenie_fid_option_shuffling} illustrates the frequencies of the predicted answers after \textit{option shuffling} with 10 distinct random seeds, testing MedGENIE-FID-Flan-T5 in all benchmarks.
Note that our seed base is 0.
The results highlight a classification bias.
In MedQA, the model is inclined to default the last option, discouraging the first, while in MedMCQA, it underestimates D.
Similarly, Table~\ref{table:medgenie_icl_option_shuffling} examines the robustness of the model to shuffled options with an unsupervised pipeline.
Shuffling can impact MedGENIE-Zephyr-$\beta$ up to 4.4 accuracy points, and MedGENIE-LLaMA-2-chat up to 5.9, registering more oscillations in MMLU.

\begin{table*}[!htb]
\begin{subtable}[t]{0.48\textwidth}
\adjustbox{width=\columnwidth, valign=t}{
\begin{tabular}{lccccccc}
    \hline
    \rowcolor{lightgrey}\multicolumn{8}{c}{\textsc{MedQA (4 options)}} \\
    \hline
    \textbf{Seed} & \textbf{A} & \textbf{B} & \textbf{C} & \textbf{D} & \textbf{E} & \textbf{Acc.} & \textbf{$p$-value} \\
    \hline
    \textit{no shuffle} & \textcolor{underestimated}{294$\blacktriangle$} & \textcolor{overestimated}{340$\blacktriangle$} & 339 & \textcolor{overestimated}{297$\blacktriangle$} & 3 & 53.1 & $ 7\cdot 10^{-4}$\\
    \textbf{data} & \textbf{353}  & \textbf{309} & \textbf{346} & \textbf{265} & \textbf{0} \\
    \hdashline
    4 & \textcolor{underestimated}{201$\blacktriangledown$} & 254 & \textcolor{overestimated}{427$\blacktriangle$} & \textcolor{overestimated}{389$\blacktriangle$} & 2 & 51.1 & $< 10^{-10}$\\
    \textbf{data} & \textbf{309}  & \textbf{265} & \textbf{353} & \textbf{346} & \textbf{0} \\
    \hdashline
    11 & \textcolor{underestimated}{170$\blacktriangledown$} & 343 & 283 & \textcolor{overestimated}{475 $\blacktriangle$} & 2 & 51.9 & $< 10^{-10}$ \\
    \textbf{data} & \textbf{265} & \textbf{346} & \textbf{309} & \textbf{353} & \textbf{0} \\
    \hdashline
    13 & \textcolor{underestimated}{233$\blacktriangledown$} & 316 & 330 & \textcolor{overestimated}{469$\blacktriangle$} & 1 & 51.1 & $< 10^{-10}$ \\
    \textbf{data} & \textbf{346} & \textbf{309} & \textbf{353} & \textbf{265} & \textbf{0} \\
    \hdashline
    40 & \textcolor{underestimated}{171$\blacktriangledown$} & 346 & \textcolor{overestimated}{402$\blacktriangle$} & \textcolor{overestimated}{351$\blacktriangle$} & 3 & 52.3 & $6 \cdot 10^{-10}$\\
    \textbf{data} & \textbf{265} & \textbf{346} & \textbf{353} & \textbf{309} & \textbf{0} \\
    \hdashline
    41 & \textcolor{underestimated}{159$\blacktriangledown$} & 306 & \textcolor{overestimated}{424$\blacktriangle$} & \textcolor{overestimated}{382$\blacktriangle$} & 2 & 52.6 & $< 10^{-10}$ \\
    \textbf{data} & \textbf{265} & \textbf{309} & \textbf{353} & \textbf{346} & \textbf{0} \\
    \hdashline
    42 & \textcolor{underestimated}{303$\blacktriangledown$} & 268 & 297 & \textcolor{overestimated}{401$\blacktriangle$} & 4 & 52.2 & $ 9\cdot 10^{-4}$ \\
    \textbf{data} & \textbf{353} & \textbf{265} & \textbf{309} & \textbf{346} & \textbf{0} \\
    \hdashline
    43 & \textcolor{underestimated}{298$\blacktriangledown$} & 334 & 244 & \textcolor{overestimated}{395$\blacktriangle$} & 2 & 52.9 & $ 2\cdot 10^{-4}$ \\
    \textbf{data} & \textbf{353} & \textbf{309} & \textbf{265} & \textbf{346} & \textbf{0} \\
    \hdashline
    45 & \textcolor{underestimated}{237$\blacktriangledown$} & 306 & 247 & \textcolor{overestimated}{482$\blacktriangle$} & 1 & 51.5 & $< 10^{-10}$ \\
    \textbf{data} & \textbf{346} & \textbf{309} & \textbf{265} & \textbf{353} & \textbf{0} \\
    \hdashline
    47 & \textcolor{underestimated}{229$\blacktriangledown$} & \textcolor{overestimated}{453$\blacktriangle$} & 294 & \textcolor{overestimated}{293$\blacktriangle$} & 4 & 51.5 & $< 10^{-10}$ \\
    \textbf{data} & \textbf{346} & \textbf{353} & \textbf{309} & \textbf{265} & \textbf{0} \\
    \hdashline
    50 & \textcolor{underestimated}{157$\blacktriangledown$} & 316 & 330 & \textcolor{overestimated}{469$\blacktriangle$} & 1 & 52.0 & $< 10^{-10}$ \\
    \textbf{data} & \textbf{265} & \textbf{309} & \textbf{346} & \textbf{353} & \textbf{0} \\
    \hline 
\end{tabular}
}
\end{subtable}
\begin{subtable}[t]{0.48\textwidth}
\adjustbox{width=\columnwidth, valign=t}{
\begin{tabular}{lccccccc}
    \hline
    \rowcolor{lightgrey}\multicolumn{8}{c}{\textsc{MedQA (5 options)}} \\
    \hline
    \textbf{Seed} & \textbf{A} & \textbf{B} & \textbf{C} & \textbf{D} & \textbf{E} & \textbf{Acc.} & \textbf{$p$-value} \\
    \hline
    \textit{no shuffle} & \textcolor{underestimated}{231$\blacktriangledown$} & 274 & \textcolor{underestimated}{221$\blacktriangledown$} & 266 & \textcolor{overestimated}{281$\blacktriangle$} & 45.6 & $5 \cdot 10^{-9}$ \\
    \textbf{data} & \textbf{273}  & \textbf{277} & \textbf{252} & \textbf{269} & \textbf{202} \\
    \hdashline
    4 & \textcolor{underestimated}{157$\blacktriangledown$} & 244 & \textcolor{overestimated}{329$\blacktriangle$} & 212 & \textcolor{overestimated}{331$\blacktriangle$} & 44.1 & $< 10^{-10}$ \\
    \textbf{data} & \textbf{277}  & \textbf{252} & \textbf{273} & \textbf{202} & \textbf{269} \\
    \hdashline
    11 & \textcolor{underestimated}{144$\blacktriangledown$} & \textcolor{underestimated}{171$\blacktriangledown$} & \textcolor{underestimated}{203$\blacktriangledown$}  & \textcolor{overestimated}{405$\blacktriangle$}  & 250 & 43.3 & $< 10^{-10}$ \\
    \textbf{data} & \textbf{269}  & \textbf{202} & \textbf{277} & \textbf{273} & \textbf{252} \\
    \hdashline
    13 & \textcolor{underestimated}{138$\blacktriangledown$} & \textcolor{underestimated}{160$\blacktriangledown$} & \textcolor{underestimated}{215$\blacktriangledown$} & \textcolor{overestimated}{377$\blacktriangle$} & \textcolor{overestimated}{383$\blacktriangle$} & 44.1 & $< 10^{-10}$ \\
    \textbf{data} & \textbf{252}  & \textbf{202} & \textbf{269} & \textbf{273} & \textbf{277} \\
    \hdashline
    40 & \textcolor{underestimated}{149$\blacktriangledown$} & \textcolor{overestimated}{324$\blacktriangle$} & \textcolor{underestimated}{148$\blacktriangledown$} & 288 & \textcolor{overestimated}{364$\blacktriangle$} & 44.4 & $< 10^{-10}$ \\
    \textbf{data} & \textbf{269}  & \textbf{273} & \textbf{202} & \textbf{277} & \textbf{252} \\
    \hdashline
    41 & \textcolor{underestimated}{150$\blacktriangledown$} & 239 & \textcolor{overestimated}{305$\blacktriangle$} & 197 & \textcolor{overestimated}{382$\blacktriangle$} & 44.6 & $< 10^{-10}$ \\
    \textbf{data} & \textbf{269}  & \textbf{252} & \textbf{273} & \textbf{202} & \textbf{277} \\
    \hdashline
    42 & \textcolor{underestimated}{224$\blacktriangledown$} & \textcolor{underestimated}{173$\blacktriangledown$} & \textcolor{underestimated}{221$\blacktriangledown$} & \textcolor{overestimated}{310$\blacktriangle$} & \textcolor{overestimated}{345$\blacktriangle$} & 45.6 & $3 \cdot 10^{-9}$\\
    \textbf{data} & \textbf{273}  & \textbf{202} & \textbf{252} & \textbf{277} & \textbf{269} \\
    \hdashline
    43 & \textcolor{underestimated}{226$\blacktriangledown$} & 239  & \textcolor{underestimated}{215$\blacktriangledown$} & 194 & \textcolor{overestimated}{399$\blacktriangle$} & 45.3 & $< 10^{-10}$ \\
    \textbf{data} & \textbf{273}  & \textbf{252} & \textbf{269} & \textbf{202} & \textbf{277} \\
    \hdashline
    45 & \textcolor{underestimated}{145$\blacktriangledown$} & \textcolor{underestimated}{220$\blacktriangledown$} & \textcolor{underestimated}{193$\blacktriangledown$} & 194 & \textcolor{overestimated}{521$\blacktriangle$} & 42.3 & $< 10^{-10}$\\
    \textbf{data} & \textbf{252}  & \textbf{269} & \textbf{277} & \textbf{202} & \textbf{273} \\
    \hdashline
    47 & \textcolor{underestimated}{149$\blacktriangledown$} & \textcolor{overestimated}{341$\blacktriangle$} & \textcolor{underestimated}{239$\blacktriangledown$} & 200 & \textcolor{overestimated}{344$\blacktriangle$} & 44.9 & $< 10^{-10}$\\
    \textbf{data} & \textbf{252}  & \textbf{273} & \textbf{277} & \textbf{202} & \textbf{269} \\
    \hdashline
    50 & \textcolor{underestimated}{142$\blacktriangledown$} & 231 & \textcolor{underestimated}{209$\blacktriangledown$} & \textcolor{overestimated}{413$\blacktriangle$} & \textcolor{overestimated}{278$\blacktriangle$} & 44.4 & $< 10^{-10}$ \\
    \textbf{data} & \textbf{269}  & \textbf{252} & \textbf{277} & \textbf{273} & \textbf{202} \\
    \hline 
\end{tabular}
}
\end{subtable}\\[5mm]
\begin{subtable}[t]{0.48\textwidth}
\adjustbox{width=\columnwidth, valign=t}{
\begin{tabular}{lcccccc}
    \hline
    \rowcolor{lightgrey}\multicolumn{7}{c}{\textsc{MedMCQA}} \\
    \hline
    \textbf{Seed} & \textbf{A} & \textbf{B} & \textbf{C} & \textbf{D} & \textbf{Acc.} & \textbf{$p$-value} \\
    \hline
    \textit{no shuffle} & 1481 & 1072 & 899 & \textcolor{underestimated}{731$\blacktriangledown$} & 52.1 & $1 \cdot 10^{-5}$ \\
    \textbf{data} & \textbf{1348} & \textbf{1085} & \textbf{925} & \textbf{825} \\
    \hdashline
    4 & \textcolor{underestimated}{933$\blacktriangledown$} & 859 & \textcolor{overestimated}{1582$\blacktriangle$} & \textcolor{underestimated}{809$\blacktriangledown$} &  51.2 & $< 10^{-10}$ \\
    \textbf{data} & \textbf{1085} & \textbf{825} & \textbf{1348} & \textbf{925} \\
    \hdashline
    11 & 810 & 941 & 1021 & 1411 & 51.2 & $6 \cdot 10^{-2}$ \\
    \textbf{data} & \textbf{825} & \textbf{925} & \textbf{1085} & \textbf{1348} \\
    \hdashline
    13 & 855 & 1042 & \textcolor{overestimated}{1565$\blacktriangle$} & \textcolor{underestimated}{721$\blacktriangledown$} & 51.7 & $< 10^{-10}$ \\
    \textbf{data} & \textbf{925} & \textbf{1085} & \textbf{1348} & \textbf{825} \\
    \hdashline
    40 & 776 & 949 & \textcolor{overestimated}{1591$\blacktriangle$} & \textcolor{underestimated}{867$\blacktriangledown$} & 51.0 & $< 10^{-10}$ \\
    \textbf{data} & \textbf{825} & \textbf{925} & \textbf{1348} & \textbf{1085} \\
    \hdashline
    41 & 787 & 1010 & \textcolor{overestimated}{1584$\blacktriangle$} & \textcolor{underestimated}{802$\blacktriangledown$} & 51.4 & $< 10^{-10}$ \\
    \textbf{data} & \textbf{825} & \textbf{1085} & \textbf{1348} & \textbf{925} \\
    \hdashline
    42 & 1438 & 893 & 1058 & \textcolor{underestimated}{794$\blacktriangledown$} & 51.1 & $9 \cdot 10^{-7}$ \\
    \textbf{data} & \textbf{1348} & \textbf{825} & \textbf{1085} & \textbf{925} \\
    \hdashline
    43 & \textcolor{overestimated}{1493$\blacktriangle$} & 1068 & 834 & \textcolor{underestimated}{788$\blacktriangledown$} & 51.9 & $6 \cdot 10^{-8}$ \\
    \textbf{data} & \textbf{1348} & \textbf{1085} & \textbf{825} & \textbf{925} \\
    \hdashline
    45 & 883 & 1085 & 829 & 1386 & 51.2 & $3 \cdot 10^{-1}$\\
    \textbf{data} & \textbf{925} & \textbf{1085} & \textbf{825} & \textbf{1348} \\
    \hdashline
    47 & 866 & \textcolor{overestimated}{1603$\blacktriangle$} & 1011 & \textcolor{underestimated}{703$\blacktriangledown$} & 51.0 & $< 10^{-10}$ \\
    \textbf{data} & \textbf{925} & \textbf{1348} & \textbf{1085} & \textbf{825} \\
    \hdashline
    50 & 802 & 1053 & 921 & 1407 & 51.5 & $2 \cdot 10^{-1}$ \\
    \textbf{data} & \textbf{825} & \textbf{1085} & \textbf{925} & \textbf{1348} \\
    \hline
\end{tabular}
}
\end{subtable}
\begin{subtable}[t]{0.48\textwidth}
\adjustbox{width=\columnwidth, valign=t}{
\begin{tabular}{lcccccc}
    \hline
    \rowcolor{lightgrey}\multicolumn{7}{c}{\textsc{MMLU-Medical}} \\
    \hline
    \textbf{Seed} & \textbf{A} & \textbf{B} & \textbf{C} & \textbf{D} & \textbf{Acc.} & \textbf{$p$-value} \\
    \hline
    \textit{no shuffle} & \textcolor{overestimated}{490$\blacktriangle$} & 441 & 471 & \textcolor{underestimated}{460$\blacktriangledown$} & 59.9 & $1 \cdot 10^{-9}$ \\
    \textbf{data} & \textbf{402} & \textbf{454} & \textbf{434} & \textbf{572} \\
    \hdashline
    4 & \textcolor{underestimated}{400$\blacktriangledown$} & \textcolor{underestimated}{456$\blacktriangledown$} & \textcolor{overestimated}{597$\blacktriangle$} & 409 & 59.1 & $< 10^{-10}$ \\
    \textbf{data} & \textbf{454} & \textbf{572} & \textbf{402} & \textbf{434} \\
    \hdashline
    11 & \textcolor{underestimated}{445$\blacktriangledown$} & 420 & 473 & \textcolor{overestimated}{524$\blacktriangle$} & 58.8 & $< 10^{-10}$ \\
    \textbf{data} & \textbf{572} & \textbf{434} & \textbf{454} & \textbf{402} \\
    \hdashline
    13 & 393 & 418 & \textcolor{overestimated}{589$\blacktriangle$} & \textcolor{underestimated}{462$\blacktriangledown$} & 59.5 & $< 10^{-10}$ \\
    \textbf{data} & \textbf{434} & \textbf{454} & \textbf{402} & \textbf{572} \\
    \hdashline
    40 & \textcolor{underestimated}{435$\blacktriangledown$} & 425 & \textcolor{overestimated}{584$\blacktriangle$} & 418 & 57.5 & $< 10^{-10}$ \\
    \textbf{data} & \textbf{572} & \textbf{434} & \textbf{402} & \textbf{454}  \\
    \hdashline
    41 & \textcolor{underestimated}{428$\blacktriangledown$} & 417 & \textcolor{overestimated}{601$\blacktriangle$} & 416 & 58.5 & $< 10^{-10}$ \\
    \textbf{data} & \textbf{572} & \textbf{454} & \textbf{402} & \textbf{434}  \\
    \hdashline
    42 & \textcolor{overestimated}{472$\blacktriangle$} & \textcolor{underestimated}{457$\blacktriangledown$} & \textcolor{overestimated}{502$\blacktriangle$} & 431 & 59.7 & $8 \cdot 10^{-9}$ \\
    \textbf{data} & \textbf{402} & \textbf{572} & \textbf{454} & \textbf{434}  \\
    \hdashline
    43 & \textcolor{overestimated}{497$\blacktriangle$} & 425 & \textcolor{underestimated}{511$\blacktriangledown$} & 429 & 59.9 & $9 \cdot 10^{-7}$\\
    \textbf{data} & \textbf{402} & \textbf{454} & \textbf{572} & \textbf{434}  \\
    \hdashline
    45 & 395 & 431 & \textcolor{underestimated}{506$\blacktriangledown$} & \textcolor{overestimated}{530$\blacktriangle$} & 58.8 & $< 10^{-10}$ \\
    \textbf{data} & \textbf{434} & \textbf{454} & \textbf{572} & \textbf{402}  \\
    \hdashline
    47 & 403 & \textcolor{overestimated}{507$\blacktriangle$} & 488 & \textcolor{underestimated}{464$\blacktriangledown$} & 60.0 & $< 10^{-10}$ \\
    \textbf{data} & \textbf{434} & \textbf{402} & \textbf{454} & \textbf{572}  \\
    \hdashline
    50 & \textcolor{underestimated}{426$\blacktriangledown$} & 427 & 467 & \textcolor{overestimated}{542$\blacktriangle$} & 58.5 & $< 10^{-10}$ \\
    \textbf{data} & \textbf{572} & \textbf{454} & \textbf{434} & \textbf{402} \\
    \hline
\end{tabular}
}
\end{subtable}
\caption{Frequencies of predicted answers after option shuffling with distinct random seeds. Classification bias of MedGENIE-FID-Flan-T5 on MedQA (4 and 5 options), MedMCQA, and MMLU-Medical benchmarks. We highlight labels that are under-estimated using the color \textcolor{underestimated}{blue $\blacktriangledown$} and over-estimated using the color \textcolor{overestimated}{red $\blacktriangle$} ($\pm$10\% of the gold label frequency). Using the $\chi^2$ test, we report the p-value for the null hypothesis \textit{"the predictive distribution of the model equals the empirically observed one."}}
\label{table:medgenie_fid_option_shuffling}
\end{table*}

\begin{table*}[!htb]
\centering
\adjustbox{width=\textwidth}{
\begin{tabular}{lcccccccccccc}
    \hline
    Shuffling Seed & \textbf{4} & \textbf{11} & \textbf{13} & \textbf{40} & \textbf{41} & \textbf{42} & \textbf{43} & \textbf{45} & \textbf{47} & \textbf{50} & \textbf{Base (0)} & \textbf{AVG} \\
    \hline
    \rowcolor{lightgrey}\multicolumn{13}{c}{MedGENIE-Zephyr-$\beta$} \\
    \hline
    MedQA & 58.9 & 60.3 & 59.1 & 58.8 & \textbf{60.6} & 58.1 & 58.8 & 59.3 & 57.8 & \underline{60.5} & 59.7 & 59.3 \\ 
    MedQA (5opt) & 50.5 & 52.2 & 52.4 & 52.2 & 50.7 & 51.9 & 52.4 & \underline{54.8} & 51.3 & 51.8 & \textbf{54.9} & 52.3 \\ 
    MedMCQA & 51.0 & 50.7 & \textbf{52.6} & 51.5 & \underline{52.0} & 49.6 & 50.2 & 51.5 & 51.0 & 51.1 & 51.0 & 51.1  \\ 
    MMLU & 64.1 & 61.9 & \underline{65.1} & 63.2 & 64.0 & 63.9 & 64.9 & 62.9 & 64.9 & 62.4 & \textbf{66.1} & 63.9 \\ 
    \hline
    \rowcolor{lightgrey}\multicolumn{13}{c}{MedGENIE-LLaMA-2-chat} \\
    \hline
    MedQA & 50.4 & \textbf{52.6} & 51.9 & 51.1 & 51.5 & 51.8 & 51.6 & 51.8 & \underline{52.2} & 51.2 & \textbf{52.6} & 51.7 \\ 
    MedQA (5opt) & \underline{46.1} & 42.4 & 44.6 & 44.1 & 44.1 & 45.8 & \textbf{46.2} & 45.9 & 45.2 & 44.3 & 46.0 & 45.0 \\
    MedMCQA & 43.5 & 43.9 & \textbf{45.1} & 44.3 & 44.5 & 42.9 & 42.9 & 42.4 & \underline{44.9} & 43.7 & 44.8 & 43.9 \\ 
    MMLU & 54.4 & 53.1 & 54.8 & 53.7 & 52.9 & \underline{56.7} & 56.6 & 54.1 & 56.1 & 53.4 & \textbf{58.8} & 55.0 \\ 
    \hline
\end{tabular}
}
\caption{Classification bias of MedGENIE-Zephyr-$\beta$ and MedGENIE-LLaMA-2-chat after option shuffling with distinct random seeds. Bold and underline denote the best and second-best scores for each model.}
\label{table:medgenie_icl_option_shuffling}
\end{table*}

\section{Artificial Context Augmentation for RAG}
\label{app:context_rag}

Table~\ref{table:rag_results} shows the detailed results of our experiments aimed at evaluating the effectiveness of combining artificial data with factual data sourced from MedWiki. We compare two approaches: (i) a standard RAG pipeline, where we retrieve the top \textit{k}=5 relevant contexts for a given query, and (ii) a modified approach where we retrieve the top \textit{k}=10 most relevant contexts and rerank them to consider only the top-5 reranked contexts.

Both artificial context and MedWiki data are segmented into chunks using the \texttt{RecursiveCharacterTextSplitter} with parameters \texttt{chunk\_size=1000} and \texttt{chunk\_overlap=200} from LangChain. Each chunk is encoded using \texttt{pubmedbert-base-embeddings}.\footnote{\url{https://huggingface.co/NeuML/pubmedbert-base-embeddings}} The embeddings of each chunk are stored using Milvus as vectorDB. To determine the similarity between vectors, we employ cosine similarity. Additionally, we use BGE-large\footnote{\url{https://huggingface.co/BAAI/bge-reranker-large}} as reranker.

\begin{table*}[!htb]
\centering
\adjustbox{width=0.8\linewidth}{
\begin{tabular}{llccccc}
\hline 
\textbf{R} & \textbf{G} & \textbf{Rerank} & \textbf{LLaMA-2-chat (7B)} & \textbf{mistral-instruct (7B)}& \textbf{Zephyr-$\beta$ (7B)}\\ 
\hline      
\rowcolor{lightgrey}\multicolumn{7}{c}{MedQA} \\
\hline
4.5M & -& \XSolidBrush & 37.2& 45.1 & 50.4 \\
4.5M & - & \CheckmarkBold & 36.3 & 44.6 & 50.5 \\
4.5M & 96K & \XSolidBrush & 40.2 & 44.9 & 50.5 \\
4.5M & 96K & \CheckmarkBold & 41.4 & 45.6 & 50.8 \\
4.5M & 2M & \XSolidBrush & 40.8 & 44.4 & 51.0 \\
4.5M & 2M & \CheckmarkBold & 40.5 & 45.9 & 51.2 \\
\hline      
\rowcolor{lightgrey}\multicolumn{7}{c}{MedQA (5opt)} \\
\hline
4.5M & -& \XSolidBrush & 32.8 & 35.1 & 44.0 \\
4.5M & - & \CheckmarkBold & 32.2 & 36.8 & 44.7  \\
4.5M & 96K & \XSolidBrush & 36.5 & 37.6 & 47.8 \\
4.5M & 96K & \CheckmarkBold & 35.8 & 37.9 & 47.5 \\
4.5M & 2M & \XSolidBrush & 33.5 & 37.2 & 47.9\\
4.5M & 2M & \CheckmarkBold & 36.3 & 37.9 & 47.8 \\
\hline
\rowcolor{lightgrey}\multicolumn{7}{c}{MedMCQA} \\
\hline
4.5M & -& \XSolidBrush & 35.8 & 44.2 & 47.0 \\
4.5M & - & \CheckmarkBold & 37.3 & 44.3 & 47.1   \\
4.5M & 96K & \XSolidBrush &  40.9 & 47.4 & 50.1 \\
4.5M & 96K & \CheckmarkBold & 41.8 & 48.1 & 50.3  \\
4.5M & 2M & \XSolidBrush &43.1 & 48.6 & 50.9 \\
4.5M & 2M & \CheckmarkBold & 43.7 & 49.8 & 50.7  \\
\hline
\rowcolor{lightgrey}\multicolumn{7}{c}{MMLU (out-of-domain)}\\
\hline
4.5M & -& \XSolidBrush & 51.8 & 58.5 & 66.9  \\
4.5M & - & \CheckmarkBold & 52.0 & 58.3 & 66.4    \\
4.5M & 96K & \XSolidBrush & 52.3 & 57.9 & 66.2   \\
4.5M & 96K & \CheckmarkBold & 53.5 & 58.3 & 67.4  \\
4.5M & 2M & \XSolidBrush & 51.5 & 58.1 & 66.5 \\
4.5M & 2M & \CheckmarkBold & 53.0 & 58.4 & 67.0  \\
\hline
\end{tabular}}
\caption{Results of experiments conducted to evaluate the effectiveness of combining artificial data with factual data sourced from MedWiki. In the absence of reranking, the top-5 retrieved context chunks are concatenated. Conversely, with reranking, we initially retrieve 10 chunks and subsequently narrow down to the top-5 through reranking. Context relevance is determined using cosine similarity. Here, R represents the total number of MedWiki chunks, while G represents the number of artificial context chunks considered.}
\label{table:rag_results}
\end{table*}

\section{Relation between Context Window and Context Length}
To ensure the feasibility and effectiveness of our approach, we implement measures to maintain manageable length of generated contexts. Primarily, our methodology involves crafting optimized prompts and selecting relevant shot examples to guide the LLM, as illustrated in Figure \ref{fig:prompt_option_focused} and Figure \ref{fig:prompt_option_free}. This approach combines prompt engineering techniques with expert insights from medical professionals. Additionaly, as detailed in Section \ref{app:implementation_details}, we establish a maximum generative hyperparameter of 512 new tokens, resulting in a maximum context length of 2560 tokens when concatenated with four other contexts. This deliberate design choice ensures compatibility with the 4096-token context window of both LLaMA-2 and Zephyr-$\beta$ models. By avoiding longer context lengths, we prevent exceeding the maximum context length in a significant proportion of test samples, particularly considering the question lengths in benchmarks like MedQA. By imposing these constraints, we guarantee that our models can effectively process and generate 
responses within the specified bounds.

\section{MedWiki as Knowledge Base for RAG}
While we acknowledge that PubMed \footnote{\href{https://pubmed.ncbi.nlm.nih.gov/}{https://pubmed.ncbi.nlm.nih.gov/}} and similar sources offer a more comprehensive repository of medical knowledge, it is important to consider practical constraints, especially in low-resource settings. Hosting such extensive datasets can be prohibitive in terms of infrastructure and cost. While exploring API services is possible, they would lack the control and flexibility needed for an optimized retrieval pipeline. It is worth noting that previous work, notably by \citet{DBLP:conf/icml/LievinMJW23}, has demonstrated the quality and effectiveness of MedWiki as a retrieval corpus. As Table \ref{table:medmcqa_results} shows, MedWiki-augmented VOD achieves SOTA accuracy scores for open-book models on MedMCQA, largely surpassing other small pre-trained language models like PubmedBERT, SciBERT, and BioBERT augmented with the full PubMed dataset. One reason for this lies in the type of knowledge required to achieve high results on MedQA, MedMCQA, and MMLU-Medical benchmarks. In fact, the resolution of the targeted questions benefits particularly from the use of broad notional knowledge, rather than chunks of text extrapolated from highly specialized scientific publications, which are not always directly transferable and generalizable to other contexts, risking to introduce noise. To further validate our choice of MedWiki as the knowledge base, we conducted additional experiments using Textbooks (the medical text corpus attached to MedQA) as an alternative knowledge base for RAG. As shown in Table \ref{table:medqa_results}, the results demonstrate that LLMs augmented with Textbooks only exhibit a marginal increase in performance with respect to WedWiki, corroborating the effectiveness of the latter. In light of these findings, we opted to maintain MedWiki as the knowledge base for all benchmarks, thereby providing a common reference point and ensuring consistency across our experiments.

\section{Qualitative Examples}
\label{app:qualitative_examples}

Table~\ref{table:qualitative_example_medqa} and Table~\ref{table:qualitative_example_medmcqa} showcase qualitative examples of contexts generated by PMC-LLaMA on MedQA and MedMCQA, respectively. Concept links are highlighted with colors, while text spans that are highly effective in deducing the correct answer option are bolded.

\begin{table*}[!htb]
\small
\centering
\begin{adjustbox}{width=\linewidth}
\begin{tabularx}{\textwidth}{l|X}
  \hline
  \multicolumn{2}{c}{\textbf{MedQA [test:50]}}\\
  \hline
  \textbf{Question} & A 65-year-old male is treated for anal carcinoma with therapy including external beam radiation. How does radiation affect cancer cells?\\
  \hline
  \textbf{Options} & \hlyellow{A. Induces the formation of thymidine dimers}\newline \hlcyan{B. Induces the formation of disulfide}\newline \hlgreen{C. Induces deamination of cytosine}\newline \hlpurple{\textbf{D. Induces breaks in double-stranded DNA}} \textcolor{checkmarkgreen}{(\checkmark)}\\
  \hline
  \textbf{\#1 Option-focused context} & \hlpurple{Radiation therapy for cancer works by \textbf{damaging cellular DNA} through the formation of free radicals, which \textbf{can cause breaks in double-stranded DNA}}. \hlyellow{It \textbf{does not} induce the formation of thymidine dimers} or \hlcyan{disulfide bonds}. \hlgreen{Deamination is a chemical reaction that affects cytosine residues in DNA and RNA, but \textbf{this does not result from radiation}}.\\
  \hline
  \textbf{\#1 Option-free context} & External beam radiation therapy (EB) describes the delivery of ionizing radiation to a target volume in the body using externally generated X-rays or gamma rays. \hlpurple{Radiation is most often used for its \textbf{cytotoxic effects} on cancer cells, and generally works by \textbf{damaging cellular DNA so that it can no longer function properly}}. Normal healthy cells are also affected by radiotherapy, however, and EB may cause acute side effects as well as possibly contribute to late side effects that can appear months or even years after treatment has concluded. The extent of these normal tissue/organ effects is dependent on several factors including: total dose administered; fractionation schedule (i.e., number of treatments delivered and their frequency); overall time course over which therapy is delivered; region(s) of body being irradiated; inherent radiosensitivity of organs in close proximity to target tumor; underlying comorbidities such as diabetes mellitus type II or coronary artery disease which may independently increase risk for complication development following exposure to ionizing radiation).\\
  \hline
\end{tabularx}
\end{adjustbox}
\caption{Examples of contexts generated by PMC-LLaMA for a MedQA test set instance.}
\label{table:qualitative_example_medqa}
\end{table*}

\begin{table*}[!htb]
\small
\centering
\begin{adjustbox}{width=\linewidth}
\begin{tabularx}{\textwidth}{l|X}
  \hline
  \multicolumn{2}{c}{\textbf{MedMCQA [eval:c6365cce-507c-40f6-90a2-46b867f47b6e]}}\\
  \hline
  \textbf{Question} & Axonal transport is:\\
  \hline
  \textbf{Options} & \hlyellow{A. Antegrade}\newline \hlcyan{B. Retrograde}\newline \textbf{C. \hlyellow{Antegrade} and \hlcyan{retrograde}} \textcolor{checkmarkgreen}{(\checkmark)}\newline D. None\\
  \hline
  \textbf{\#1 Option-focused context} & Axons are long tubular structures that conduct nerve impulses. \hlyellow{\textbf{Antegrade} transport occurs in the direction of the axon's tip}, \textbf{whereas} \hlcyan{\textbf{retrograde} transport is in the opposite direction, toward the cell body}.\\
  \hline
  \textbf{\#2 Option-focused context} & The question is asking about the \textbf{direction} of transport along axons. Axons are specialized structures that conduct electrical impulses in nerve cells and can extend up to a meter or more in length. Materials such as proteins, organelles, and other cellular components need to be transpoed along the axons between the cell body and the synapses (ends of neuron that connect with those of other neurons). \textbf{Transport may occur in either direction}: \hlyellow{\textbf{Antegrade} transport refers to movement from the cell body toward the synapse}; \hlcyan{\textbf{retrograde} transport is movement from synaptic end back toward the cell body}.\\
  \hline
  \textbf{\#1 Option-free context} & Axonal transport is a part of the intracellular transport system in which substances are carried from the cell body along microtubules to the axon terminals. \textbf{It includes:}\newline \hlyellow{- Transports towards axon terminal (\textbf{antegrade})}\newline \hlcyan{- Transports back to cell body (\textbf{retrograde})}\newline \hlyellow{One example of antegrade movement is when materials, such as vesicles containing neurotransmitters, are carried from the soma through the axon and out its endings where they fuse with plasma membrane for release into synaptic cleft}. \hlcyan{An example of retrograde movement would be when sugar molecules and proteins are returned to their origin within cells. In this way, neurons can communicate information about incoming stimuli using chemical signals called neurotransmitters that act upon receptors found on dendrites or cell bodies}. \textit{[...]} \\
  \hline
\end{tabularx}
\end{adjustbox}
\caption{Examples of contexts generated by PMC-LLaMA for a MedMCQA test set instance.}
\label{table:qualitative_example_medmcqa}
\end{table*}

\section{Prompt Templates}
\label{app:prompt_templates}

Figure~\ref{fig:prompt_option_focused} and Figure~\ref{fig:prompt_option_free} report the constructed prompts for option-focused and option-free artificial context generation, with expert-crafted examples.
Figure~\ref{fig:llama3_medqa} and Figure~\ref{fig:phi_medqa} illustrate the prompt template used to perform 1-shot inference on MedQA with LLaMA-3-instruct and Phi-3-mini-instruct for artificial grounding (MedGENIE). 
Figure~\ref{fig:prompt_zephyr_medqa} and Figure~\ref{fig:prompt_llama2chat_medqa} show the prompt template used to perform 2-shot inference on MedQA with Zephyr-$\beta$ and LLaMA-2-chat for both artificial grounding and MedWiki grounding. Figure~\ref{fig:prompt_lzephyr_medqa_no_ctxs}, Figure~\ref{fig:prompt_llama_medqa_no_ctxs} and Figure~\ref{fig:prompt_pmcllama_medqa} instead show how Zephyr-$\beta$, LLaMA-2-chat and PMC-LLaMA perform 2-shot inference on MedQA without grounding, respectively. These prompt templates are also applicable to MedMCQA and MMLU, with adjustments made by interchanging the provided 2-shot examples.
Figure~\ref{fig:prompt_zephyr_medmcqa} and Figure~\ref{fig:prompt_llama2chat_medmcqa} display the templates adopted with Zephyr-$\beta$ and LLaMA-2-chat for performing 2-shot inference with either artificial grounding (MedGENIE) or MedWiki grounding on MedMCQA. The templates utilized for MMLU are presented in Figure~\ref{fig:prompt_zephyr_mmlu} and Figure~\ref{fig:prompt_llama2chat_mmlu} instead.

\begin{figure*}[!htb]
\begin{tcolorbox}[colback=lightgrey, colframe=darkgrey, arc=10pt, boxrule=0.5pt, fontupper=\ttfamily, width=1\textwidth]
\small \textcolor{purple}{\textbf{\#\#\# Question:}}\\
Chronic urethral obstruction due to benign prismatic hyperplasia can lead to the following change in kidney parenchyma:\\	
- Hyperplasia\\
- Hyperophy	\\
- Atrophy\\	
- Dyplasia\\

\textcolor{electricblue}{\textbf{\#\#\# Context:}}\\
In the scenario of chronic urethral obstruction caused by benign prostatic hyperplasia (BPH), it's essential to consider the impact on the kidney parenchyma. The continuous blockage of the urethra, a tube responsible for carrying urine from the bladder, can result in a condition known as hydronephrosis.\\

Chronic Urethral Obstruction:\\
- This condition involves a persistent blockage in the urethra, commonly caused by the non-cancerous enlargement of the prostate gland, known as benign prostatic hyperplasia (BPH).\\

Kidney Parenchyma and Hydronephrosis:\\
- The kidney parenchyma is the functional tissue of the kidney responsible for filtration and urine production.\\
- Hydronephrosis is the swelling or enlargement of the kidney due to the backup of urine caused by the obstruction. The increased pressure within the kidney can lead to changes in its structure and function.\\

Understanding the Terms:\\
- Hyperplasia: Refers to an increase in the number of cells. Consider whether this is the likely change in the kidney parenchyma due to chronic urethral obstruction.\\
- Hyperophy: This seems to be a misspelling. It might be intended as "hypertrophy," which refers to an increase in the size of cells. Consider if this is a probable outcome in the context of the described condition.\\
- Atrophy: Describes a reduction in the size or function of an organ or tissue. Consider whether atrophy is a likely change in the kidney parenchyma due to prolonged obstruction.\\
- Dyplasia: Likely a typographical error. The correct term is "dysplasia," which refers to abnormal development or growth of cells. Consider whether dysplasia is a probable outcome in the kidney parenchyma.\\

By understanding the impact of chronic urethral obstruction on the kidney and considering the definitions of the provided terms, you can deduce the potential change in kidney parenchyma.\\

\textcolor{purple}{\textbf{\#\#\# Question:}}\\
Which vitamin is supplied from only animal source:	\\
- Vitamin C \\
- Vitamin B7\\
- Vitamin B12\\
- Vitamin D\\

\textcolor{electricblue}{\textbf{\#\#\# Context:}}\\
Vitamins are essential micronutrients required by the human body for various physiological functions. They are classified into different groups, and each vitamin plays a specific role in maintaining health. The question pertains to identifying the vitamin that is exclusively supplied from animal sources.\\
- Vitamin C (Ascorbic Acid): This vitamin is found in various fruits and vegetables, particularly citrus fruits, berries, and leafy greens. It is not exclusive to animal sources.\\
- Vitamin B7 (Biotin): Biotin is a water-soluble vitamin found in a variety of foods, including meat, fish, eggs, nuts, seeds, and certain vegetables. While it is present in some animal products, it is not exclusively derived from animals.\\
- Vitamin B12 (Cobalamin): Vitamin B12 is unique in that it is primarily found in animal-based sources. It plays a crucial role in neurological function and the formation of red blood cells. Dietary sources include meat, fish, eggs, and dairy products. Vegetarians and vegans may need to supplement or rely on fortified foods to meet their B12 requirements since plant-based foods generally lack this vitamin.\\
- Vitamin D: Vitamin D is synthesized in the skin upon exposure to sunlight and is also found in some food sources. While animal products such as fatty fish, liver, and egg yolks contain vitamin D, it can also be obtained from fortified foods and supplements. Therefore, vitamin D is not exclusively derived from animal sources.\\

\textcolor{purple}{\textbf{\#\#\# Question:}}\\
\textcolor{orange}{\{\{}\textcolor{blue}{new\_question}\textcolor{orange}{\}\}}\\

\textcolor{electricblue}{\textbf{\#\#\# Context:}}
\end{tcolorbox}
\caption{Prompt template for generating an \textit{option-focused} context with PMC-LLaMA.}
\label{fig:prompt_option_focused}
\end{figure*}

\begin{figure*}[!htb]
\begin{tcolorbox}[colback=lightgrey, colframe=darkgrey, arc=10pt, boxrule=0.5pt, fontupper=\ttfamily, width=1\textwidth]
\small \textcolor{purple}{\textbf{\#\#\# Question:}}\\
Chronic urethral obstruction due to benign prismatic hyperplasia can lead to the following change in kidney parenchyma:\\

\textcolor{electricblue}{\textbf{\#\#\# Context:}}\\
Benign prostatic hyperplasia (BPH) is a common condition in aging men characterized by the non-malignant enlargement of the prostate gland. The prostate surrounds the urethra, and its enlargement can lead to various urinary symptoms such as difficulty in urination, incomplete emptying of the bladder, and increased frequency of urination.
When BPH causes chronic urethral obstruction, it can have implications for the kidneys and their parenchyma. The term "parenchyma" refers to the functional tissue of an organ, and in the case of the kidneys, it includes the renal cortex and medulla, where vital functions such as filtration, reabsorption, and secretion occur.
Chronic urethral obstruction can create back pressure on the urinary system, impacting the flow of urine from the kidneys to the bladder. This increased pressure in the urinary tract can lead to several changes in the kidney parenchyma, collectively referred to as obstructive nephropathy. Some of the key changes include:\\
Hydronephrosis: The prolonged obstruction of urine flow can cause the renal pelvis and calyces to dilate, a condition known as hydronephrosis. This dilation is a result of the accumulation of urine upstream of the obstruction, causing stretching and expansion of the renal structures.
Interstitial fibrosis: Chronic obstruction may lead to inflammation and fibrosis in the interstitium of the kidney. Fibrosis is the excessive formation of connective tissue, and in this context, it can replace normal kidney tissue, impairing its function.\\
Renal atrophy: Prolonged obstruction and the associated changes can lead to the atrophy of renal tubules and glomeruli. This atrophy is a consequence of the reduced blood flow and the pressure exerted on the kidney tissues.\\
Impaired renal function: Over time, the structural changes in the kidney parenchyma can result in impaired renal function. The ability of the kidneys to filter waste products, regulate electrolytes, and maintain fluid balance may be compromised.\\
Renal failure: In severe cases, chronic urethral obstruction due to BPH can progress to renal failure, where the kidneys are no longer able to adequately perform their vital functions. This is a serious and potentially life-threatening condition that may require medical intervention, such as surgery to relieve the obstruction.\\

\textcolor{purple}{\textbf{\#\#\# Question:}}\\
Which vitamin is supplied from only animal source:\\

\textcolor{electricblue}{\textbf{\#\#\# Context:}}\\
Vitamin B12, also known as cobalamin, is the vitamin that is primarily supplied from only animal sources. Unlike many other vitamins that can be obtained from both plant and animal sources, vitamin B12 is unique in its occurrence primarily in animal-derived foods.
Vitamin B12 plays a crucial role in various physiological processes, including the formation of red blood cells, neurological function, and DNA synthesis. It is essential for maintaining the health of nerve cells and aiding in the production of DNA and RNA.\\
Common sources of vitamin B12 from animal products include:\\
- Meat: Particularly, beef, pork, and lamb are good sources of vitamin B12.\\
- Poultry: Chicken and turkey also contain vitamin B12.\\
- Fish: Fatty fish such as salmon, trout, and tuna are good sources.\\
- Shellfish: Clams, oysters, and mussels are rich in vitamin B12.\\
- Dairy products: Milk, cheese, and eggs contain vitamin B12, although in smaller amounts compared to meat and fish.\\
Since vitamin B12 is not found in significant amounts in plant foods, individuals following a strict vegetarian or vegan diet may be at risk of B12 deficiency and may need to consider supplementation or fortified foods to meet their dietary requirements. It's an essential nutrient for overall health, and a deficiency can lead to various health issues, including anemia and neurological problems.\\

\textcolor{purple}{\textbf{\#\#\# Question:}}\\
\textcolor{orange}{\{\{}\textcolor{blue}{new\_question}\textcolor{orange}{\}\}}\\

\textcolor{electricblue}{\textbf{\#\#\# Context:}}
\end{tcolorbox}
\caption{Prompt template for generating an \textit{option-free} context with PMC-LLaMA.}
\label{fig:prompt_option_free}
\end{figure*}

\begin{figure*}[!htb]
\begin{tcolorbox}[colback=lightgrey, colframe=darkgrey, arc=10pt, boxrule=0.5pt, fontupper=\ttfamily, width=1\textwidth]
\small {\textbf{<|begin\_of\_text|><|start\_header\_id|>system<|end\_header\_id|>}} \\

You are a medical expert. Your task is to answer questions related to medical exams based on a given context by selecting the correct option. Return as output only the selected option.\textbf{<|eot\_id|><|start\_header\_id|>user<|end\_header\_id|>} \\

Select only one option. Don't explain your choice. \\

\textcolor{electricblue}{\textbf{Context:}} The scenario suggests sudden infant death syndrome (SIDS). SIDS is the unexpected, sudden death of an infant that cannot be explained by history or autopsy findings. This condition occurs when infants are placed to sleep on their stomachs and are not covered by a firm bedding. Placing babies on a firm mattress in a supine position reduces the risk of SIDS significantly. \\
Sudden infant death syndrome (SIDS) is defined as the sudden and unexplained death of an appearance healthy infant younger than one year old. Risk factors include sleeping in the prone position, soft surfaces on which to sleep, bed sharing with adults, late or no immunization for infectious diseases like HIV. \\
Sudden infant death syndrome (SIDS) is the unexpected, sudden death of a child under one year old. An autopsy does not show an explainable cause of death in cases with SIDS. Placing the child in a supine position on a firm mattress while sleeping decreases the risk of SIDS by preventing potential hazards such as soft bedding material or entrapment risks that could compromise respiration. The remaining options do not significantly reduce SIDS risks and are therefore incorrect answers to this question. \\
Sudden infant death syndrome (SIDS) is the sudden, unexplained death of an apparently healthy baby. Evidence suggests that the risk of SIDS can be reduced by: \\
- Placing babies in a supine position (on their backs) for sleep \\
- Using a firm sleep surface, such as a crib mattress covered with a fitted sheet \\
- Maintaining soft bedding and loose clothing tucked around the baby to prevent facility to wedge herself between two surfaces or get trapped or wedged between beds, furniture or other objects. [...] \\

\textcolor{purple}{\textbf{Question:}} A 3-month-old baby died suddenly at night while asleep. His mother noticed that he had died only after she awoke in the morning. No cause of death was determined based on the autopsy. Which of the following precautions could have prevented the death of the baby? \\
A. Placing the infant in a supine position on a firm mattress while sleeping \\
B. Keeping the infant covered and maintaining a high room temperature \\
C. Application of a device to maintain the sleeping position \\
D. Avoiding pacifier use during sleep \\
Answer:\textbf{<|eot\_id|><|start\_header\_id|>assistant<|end\_header\_id|>} \\

A. Placing the infant in a supine position on a firm mattress while sleeping.\\ \textbf{<|eot\_id|><|start\_header\_id|>user<|end\_header\_id|>} \\

Now help me with another question. Just select only one option as you did so far. Don't explain your choice. \\

\textcolor{electricblue}{\textbf{Context:}} \textcolor{orange}{\{\{}\textcolor{blue}{new\_context}\textcolor{orange}{\}\}}\\

\textcolor{purple}{\textbf{Question:}} \textcolor{orange}{\{\{}\textcolor{blue}{new\_question}\textcolor{orange}{\}\}}\\
\textcolor{orange}{\{\{}\textcolor{blue}{new\_option\_set}\textcolor{orange}{\}\}}\\
Answer:\textbf{<|eot\_id|><|start\_header\_id|>assistant<|end\_header\_id|>}
\end{tcolorbox}
\caption{Prompt template for question answering with LLaMA-3-Instruct on MedQA by leveraging artificial
grounding. }
\label{fig:llama3_medqa}
\end{figure*}

\begin{figure*}[!htb]
\begin{tcolorbox}[colback=lightgrey, colframe=darkgrey, arc=10pt, boxrule=0.5pt, fontupper=\ttfamily, width=1\textwidth]
\small {\textbf{<s><|user|>}} \\
You are a medical expert. Your task is to answer questions related to medical exams based on a given context by selecting the correct option. Return as output only the selected option. \\

\textcolor{electricblue}{\textbf{Context:}} The scenario suggests sudden infant death syndrome (SIDS). SIDS is the unexpected, sudden death of an infant that cannot be explained by history or autopsy findings. This condition occurs when infants are placed to sleep on their stomachs and are not covered by a firm bedding. Placing babies on a firm mattress in a supine position reduces the risk of SIDS significantly. \\
Sudden infant death syndrome (SIDS) is defined as the sudden and unexplained death of an appearance healthy infant younger than one year old. Risk factors include sleeping in the prone position, soft surfaces on which to sleep, bed sharing with adults, late or no immunization for infectious diseases like HIV. \\
Sudden infant death syndrome (SIDS) is the unexpected, sudden death of a child under one year old. An autopsy does not show an explainable cause of death in cases with SIDS. Placing the child in a supine position on a firm mattress while sleeping decreases the risk of SIDS by preventing potential hazards such as soft bedding material or entrapment risks that could compromise respiration. The remaining options do not significantly reduce SIDS risks and are therefore incorrect answers to this question. \\
Sudden infant death syndrome (SIDS) is the sudden, unexplained death of an apparently healthy baby. Evidence suggests that the risk of SIDS can be reduced by: \\
- Placing babies in a supine position (on their backs) for sleep \\
- Using a firm sleep surface, such as a crib mattress covered with a fitted sheet \\
- Maintaining soft bedding and loose clothing tucked around the baby to prevent facility to wedge herself between two surfaces or get trapped or wedged between beds, furniture or other objects. [...] \\

\textcolor{purple}{\textbf{Question:}} A 3-month-old baby died suddenly at night while asleep. His mother noticed that he had died only after she awoke in the morning. No cause of death was determined based on the autopsy. Which of the following precautions could have prevented the death of the baby? \\
A. Placing the infant in a supine position on a firm mattress while sleeping \\
B. Keeping the infant covered and maintaining a high room temperature \\
C. Application of a device to maintain the sleeping position \\
D. Avoiding pacifier use during sleep \\

Answer by returning only the correct letter among A, B, C or D. Don't explain your choice. \\
Answer:\textbf{<|end|>} \\
\textbf{<|assistant|>} \\
A\textbf{<|end|>} \\
\textbf{<|user|>} \\
\textcolor{electricblue}{\textbf{Context:}} \textcolor{orange}{\{\{}\textcolor{blue}{new\_context}\textcolor{orange}{\}\}}\\

\textcolor{purple}{\textbf{Question:}} \textcolor{orange}{\{\{}\textcolor{blue}{new\_question}\textcolor{orange}{\}\}}\\
\textcolor{orange}{\{\{}\textcolor{blue}{new\_option\_set}\textcolor{orange}{\}\}}\\

Answer by returning only the correct letter among A, B, C or D. Don't explain your choice.\\
Answer:\textbf{<|end|>} \\
\textbf{<|assistant|>}
\end{tcolorbox}
\caption{Prompt template for question answering with Phi-3-mini on MedQA by leveraging artificial grounding. }
\label{fig:phi_medqa}
\end{figure*}

\begin{figure*}[!htb]
\begin{tcolorbox}[colback=lightgrey, colframe=darkgrey, arc=10pt, boxrule=0.5pt, fontupper=\ttfamily, width=1\textwidth]
\small \textbf{<|system|>}\\
You are a medical expert. Your task is to answer questions related to medical exams based on a given context by selecting the correct option. Return as output only the selected option.\textbf{</s>}\\
\textbf{<|user|>}\\
Select only one option. Don't explain your choice.\\

\textcolor{electricblue}{\textbf{\#\#\# Context:}}\\
Nitrofurantoin is a commonly used antibiotic for the treatment of uncomplicated urinary tract infections (UTIs) in pregnant women. It is considered safe during pregnancy and is effective against common pathogens causing UTIs. Ampicillin and ceftriaxone are not the first-line choices for treating uncomplicated UTIs, and doxycycline is contraindicated in pregnancy due to potential adverse effects on fetal development.\\

\textcolor{purple}{\textbf{\#\#\# Question:}}\\
A 23-year-old pregnant woman at 22 weeks gestation presents with burning upon urination. She states it started 1 day ago and has been worsening despite drinking more water and taking cranberry extract. She otherwise feels well and is followed by a doctor for her pregnancy. Her temperature is 97.7°F (36.5°C), blood pressure is 122/77 mmHg, pulse is 80/min, respirations are 19/min, and oxygen saturation is 98\% on room air. Physical exam is notable for an absence of costovertebral angle tenderness and a gravid uterus. Which of the following is the best treatment for this patient?\\
(A) Ampicillin\\
(B) Ceftriaxone\\
(C) Doxycycline\\
(D) Nitrofurantoin\textbf{</s>}\\
\textbf{<|assistant|>}\\
(D) Nitrofurantoin.\textbf{</s>}\\
\textbf{<|user|>}\\
Select only one option. Don't explain your choice.\\

\textcolor{electricblue}{\textbf{\#\#\# Context:}}\\
Placing the infant in a supine position on a firm mattress while sleeping is the recommended precaution to reduce the risk of sudden infant death syndrome (SIDS). This position helps maintain clear airways and minimizes the risk of suffocation. Avoiding practices such as covering the infant excessively, using devices to maintain sleeping position, and prohibiting pacifier use during sleep are not recommended and may pose additional risks.\\

\textcolor{purple}{\textbf{\#\#\# Question:}}\\
A 3-month-old baby died suddenly at night while asleep. His mother noticed that he had died only after she awoke in the morning. No cause of death was determined based on the autopsy. Which of the following precautions could have prevented the death of the baby?\\
(A) Placing the infant in a supine position on a firm mattress while sleeping\\
(B) Keeping the infant covered and maintaining a high room temperature\\
(C) Application of a device to maintain the sleeping position\\
(D) Avoiding pacifier use during sleep</s>\\
\textbf{<|assistant|>}\\
(A) Placing the infant in a supine position on a firm mattress while sleeping.\textbf{</s>}\\
\textbf{<|user|>}\\
Now help me with another question. Just select only one option as you did so far. Don't explain your choice.\\

\textcolor{electricblue}{\textbf{\#\#\# Context:}}\\
\textcolor{orange}{\{\{}\textcolor{blue}{new\_context}\textcolor{orange}{\}\}}\\

\textcolor{purple}{\textbf{\#\#\# Question:}}\\
\textcolor{orange}{\{\{}\textcolor{blue}{new\_question}\textcolor{orange}{\}\}}\textbf{</s>}\\
\textbf{<|assistant|>}
\end{tcolorbox}
\caption{Prompt template for question answering with Zephyr-$\beta$ on MedQA by leveraging artificial/MedWiki grounding. We include only the best 2-shot pairs as per Table \ref{table:best_icl_prompt}. It should be noted that for MedWiki grounding, these pairs always correspond to the human-made ones (H) according to Table \ref{table:shots_medqa}.}
\label{fig:prompt_zephyr_medqa}
\end{figure*}

\begin{figure*}[!htb]
\begin{tcolorbox}[colback=lightgrey, colframe=darkgrey, arc=10pt, boxrule=0.5pt, fontupper=\ttfamily, width=1\textwidth]
\small \textbf{$[$INST$]$ <<SYS>>} \\
You are a medical expert. Your task is to answer questions related to medical exams based on a given context. Answer as concise as possible. Your answer must be always a string of one line starting with "The answer is", followed by your final choice. Nothing more.\\
\textbf{<</SYS>>}\\
Make a choice based on the context and question. Take the following two questions as examples.\\

\textbf{\# Example 1}\\
\textcolor{electricblue}{\textbf{\#\#\# Context:}}\\
Most outpatient physicians treat asymptomatic bacteriuria with sulfate-based cephalosporins such as nitrofurantoin (100 mg BID for 7 days) or cephalexin (500mg tid for 7 days). Both drugs are considered safe during pregnancy.\\

\textcolor{purple}{\textbf{\#\#\# Question:}}\\
A 23-year-old pregnant woman at 22 weeks gestation presents with burning upon urination. She states it started 1 day ago and has been worsening despite drinking more water and taking cranberry extract. She otherwise feels well and is followed by a doctor for her pregnancy. Her temperature is 97.7°F (36.5°C), blood pressure is 122/77 mmHg, pulse is 80/min, respirations are 19/min, and oxygen saturation is 98\% on room air. Physical exam is notable for an absence of costovertebral angle tenderness and a gravid uterus. Which of the following is the best treatment for this patient?\\
(A) Ampicillin\\
(B) Ceftriaxone\\
(C) Doxycycline\\
(D) Nitrofurantoin\\

The answer is (D) Nitrofurantoin.\\

\textbf{\# Example 2}\\
\textcolor{electricblue}{\textbf{\#\#\# Context:}}\\
Sudden infant death syndrome (SIDS) is the unexpected, sudden death of a child under one year old. An autopsy does not show an explainable cause of death in cases with SIDS. Placing the child in a supine position on a firm mattress while sleeping decreases the risk of SIDS by preventing potential hazards such as soft bedding material or entrapment risks that could compromise respiration.\\

\textcolor{purple}{\textbf{\#\#\# Question:}}\\
A 3-month-old baby died suddenly at night while asleep. His mother noticed that he had died only after she awoke in the morning. No cause of death was determined based on the autopsy. Which of the following precautions could have prevented the death of the baby?\\
(A) Placing the infant in a supine position on a firm mattress while sleeping\\
(B) Keeping the infant covered and maintaining a high room temperature\\
(C) Application of a device to maintain the sleeping position\\
(D) Avoiding pacifier use during sleep\\

The answer is (A) Placing the infant in a supine position on a firm mattress while sleeping.\\

Now, help me with this question. Remember to answer with just a string of one line starting with "The answer is" as shown by the previous examples.\\

\textcolor{electricblue}{\textbf{\#\#\# Context:}}\\
\textcolor{orange}{\{\{}\textcolor{blue}{new\_context}\textcolor{orange}{\}\}}\\

\textcolor{purple}{\textbf{\#\#\# Question:}}\\
\textcolor{orange}{\{\{}\textcolor{blue}{new\_question}\textcolor{orange}{\}\}} \textbf{$[$/INST$]$}
\end{tcolorbox}
\caption{Prompt template for question answering with LLaMA-2-chat on MedQA by leveraging artificial/MedWiki grounding. We include only the best 2-shot pairs as per Table \ref{table:best_icl_prompt}. It should be noted that for MedWiki grounding, these pairs always correspond to the human-made ones (H) according to Table \ref{table:shots_medqa}.}
\label{fig:prompt_llama2chat_medqa}
\end{figure*}

\begin{figure*}[!htb]
\begin{tcolorbox}[colback=lightgrey, colframe=darkgrey, arc=10pt, boxrule=0.5pt, fontupper=\ttfamily, width=1\textwidth]
\textbf{<|system|>}\\
You are a medical expert. Your task is to answer questions related to medical exams by selecting the correct option. Return as output only the selected option.\textbf{</s>}\\
\textbf{<|user|>}\\
Select only one option. Don't explain your choice.\\

\textcolor{purple}{\textbf{\#\#\# Question:}}\\
A 23-year-old pregnant woman at 22 weeks gestation presents with burning upon urination. She states it started 1 day ago and has been worsening despite drinking more water and taking cranberry extract. She otherwise feels well and is followed by a doctor for her pregnancy. Her temperature is 97.7°F (36.5°C), blood pressure is 122/77 mmHg, pulse is 80/min, respirations are 19/min, and oxygen saturation is 98\% on room air. Physical exam is notable for an absence of costovertebral angle tenderness and a gravid uterus. Which of the following is the best treatment for this patient?\\
(A) Ampicillin\\
(B) Ceftriaxone\\
(C) Doxycycline\\
(D) Nitrofurantoin\textbf{</s>}\\
\textbf{<|assistant|>}\\
(D) Nitrofurantoin.\textbf{</s>}\\
\textbf{<|user|>}\\
Select only one option. Don't explain your choice.\\

\textcolor{purple}{\textbf{\#\#\# Question:}}\\
A 3-month-old baby died suddenly at night while asleep. His mother noticed that he had died only after she awoke in the morning. No cause of death was determined based on the autopsy. Which of the following precautions could have prevented the death of the baby?\\
(A) Placing the infant in a supine position on a firm mattress while sleeping\\
(B) Keeping the infant covered and maintaining a high room temperature\\
(C) Application of a device to maintain the sleeping position\\
(D) Avoiding pacifier use during sleep\textbf{</s>}\\
\textbf{<|assistant|>}\\
(A) Placing the infant in a supine position on a firm mattress while sleeping.\textbf{</s>}\\
\textbf{<|user|>}\\
Now help me with another question. Just select only one option as you did so far. Don't explain your choice.\\

\textcolor{purple}{\textbf{\#\#\# Question:}}\\
\textcolor{orange}{\{\{}\textcolor{blue}{new\_question}\textcolor{orange}{\}\}}\textbf{</s>}\\
\textbf{<|assistant|>}\\
\end{tcolorbox}
\caption{Prompt template for question answering with Zephyr-$\beta$ \textbf{without grounding} on MedQA.}
\label{fig:prompt_lzephyr_medqa_no_ctxs}
\end{figure*}

\begin{figure*}[!htb]
\begin{tcolorbox}[colback=lightgrey, colframe=darkgrey, arc=10pt, boxrule=0.5pt, fontupper=\ttfamily, width=1\textwidth]
\textbf{\textbf{$[$INST$]$}<<SYS>>}\\
You are a medical expert. Your task is to answer questions related to medical exams. Answer as concise as possible. Your answer must be always a string of one line starting with "The answer is", followed by your final choice. Nothing more.\\
\textbf{<</SYS>>}\\
\textbf{\# Example 1}\\
\textcolor{purple}{\textbf{\#\#\# Question:}}\\
A 23-year-old pregnant woman at 22 weeks gestation presents with burning upon urination. She states it started 1 day ago and has been worsening despite drinking more water and taking cranberry extract. She otherwise feels well and is followed by a doctor for her pregnancy. Her temperature is 97.7°F (36.5°C), blood pressure is 122/77 mmHg, pulse is 80/min, respirations are 19/min, and oxygen saturation is 98\% on room air. Physical exam is notable for an absence of costovertebral angle tenderness and a gravid uterus. Which of the following is the best treatment for this patient?\\
A. Ampicillin\\
B. Ceftriaxone\\
C. Doxycycline\\
D. Nitrofurantoin\\

\textcolor{checkmarkgreen}{\textbf{\#\#\# Answer:}} \\
D. Nitrofurantoin\\

\textbf{\# Example 2}\\
\textcolor{purple}{\textbf{\#\#\# Question:}}\\
A 3-month-old baby died suddenly at night while asleep. His mother noticed that he had died only after she awoke in the morning. No cause of death was determined based on the autopsy. Which of the following precautions could have prevented the death of the baby?\\
(A) Placing the infant in a supine position on a firm mattress while sleeping\\
(B) Keeping the infant covered and maintaining a high room temperature\\
(C) Application of a device to maintain the sleeping position\\
(D) Avoiding pacifier use during sleep\\

The answer is (A) Placing the infant in a supine position on a firm mattress while sleeping.\\

Now, help me with this question. Remember to answer with just a string of one line starting with "The answer is" as shown by the previous examples.\\

\textcolor{purple}{\textbf{\#\#\# Question:}}\\
\textcolor{orange}{\{\{}\textcolor{blue}{new\_question}\textcolor{orange}{\}\}}\textbf{$[$/INST$]$}\\
\end{tcolorbox}
\caption{Prompt template for question answering with LLaMA-2-chat \textbf{without grounding} on MedQA.}
\label{fig:prompt_llama_medqa_no_ctxs}
\end{figure*}

\begin{figure*}[!htb]
\begin{tcolorbox}[colback=lightgrey, colframe=darkgrey, arc=10pt, boxrule=0.5pt, fontupper=\ttfamily, width=1\textwidth]
\small \textbf{\#\#\# Instruction:} \\
Make a choice based on the question and options. Take the following two questions as examples.\\

\textbf{\# Few-shot Example 1}\\
\textcolor{purple}{\textbf{\#\#\# Question:}}\\
A 23-year-old pregnant woman at 22 weeks gestation presents with burning upon urination. She states it started 1 day ago and has been worsening despite drinking more water and taking cranberry extract. She otherwise feels well and is followed by a doctor for her pregnancy. Her temperature is 97.7°F (36.5°C), blood pressure is 122/77 mmHg, pulse is 80/min, respirations are 19/min, and oxygen saturation is 98\% on room air. Physical exam is notable for an absence of costovertebral angle tenderness and a gravid uterus. Which of the following is the best treatment for this patient?\\
A. Ampicillin\\
B. Ceftriaxone\\
C. Doxycycline\\
D. Nitrofurantoin\\

\textcolor{checkmarkgreen}{\textbf{\#\#\# Answer:}} \\
D. Nitrofurantoin\\

\textbf{\# Few-shot Example 2}\\
\textcolor{purple}{\textbf{\#\#\# Question:}}\\
A 3-month-old baby died suddenly at night while asleep. His mother noticed that he had died only after she awoke in the morning. No cause of death was determined based on the autopsy. Which of the following precautions could have prevented the death of the baby?\\
A. Placing the infant in a supine position on a firm mattress while sleeping\\
B. Keeping the infant covered and maintaining a high room temperature\\
C. Application of a device to maintain the sleeping position\\
D. Avoiding pacifier use during sleep\\

\textcolor{checkmarkgreen}{\textbf{\#\#\# Answer:}}\\
A. Placing the infant in a supine position on a firm mattress while sleeping\\

Now help me with another question\\

\textcolor{purple}{\textbf{\#\#\# Question:}}\\
\textcolor{orange}{\{\{}\textcolor{blue}{new\_question}\textcolor{orange}{\}\}}\\

\textcolor{checkmarkgreen}{\textbf{\#\#\# Answer:}}\\
\end{tcolorbox}
\caption{Prompt template for question answering with PMC-LLaMA \textbf{without grounding} on MedQA.}
\label{fig:prompt_pmcllama_medqa}
\end{figure*}

\begin{figure*}[!htb]
\begin{tcolorbox}[colback=lightgrey, colframe=darkgrey, arc=10pt, boxrule=0.5pt, fontupper=\ttfamily, width=1\textwidth]
\small \textbf{<|system|>}\\
You are a medical expert. Your task is to answer questions related to medical exams based on a given context by selecting the correct option. Return as output only the selected option.\textbf{</s>}\\
\textbf{<|user|>}\\
Select only one option. Don't explain your choice.\\

\textcolor{electricblue}{\textbf{\#\#\# Context:}}\\
Antibiotic prophylaxis for infective endocarditis is indicated in individuals with predisposing cardiac conditions. In this scenario, determining if an isolated secundum ASD and mitral valve prolapse without regurgitation are associated with the potential risk of developing infective endocarditis requires further information. The presence of a prior coronary aery bypass graft and coarctation of aoa are both established indications for antibiotic prophylaxis due to their association with infective endocarditis risk.\\

\textcolor{purple}{\textbf{\#\#\# Question:}}\\
Antiboiotic Prophylaxis for infective endocarditis is indicated in:\\
(A) Isolated secundum ASD \\
(B) Mitral valve prolapse without regurgitation\\
(C) Prior coronary aery bypass graft\\
(D) Coarctation of aoa\textbf{</s>}\\
\textbf{<|assistant|>}\\
(D) Coarctation of aoa.\textbf{</s>}\\
\textbf{<|user|>}\\
Select only one option. Don't explain your choice.\\

\textcolor{electricblue}{\textbf{\#\#\# Context:}}\\
The Anterolateral portal is also known as the lateral portal. It is used for viewing the patellofemoral joint, inserting probe or laser for soft-tissue procedures.\\

\textcolor{purple}{\textbf{\#\#\# Question:}}\\
Anterolateral ahroscopy of knee is for:\\
(A) To see patellofemoral aiculation\\
(B) To see the posterior cruciate ligament\\
(C) To see the anterior poion of lateral meniscus\\
(D) To see the periphery of the posterior horn of medial meniscus\textbf{</s>}\\
\textbf{<|assistant|>}\\
(A) To see patellofemoral aiculation.\textbf{</s>}\\
\textbf{<|user|>}\\
Now help me with another question. Just select only one option as you did so far. Don't explain your choice.\\

\textcolor{electricblue}{\textbf{\#\#\# Context:}}\\
\textcolor{orange}{\{\{}\textcolor{blue}{new\_context}\textcolor{orange}{\}\}}\\

\textcolor{purple}{\textbf{\#\#\# Question:}}\\
\textcolor{orange}{\{\{}\textcolor{blue}{new\_question}\textcolor{orange}{\}\}}
\end{tcolorbox}
\caption{Prompt template for question answering with Zephyr-$\beta$ on MedMCQA by leveraging artificial/MedWiki grounding. We include only the best 2-shot pairs as per Table \ref{table:best_icl_prompt}. It should be noted that for MedWiki grounding, these pairs always correspond to the human-made ones (H) according to Table \ref{table:shots_medqa}.}
\label{fig:prompt_zephyr_medmcqa}
\end{figure*}

\begin{figure*}[!htb]
\begin{tcolorbox}[colback=lightgrey, colframe=darkgrey, arc=10pt, boxrule=0.5pt, fontupper=\ttfamily, width=1\textwidth]
\small \textbf{$[$INST$]$ <<SYS>>}\\
You are a medical expert. Your task is to answer questions related to medical exams based on a given context. Answer as concise as possible. Your answer must be always a string of one line starting with "The answer is", followed by your final choice. Nothing more.\\
\textbf{<</SYS>>}\\
Make a choice based on the context and question. Take the following two questions as examples.\\

\textbf{\# Example 1}\\
\textcolor{electricblue}{\textbf{\#\#\# Context:}}\\
Hyperviscosity is a condition where the blood becomes abnormally thick, hindering its ability to flow properly. Cryoglobulinemia is a condition characterized by abnormal antibodies in the blood (antibodies are specialized cells that recognize and attack foreign invaders). These abnormal antibodies become solid at cold temperatures and lead to clumping of red blood cells, an increase in viscosity, and subsequent obstruction of small vessels.\\

\textcolor{purple}{\textbf{\#\#\# Question:}}\\
Hyper viscosity is seen in\\
(A) Cryoglobulinemia\\
(B) Multiple myeloma\\
(C) MGUS\\
(D) Lymphoma\\

The answer is (A) Cryoglobulinemia.\\

\textbf{\# Example 2}\\
\textcolor{electricblue}{\textbf{\#\#\# Context:}}\\
Inversion of the foot refers to a foot in which its sole faces medially. Since the plantar surface of the foot is in contact with the ground at all times, this condition occurs when one or more muscles responsible for moving it become tight or weak, resulting in an alteration in normal alignment.\\

\textcolor{purple}{\textbf{\#\#\# Question:}}\\
In inversion of the foot, the sole will face:\\
(A) Upwards\\
(B) Downwards\\
(C) Laterally\\
(D) Medially\\

The answer is (D) Medially.\\

Now, help me with this question. Remember to answer with just a string of one line starting with "The answer is" as shown by the previous examples.\\

\textcolor{electricblue}{\textbf{\#\#\# Context:}}\\
\textcolor{orange}{\{\{}\textcolor{blue}{new\_context}\textcolor{orange}{\}\}}\\

\textcolor{purple}{\textbf{\#\#\# Question:}}\\
\textcolor{orange}{\{\{}\textcolor{blue}{new\_question}\textcolor{orange}{\}\}}\textbf{$[$/INST$]$}
\end{tcolorbox}
\caption{Prompt template for question answering with LLaMA-2-chat on MedMCQA by leveraging artificial/MedWiki grounding. We include only the best 2-shot pairs as per Table \ref{table:best_icl_prompt}. It should be noted that for MedWiki grounding, these pairs always correspond to the human-made ones (H) according to Table \ref{table:shots_medqa}.}
\label{fig:prompt_llama2chat_medmcqa}
\end{figure*}

\begin{figure*}[!htb]
\begin{tcolorbox}[colback=lightgrey, colframe=darkgrey, arc=10pt, boxrule=0.5pt, fontupper=\ttfamily, width=1\textwidth]
\small \textbf{<|system|>}\\
You are a medical expert. Your task is to answer questions related to medical exams based on a given context by selecting the correct option. Return as output only the selected option.\textbf{</s>}\\
\textbf{<|user|>}\\
Select only one option. Don't explain your choice.\\

\textcolor{electricblue}{\textbf{\#\#\# Context:}}\\
Hyperviscosity is a condition where the blood becomes abnormally thick, hindering its ability to flow properly. Cryoglobulinemia is a condition characterized by abnormal antibodies in the blood (antibodies are specialized cells that recognize and attack foreign invaders). These abnormal antibodies become solid at cold temperatures and lead to clumping of red blood cells, an increase in viscosity, and subsequent obstruction of small vessels.\\

\textcolor{purple}{\textbf{\#\#\# Question:}}\\
Hyper viscosity is seen in\\
(A) Cryoglobulinemia\\
(B) Multiple myeloma\\
(C) MGUS\\
(D) Lymphoma\textbf{</s>}\\
\textbf{<|assistant|>}\\
(A) Cryoglobulinemia.\textbf{</s>}\\
\textbf{<|user|>}\\
Select only one option. Don't explain your choice.\\

\textcolor{electricblue}{\textbf{\#\#\# Context:}}\\
Inversion of the foot refers to a foot in which its sole faces medially. Since the plantar surface of the foot is in contact with the ground at all times, this condition occurs when one or more muscles responsible for moving it become tight or weak, resulting in an alteration in normal alignment.\\

\textcolor{purple}{\textbf{\#\#\# Question:}}\\
In inversion of the foot, the sole will face:\\
(A) Upwards\\
(B) Downwards\\
(C) Laterally\\
(D) Medially\textbf{</s>}\\
\textbf{<|assistant|>}\\
(D) Medially.\textbf{</s>}\\
\textbf{<|user|>}\\
Now help me with another question. Just select only one option as you did so far. Don't explain your choice.\\

\textcolor{electricblue}{\textbf{\#\#\# Context:}}\\
\textcolor{orange}{\{\{}\textcolor{blue}{new\_context}\textcolor{orange}{\}\}}\\

\textcolor{purple}{\textbf{\#\#\# Question:}}\\
\textcolor{orange}{\{\{}\textcolor{blue}{new\_question}\textcolor{orange}{\}\}}
\end{tcolorbox}
\caption{Prompt template for question answering with Zephyr-$\beta$ on MMLU-Medical by leveraging artificial/MedWiki grounding. We include only the best 2-shot pairs as per Table \ref{table:best_icl_prompt}. It should be noted that for MedWiki grounding, these pairs always correspond to the human-made ones (H) according to Table \ref{table:shots_medqa}.}
\label{fig:prompt_zephyr_mmlu}
\end{figure*}

\begin{figure*}[!htb]
\begin{tcolorbox}[colback=lightgrey, colframe=darkgrey, arc=10pt, boxrule=0.5pt, fontupper=\ttfamily, width=1\textwidth]
\small \textbf{$[$INST$]$ <<SYS>>}\\
You are a medical expert. Your task is to answer questions related to medical exams based on a given context. Answer as concise as possible. Your answer must be always a string of one line starting with "The answer is", followed by your final choice. Nothing more.\\
\textbf{<</SYS>>}\\
Make a choice based on the context and question. Take the following two questions as examples.\\

\textbf{\# Example 1}\\
\textcolor{electricblue}{\textbf{\#\#\# Context:}}\\
Antibiotic prophylaxis for infective endocarditis is indicated in individuals with predisposing cardiac conditions. In this scenario, determining if an isolated secundum ASD and mitral valve prolapse without regurgitation are associated with the potential risk of developing infective endocarditis requires further information. The presence of a prior coronary aery bypass graft and coarctation of aoa are both established indications for antibiotic prophylaxis due to their association with infective endocarditis risk.\\

\textcolor{purple}{\textbf{\#\#\# Question:}}\\
Antiboiotic Prophylaxis for infective endocarditis is indicated in:\\
(A) Isolated secundum ASD\\
(B) Mitral valve prolapse without regurgitation\\
(C) Prior coronary aery bypass graft\\
(D) Coarctation of aoa\\

The answer is (D) Coarctation of aoa.\\

\textbf{\# Example 2}\\
\textcolor{electricblue}{\textbf{\#\#\# Context:}}\\
The Anterolateral portal is also known as the lateral portal. It is used for viewing the patellofemoral joint, inserting probe or laser for soft-tissue procedures.\\

\textcolor{purple}{\textbf{\#\#\# Question:}}\\
Anterolateral ahroscopy of knee is for:\\
(A) To see patellofemoral aiculation\\
(B) To see the posterior cruciate ligament\\
(C) To see the anterior poion of lateral meniscus\\
(D) To see the periphery of the posterior horn of medial meniscus\\

The answer is (A) To see patellofemoral aiculation.\\

Now, help me with this question. Remember to answer with just a string of one line starting with "The answer is" as shown by the previous examples.\\

\textcolor{electricblue}{\textbf{\#\#\# Context:}}\\
\textcolor{orange}{\{\{}\textcolor{blue}{new\_context}\textcolor{orange}{\}\}}\\

\textcolor{purple}{\textbf{\#\#\# Question:}}\\
\textcolor{orange}{\{\{}\textcolor{blue}{new\_question}\textcolor{orange}{\}\}}\textbf{$[$/INST$]$}
\end{tcolorbox}
\caption{Prompt template for question answering with LLaMA-2-chat on MMLU-Medical. We include only the best 2-shot pairs as per Table \ref{table:best_icl_prompt}. It should be noted that for MedWiki grounding, these pairs always correspond to the human-made ones (H) according to Table \ref{table:shots_medqa}.}
\label{fig:prompt_llama2chat_mmlu}
\end{figure*}

\end{document}